\documentclass[manuscript,screen,nonacm]{acmart}

\makeatletter
\def\@authorsaddresses{}
\makeatother

\usepackage{subcaption}
\usepackage{multirow}
\usepackage{soul}

\begin{document}

\title{Who Defines Fairness? Target-Based Prompting for Demographic Representation in Generative Models}

\author{Marzia Binta Nizam}\thanks{Corresponding Author: manizam@ucsc.edu}
\author{James Davis}

\affiliation{%
  \institution{University of California, Santa Cruz}
  \city{Santa Cruz}
  \state{California}
  \country{USA}
}

\begin{abstract}
Text-to-image (T2I) models like Stable Diffusion and DALL·E have made generative AI widely accessible, yet recent studies reveal that these systems often replicate societal biases, particularly in how they depict demographic groups across professions. Prompts such as “doctor” or “CEO” frequently yield lighter-skinned outputs, while lower-status roles like “janitor” show more diversity, reinforcing stereotypes. Existing mitigation methods typically require retraining or curated datasets, making them inaccessible to most users. We propose a lightweight, inference-time framework that mitigates representational bias through prompt-level intervention without modifying the underlying model. Instead of assuming a single definition of fairness, our approach allows users to select among multiple fairness specifications—ranging from simple choices such as a uniform distribution to more complex definitions informed by a large language model (LLM) that cites sources and provides confidence estimates. These distributions guide the construction of demographic-specific prompt variants in the corresponding proportions, and we evaluate alignment by auditing adherence to the declared target and measuring the resulting skin tone distribution rather than assuming uniformity as “fairness.” Across 36 prompts spanning 30 occupations and 6 non-occupational contexts, our method shifts observed skin-tone outcomes in directions consistent with the declared target, and reduces deviation from targets when the target is defined directly in skin-tone space (fallback). This work demonstrates how fairness interventions can be made transparent, controllable, and usable at inference time, directly empowering users of generative AI.

\end{abstract}



\maketitle
\vspace{-0.9em}
\section{Introduction}
Text-to-image (T2I) models \cite{chen2023pixart,ramesh2021zero,rombach2022high} such as
Stable Diffusion and DALL·E enable users to generate high-quality images from natural language prompts. Despite this accessibility, growing evidence shows that these systems
often replicate and amplify societal biases. For instance, prompts such as ``doctor'' or ``CEO'' frequently yield images dominated by lighter skin tones, while lower-prestige roles (e.g., ``janitor'' or ``construction worker'') tend to yield a higher share of darker skin tones, mirroring long-standing visual stereotypes tied to socioeconomic status.

Our analysis of 30 occupations illustrates the extent of this imbalance. For high-status
roles such as ``doctor'' and ``CEO,'' over two-thirds of generated images depict individuals
with light skin tones, with darker skin tones largely absent. In contrast, lower-status roles
produce a noticeably higher share of darker-skinned individuals. Even ostensibly neutral
prompts such as ``a smiling person'' show strong skew toward lighter skin tones, suggesting
that bias extends beyond occupational contexts.

While bias in T2I systems is now well established---often attributed to skewed 
training data and model optimization pipelines---many mitigation approaches 
focus on modifying the models themselves, such as retraining on curated datasets 
or fine-tuning with specialized objectives~\cite{he2024debiasing, 
huang2025debiasing, kim2025rethinking}. These interventions are costly, opaque, and typically inaccessible to end users.  Inference-time approaches reduce this cost but 
typically assume a single, fixed notion of what ``fair'' representation means, 
leaving limited room for users to adapt interventions to different cultural, 
occupational, or application-specific contexts \cite{friedrich2023fair, bansal2022Enti, zhang2023iti} .

In this work, we propose a lightweight, user-centric framework that intervenes
at the prompt level using a large language model (LLM) as an intermediary agent.
Our framework differs from prior approaches in three fundamental ways. First,
rather than assuming that fairness always equals uniformity, we introduce a
\textbf{target-based prompting} mechanism that allows users to select from a
\emph{spectrum of fairness specifications}---ranging from uniform parity to
distributions informed by real-world demographic statistics or user-defined
sociopolitical goals---making the choice of what counts as fair explicit and
contestable.  Second, rather than relying on rigid, manually selected demographic
terms, our system leverages an LLM as a \textbf{knowledge engine} that
dynamically identifies relevant demographic dimensions for a given prompt,
retrieves evidence-based proportions with cited sources, and yields confidence
estimates, ensuring that interventions are \emph{contextually grounded} rather
than arbitrary. Third, by maintaining an explicit \emph{declared target
distribution}, our framework enables quantitative auditing of the gap between
the intended demographic balance and the generated outcome---a transparency
property absent from existing black-box intervention approaches.

\begin{figure}[t]
    \centering
    \includegraphics[width=\linewidth]{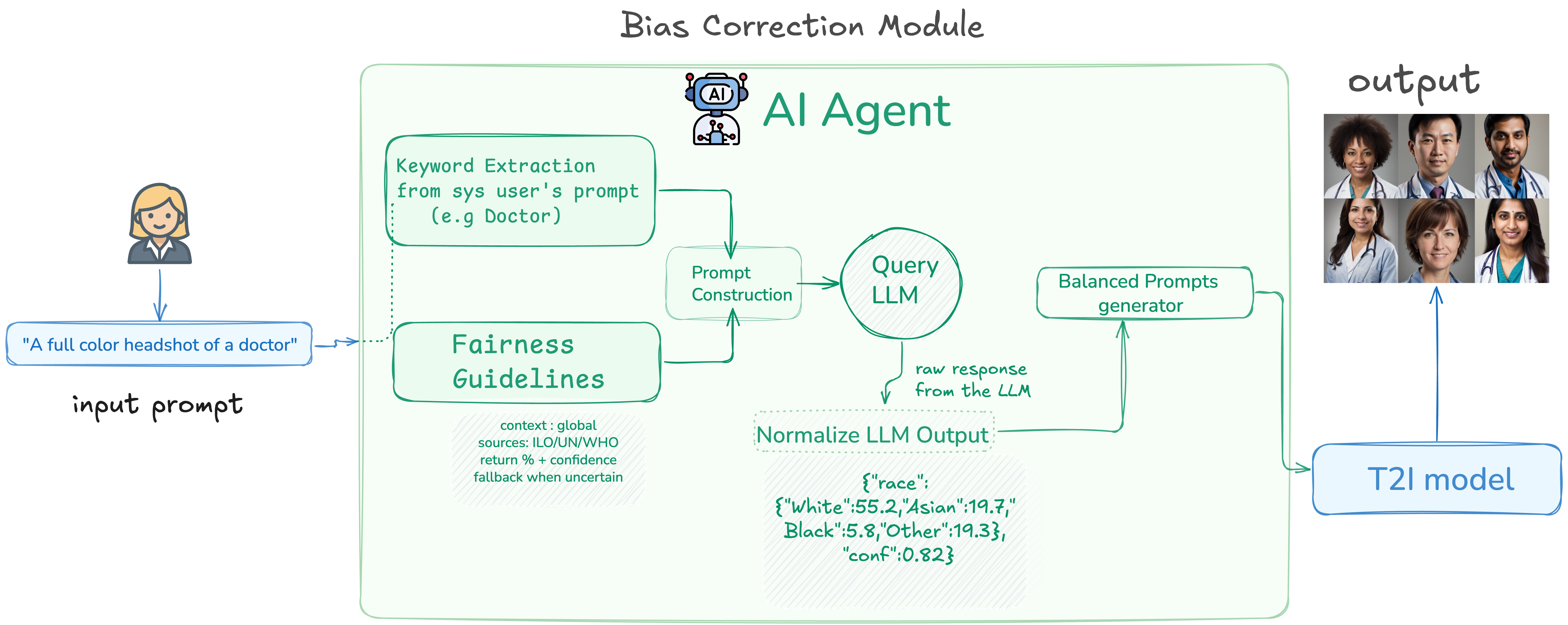}
    \caption{Visual pipeline of our demographically-aware image generation system. Given an input prompt (e.g., “A full-color headshot of a doctor”), an agent extracts the relevant concept and queries an LLM for a demographic target distribution under a user-specified scope. The LLM returns demographic proportions along with confidence and cited sources, which are logged for transparency and auditing. The target distribution is then used to generate subgroup prompt variants, which are sampled to produce a target-conditioned image set using a fixed text-to-image diffusion model. When statistics are unavailable or low-confidence, the pipeline falls back to a predefined target (uniform Fitzpatrick by default).}
    \label{fig:system_pipeline}
\end{figure}

We evaluate our framework across four T2I models (SD Realistic Vision v5.1, 
SDXL Turbo, SD~1.5, and DALL-E~2) and 36 prompts spanning 30 occupations and 
6 non-occupational contexts. Our method requires no model fine-tuning and 
integrates with any T2I pipeline via its text interface. Results demonstrate 
that target-conditioned prompting substantially shifts measured skin-tone 
outcomes relative to baseline generation, consistently outperforming 
inference-time baselines across all occupational status groups and model 
checkpoints. Beyond skin tone, the same mechanism generalizes to other 
attributes such as gender presentation, without sacrificing visual fidelity.

\begin{figure}[t]
    \centering
    \begin{tabular}{@{}c c@{}}
        
     \includegraphics[width=0.95\linewidth]{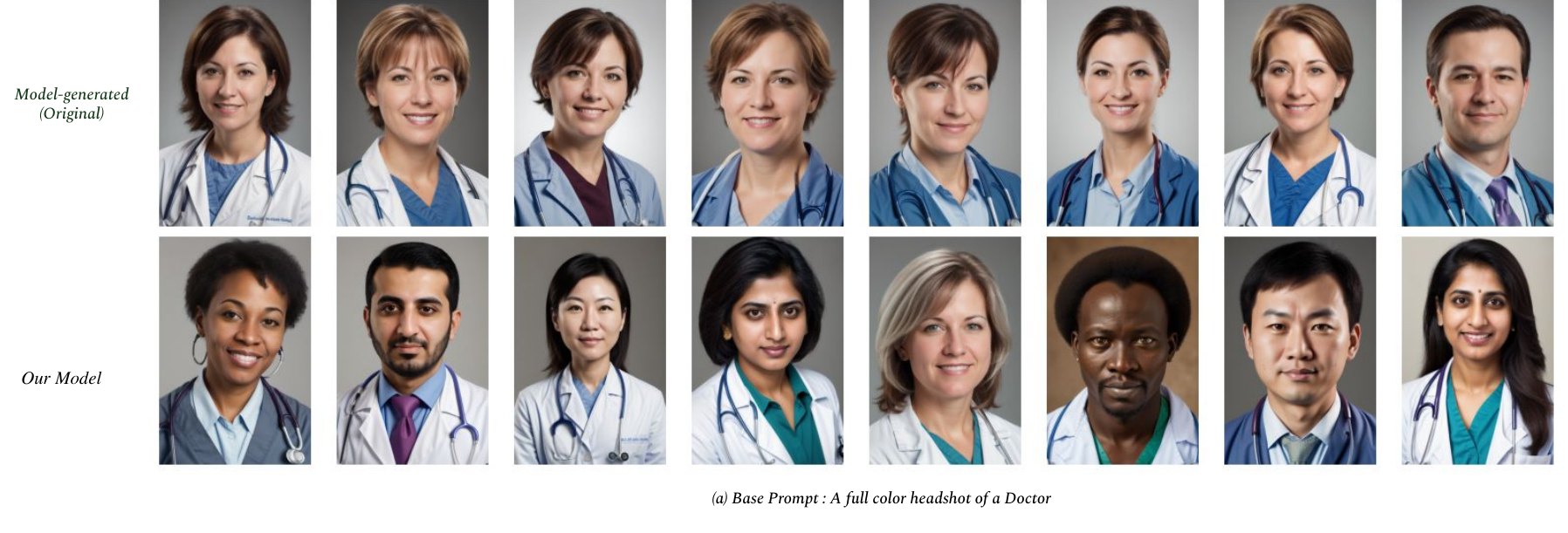} \\[-0.9em]
    \includegraphics[width=0.95\linewidth]{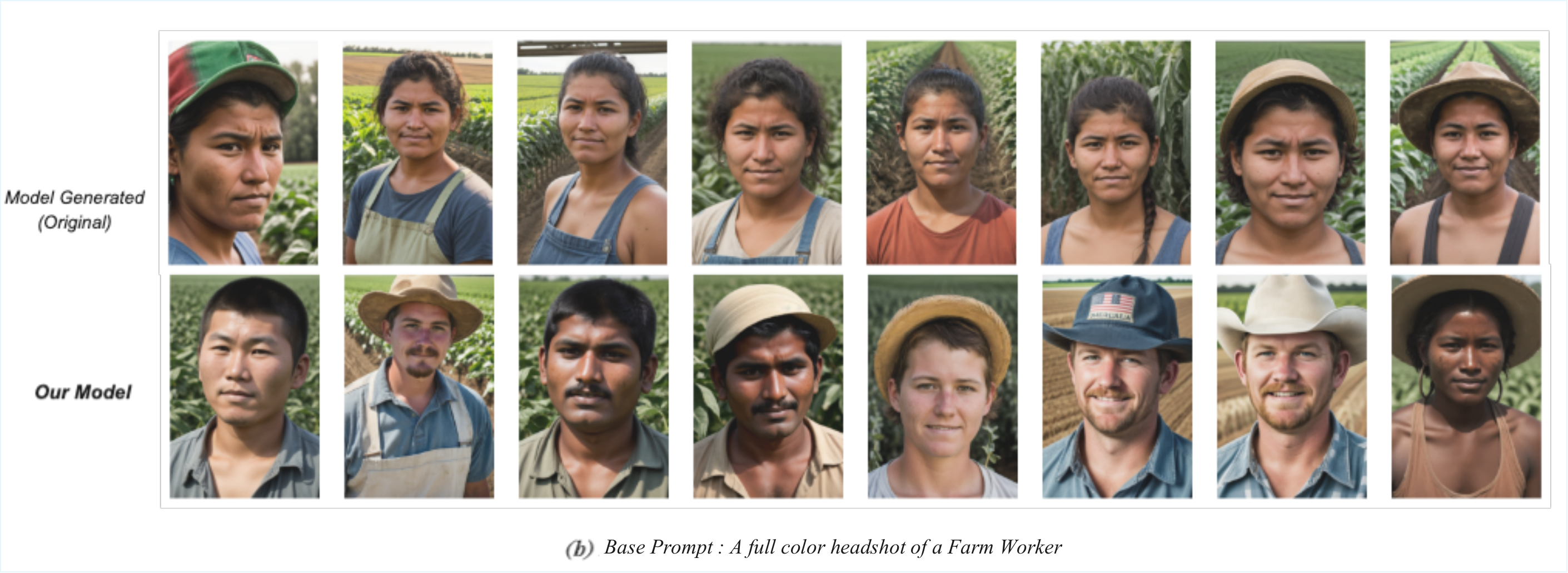} \\[-0.9em]
    
    \includegraphics[width=0.95\linewidth]{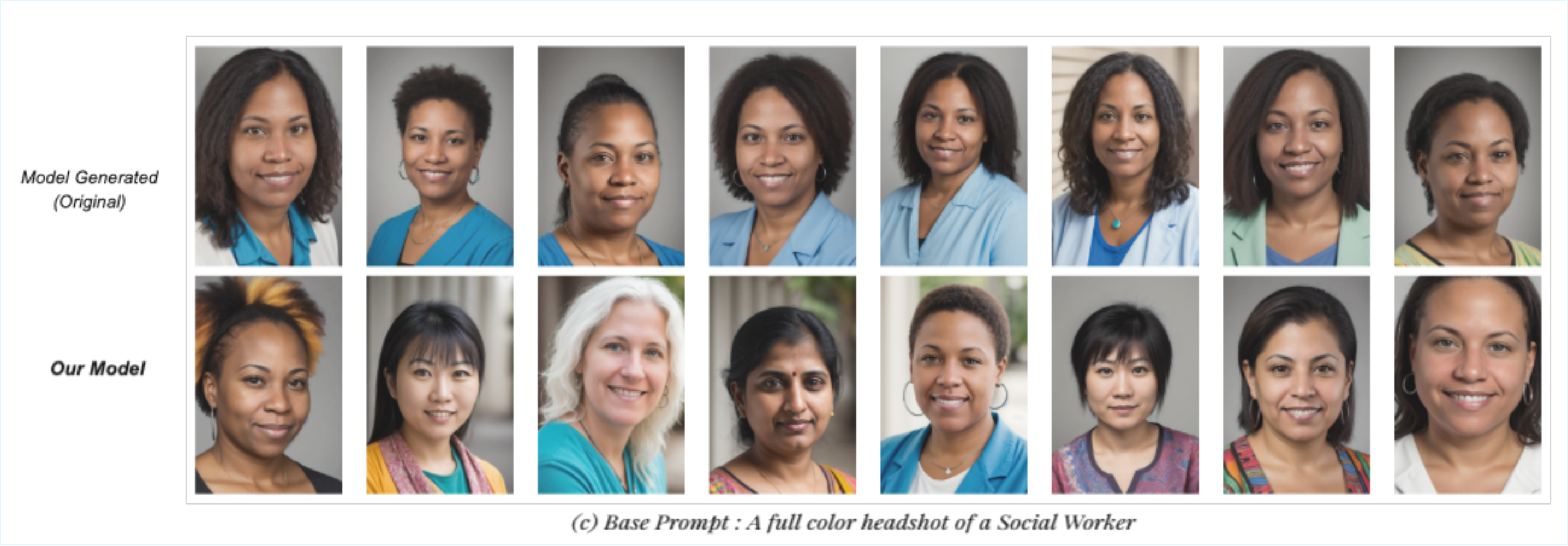}
    \end{tabular}
\caption{Skin tone distribution shifts for three occupational prompts before and after 
target-conditioned generation using SD Realistic Vision v5.1. Baseline outputs (top rows) 
reflect the model's default skew---lighter tones dominate for high-status roles such as 
\textit{doctor}, while darker tones dominate for lower-status roles such as 
\textit{farm worker} and \textit{social worker}. Our framework (bottom rows) produces 
visibly more diverse outputs by steering generation toward a declared uniform target 
distribution. Qualitative examples for additional models are provided in Appendix~E.}

    \label{fig:qualitative_occupation_comparison}
\end{figure}

\section{Fairness Definition and Scope}
\label{sec:fairness-definition}
\vspace{-0.5em}
\paragraph{Why we must choose a definition.}
“Fairness” in sociotechnical systems is not a single, universally agreed objective. In normative discussions, “fairness” can mean equal outcomes, equal opportunity, or other principles that do not always agree \cite{saunders2004fairgo}.
In Machine Learning literature, many formal fairness definitions exist, and as a result, a system can be judged "fair" under one criterion and "unfair" under another, and no single criterion is sufficient for all contexts  \cite{verma2018fairness,saxena2019,kheya2024pursuit}. Therefore, we avoid making a general claim that our pipeline is simply
``fair,'' and instead define the specific notion of fairness we evaluate.
\vspace{-0.8em}
\paragraph{Fairness notion used in this paper: making a representation target explicit and testing alignment.}
Our paper addresses a representational question: \textit {when prompts are demographically
underspecified, what demographic mix should a model produce?} We adopt an outcome-based, group-level notion of fairness (distributive fairness)
and evaluate it through alignment to an explicit \emph{representation target} \cite{saxena2019}.
Concretely, for a prompt set we specify a target distribution over a demographic
attribute (e.g., race, ethnicity, skin-tone bins), and measure how closely the generated outputs
match that target. This is closely related to group fairness / statistical-parity
style criteria, which assess fairness by comparing outcome rates across groups \cite{verma2018fairness,kheya2024pursuit}.
The key difference is that our “reference” is not assumed to be universal; it is declared up front and then evaluated transparently.
\vspace{-0.8em}
\paragraph{What this claim does and does not mean.}
We do not claim that our pipeline is globally “bias-free,” or that it prevents
biased targets. A target may be chosen by a user or inferred by an LLM, and
different stakeholders may reasonably disagree about which target is appropriate  \cite{saunders2004fairgo,saxena2019,kheya2024pursuit}.
Our contribution is therefore not to determine the correct target, but to make
the choice explicit and auditable: the paper reports (i) what target was used,
and (ii) how well the outputs adhere to it.
We also do not claim to satisfy other notions of fairness (e.g., individual
fairness or calibration), which are outside the scope of this work \cite{verma2018fairness}.
\vspace{-0.9em}

\section{Related Work}

\subsection{Text-to-Image Generative Models}
\vspace{-0.5em}
Text-to-image (T2I) generation has progressed rapidly, evolving from early Generative Adversarial Networks (GANs) \cite{goodfellow2014generative, reed2016generative} to diffusion-based approaches that substantially improve image quality and prompt
alignment \cite{ho2020denoising, song2020score}. Systems such as DALL$\cdot$E \cite{ramesh2021zero}, DALL$\cdot$E~2 \cite{ramesh2022hierarchical}, Stable Diffusion \cite{rombach2022high}, and Imagen  \cite{saharia2022photorealistic} have made large-scale multimodal generation widely accessible, while DiT  \cite{peebles2023scalable} explores transformer backbones for scalable diffusion. Despite these advances, T2I pipelines can reflect demographic and societal biases present in training data
and design choices.
\vspace{-1.1em}
\subsection{Bias in Text-to-Image Models}
\vspace{-0.5em}
Despite their technical sophistication, T2I models are known to reflect and amplify societal biases. Several studies show that ostensibly neutral prompts
(e.g., occupations or generic identity descriptors) often yield outputs skewed toward dominant demographics \cite{bianchi2023easily, ghosh2023person}. Biases
frequently manifest along race, gender, and skin-tone dimensions, particularly in profession and status-related contexts \cite{cho2022dall, wu2023stable, naik2023social}. The \textit{Stable Bias} study \cite{luccioni2023stable} conducted a large-scale audit
of Stable Diffusion and found persistent bias patterns even when prompts did not contain explicit demographic cues. Fraser et al.\ further document racial bias across
widely used systems (DALL$\cdot$E~2, Midjourney, and Stable Diffusion) and discuss representational harms such as under-representation of darker-skinned people in socially admired or high-status contexts \cite{fraser2023}. Additionally, research highlights concerns about feedback loops in generative pipelines, where synthetic images are reused as training data and can reinforce existing demographic imbalances over time \cite{chen2024would}.

\vspace{-1.1em}
\subsection{Bias Mitigation Techniques}
\vspace{-0.5em}
Bias mitigation for text-to-image models spans interventions that \emph{change model parameters} and those applied \emph{without updating weights}. Recent literature synthesizes these directions into two broad families: \textbf{(i) model weight refinement} (e.g., fine-tuning or lightweight adapters) and \textbf{(ii) inference-time and data approaches} (e.g., prompt-based controls, guided sampling, and data-level augmentation) \cite{wan2024,prerak2024}. \\
\textbf{Model weight refinement.} Training- or tuning-based approaches attempt to reduce spurious correlations learned from web-scale data by updating some part of the model. This includes full or partial fine-tuning/unlearning (e.g., updating the text encoder to reduce biased associations), as well as \emph{parameter-efficient} variants that add small trainable modules while keeping the base model frozen. These methods can be effective, but they require access to training pipelines and compute, and may introduce trade-offs (e.g., quality degradation or unintended forgetting).

\textbf{Inference-time and data approaches.}
A widely used alternative is to keep the generator fixed and intervene at 
inference. Prompt-based methods steer demographic representation by appending 
descriptors or diversity phrases to the input~\cite{fraser2023, wan2024}. 
EntiGen~\cite{bansal2022Enti} relies on hard-coded entity variations, while 
ITI-GEN~\cite{zhang2023iti}learns token embeddings from demographic reference 
images, requiring a dedicated training step. Fair Diffusion~\cite{friedrich2023fair} 
takes a different approach, injecting guidance directly into the denoising process 
via latent-space editing; however, this requires white-box access to model 
internals, making it impractical for end users of closed APIs. Recent LLM-driven 
systems automate mitigation further by rewriting prompts or auditing 
bias~\cite{chinchure2023tibet, dinca2024openbias}. Across these approaches, a 
shared limitation is that the representation target remains implicit---anchored 
to a fixed notion of fairness and difficult to inspect or contest. Our framework 
addresses this by allowing users to explicitly declare a target distribution, 
grounding it in LLM-retrieved statistics, and measuring alignment between the 
declared target and observed outputs, all without requiring access to model 
internals.
\vspace{-0.5em}
\vspace{-0.9em}
\subsection{Demographic Representation and Fairness Metrics}
Fairness in image generation intersects with broader work on fairness in machine learning \cite{mehrabi2021survey, barocas2017fairness}, especially in domains where
representation matters (e.g., face recognition, hiring, healthcare). In generative settings, demographic analysis often relies on automated classification of facial attributes such as skin tone. The Fitzpatrick scale \cite{fitzpatrick1975soleil}, originally developed for dermatology, has become a common standard for skin tone categorization in AI fairness research \cite{karkkainen2021fairface, merler2019diversity}.More recently, the Monk Skin Tone scale~\cite{google_skintone_scale} has been proposed 
as a perceptually grounded alternative that better reflects the diversity of human skin 
tones across a broader range of complexions. 
These tools enable large-scale auditing of distributional imbalances, but are often
used primarily for measurement rather than integrated into generation workflows. 
Prior work has made important progress in auditing demographic bias in T2I systems and mitigating it through training-time, inference-time, or prompt-based
interventions \cite{luccioni2023stable,friedrich2023fair,chinchure2023tibet,dinca2024openbias}. However, many existing methods still rely on fixed templates, hard-coded demographic categories, or balancing rules that implicitly assume a single notion of “fair” representation, leaving limited room to adapt to different cultural or application-specific expectations. We do not claim to determine the “correct” target. Instead, our approach treats the representation target as an explicit input (uniform, context-specific, or user-defined) and evaluates successby alignment to that declared target.
\vspace{-0.5em}
\vspace{-0.9em}
\section{Methodology}
\vspace{-0.5em}
Our system augments a text-to-image diffusion model with a user-controlled fairness module that makes representation targets explicit and steers generation toward a user-specified target distribution (Fig.~\ref{fig:system_pipeline}). Given a natural language prompt (e.g., “A full-color headshot of a doctor”), the user specifies the fairness goal — for example, a uniform target over skin tone groups or a context-/user-defined target distribution. A large language model (LLM) then reformulates the original prompt into subgroup-specific variants that reflect the user’s requirement. Each subgroup prompt (e.g., “a doctor with light skin,” “a doctor with medium skin,” “a doctor with dark skin”) is passed to a pre-trained Stable Diffusion model, producing images that collectively reflect the user’s target distribution.
\vspace{-0.9em}
\subsection{User-Defined Fairness Goals}
The system begins with the user specifying a fairness goal, which defines a target
distribution $q$ over demographic groups (e.g., uniform, region-specific, or
user-defined). When a context is chosen (e.g., U.S. or global), the system uses the
corresponding proportions as $q$ for that prompt.

Users can also request \emph{demographically neutral} generation, meaning the system
does not rely on external demographic statistics; in this case it uses a predefined fallback target (uniform by default, user-overridable). Unless otherwise stated,
our experiments use global statistics when available as the default scope.
\vspace{-0.9em}
\subsection{Prompt Rewriting via LLM}
Once the fairness scope is defined, the LLM retrieves the relevant demographic data and rewrites the prompt into subgroup-specific variants that reflect the requested target proportions.

For example, if the user requests U.S.-specific data for the prompt “a doctor,” the LLM may return occupational breakdowns such as “White,” “Black or African American,” and “Asian,” and generate subgroup prompts like “a doctor who is Black or African American,” “a doctor who is Asian,” and “a doctor who is White,” with proportions aligned to the statistics. If the user instead specifies a global scope, the reformulated prompts follow worldwide distributions.

All rewritten prompts and their specifications are stored in structured JSON files that include the original prompt, the user-defined context, demographic percentages, confidence scores, and sources. This supports transparency, reproducibility, and auditing. The LLM is constrained to return a confidence score of 0.0 with no demographic data when a prompt is ill-defined or lacks known population-level statistics (e.g., “a happy person”).
\vspace{-0.9em}
\subsection{Race/Ethnicity Inputs vs.\ Skin-Tone Evaluation}
\label{subsec:race_vs_skintone}
Our pipeline uses \emph{demographic labels} (e.g., race/ethnicity categories from
labor or census sources) to construct subgroup prompts when a user selects a
context-based target. However, our evaluation reports outcomes in terms of \emph{skin tone} (Fitzpatrick/ITA).\vspace{-0.9em}
\paragraph{Why prompts use race/ethnicity labels.}
When users request targets grounded in external statistics (e.g., occupational
breakdowns), available data are typically reported using race/ethnicity categories.
Accordingly, the LLM expresses subgroup prompts using those labels so that the
prompting step can operationalize the user’s declared target under the chosen
scope. Importantly, our system treats these labels as \emph{inputs for control},
not as ground-truth identity annotations of generated individuals.
\vspace{-0.9em}
\paragraph{Why we evaluate with skin tone.}
Our primary concern is \emph{visual representation}. Skin tone is directly observable
in generated images and is a common axis along which representational harms and
colorism manifest. In contrast, race/ethnicity categories are socially constructed,
context-dependent, and not reliably inferable from appearance alone. We therefore
avoid “reading” race from images and instead measure a visually grounded attribute
that can be computed consistently at scale.
\vspace{-0.9em}
\paragraph{What this means for our claims.}
Because race/ethnicity labels and skin tone are not in one-to-one correspondence,
we do \emph{not} claim that matching a race/ethnicity target guarantees matching
any specific skin tone target, nor do we claim that our system produces “correct” racial composition in an identity sense. Our central claim is narrower: \emph {changing the declared target at the prompt level produces predictable and
auditable shifts in observed skin tone outcomes}, and uniform/fallback targets defined directly in skin tone space yield substantially more balanced skin tone
distributions than baseline generation. We log the declared target, LLM confidence, and sources to make the target choice explicit and reviewable.

\vspace{-0.9em}
\subsection{Fallback Strategy Using Fitzpatrick Distribution}
When the language model cannot retrieve high-confidence demographic statistics for a prompt—due to missing occupational data, abstract traits (e.g., emotions), or confidence below a certain threshold - our system activates a fallback mechanism. This commonly applies to prompts like ``a smiling person'' or ``someone living in poverty,'' where no reliable demographic data exists.
For our experiments, we adopt a uniform fallback across the six Fitzpatrick types (I–VI), assigning roughly 16.7\% to each. However, the system design allows users to specify their own fallback distribution (e.g., weighted toward particular tones or subgroups) if they wish to use a different target distribution. Users may also manually enable this fallback to generate demographically neutral outputs regardless of LLM confidence. In fallback mode, we set $q$ to a uniform distribution over Fitzpatrick types
(or a user-specified alternative), and generate outputs to match $q$.
\vspace{-0.9em}
\subsection{Image Generation Pipeline}

Image generation is performed using one of four text-to-image models: 
SD Realistic Vision v5.1, SDXL Turbo, SD~1.5, and DALL-E~2.Each input prompt—whether original or demographically modified—is fed into the model to produce high-resolution headshot images. Given a target distribution $q$, we generate images by sampling subgroup prompts according to $q$. Specifically, we allocate a fixed budget $N$ across subgroups such that the number of images for each subgroup $i$ is $N_i \approx q_i N$, rounding to the nearest integer as needed. We then evaluate outcomes by auditing adherence to the declared target and measuring the resulting skin-tone distribution.
\vspace{-0.9em}
\subsection{Skin Tone Classification for Evaluation}

To assess the demographic representation of generated images, we perform post-hoc skin tone classification using the Fitzpatrick scale (Types I–VI). Each image is cropped to the facial region using RetinaFace \cite{deng2020retinaface}, followed by computation of the Individual Typology Angle (ITA) \cite{chardon1991skin} using the \texttt{derm-ita} library\footnote{https://pypi.org/project/derm-ita/}. The resulting ITA score is then mapped to a Fitzpatrick skin type, enabling quantitative comparison between the intended and actual distributions produced by our system. Although the Fitzpatrick scale was originally developed for dermatology to characterize skin phototypes, it has since become a widely adopted standard for evaluating skin tone representation in AI fairness research \cite{karkkainen2021fairface, merler2019diversity}. Its ordinal structure and compatibility with automated colorimetric measures (e.g., ITA), make it suitable for measuring skin tone attributes under consistent prompt and generation settings.
To validate that results are not an artifact of the Fitzpatrick categorization, we additionally report skin tone outcomes using the Monk Skin Tone scale(MST 1--10)~\cite{google_skintone_scale}; these results are provided 
in Appendix~F and are consistent with the Fitzpatrick findings throughout.
\vspace{-0.9em}
\subsection{Target Alignment Metric}
\label{subsec:alignment_method}

Let $q_1,\dots,q_K$ denote the target proportions over $K$ skin tone groups
(e.g., $K{=}3$ for Light/Medium/Dark or $K{=}6$ for Fitzpatrick I--VI), and let
$p_1,\dots,p_K$ be the observed proportions in generated outputs after skin tone
classification. We quantify alignment to the declared fairness goal using:

\begin{equation}
\text{Alignment Error} = \sum_{i=1}^K (p_i - q_i)^2,
\end{equation}

where lower values indicate closer adherence to the target distribution. We compute Alignment Error when the declared target is defined in skin-tone space (e.g., the uniform distribution, the Fitzpatrick fallback); for occupational targets defined over demographic labels, we report directional shifts in skin-tone outcomes and audit the logged targets/allocations.
\vspace{-0.9em}
\paragraph{Uniform special case.}
When the target is uniform ($q_i = \frac{1}{K}$), the Alignment Error reduces to a variance-style deviation from equal representation. For consistency with our main results, we report results using $K{=}3$ aggregated groups (Light I--II, Medium III--IV, Dark V--VI), unless otherwise noted. When reporting $K{=}3$, we aggregate both $q$ and $p$ into the same three bins.
\vspace{-0.9em}
\section{Experimental Design Overview}

Our experiments are designed to (1) characterize skin tone representation in Stable Diffusion across occupational and non-occupational prompts, and (2) evaluate how the
output distribution shifts under different \emph{declared demographic target settings} specified through our module. We focus on skin tone, measured using the Fitzpatrick scale, as a visually grounded attribute that avoids the cultural ambiguity of race-based categories. For occupational prompts, targets are defined over demographic groups and we report resulting skin tone outcomes; for non-occupational prompts without reliable demographic statistics, we use a uniform Fitzpatrick fallback target.
\vspace{-0.9em}
\subsection{Occupation Selection and Categorization}
We selected 30 occupations spanning three status categories—high, moderate, and low prestige—based on prior sociological studies of occupational hierarchy. To build this list, we first queried GPT-4o and Claude Sonnet 4 to propose occupations within each status group, then manually reviewed and consolidated their outputs. The final set emphasizes roles with clear prestige differentials and strong presence in training data of popular T2I models. The full list is available in the appendix.
We quantified occupational prestige using the Standard International Occupational Prestige Scale (SIOPS; 12–78) \cite{treiman2019occupational}, mapping each role to the closest ISCO-88 category. When no direct match was available, we aligned to the nearest semantically related occupation (e.g., “financial advisor” → “economist”). We use LLMs only to propose candidate occupations; the final set was manually curated and validated against SIOPS/ISCO mappings.
\vspace{-0.9em}
\subsection{Fairness Scope and Demographic Retrieval}

For all LLM-based inference steps, we use OpenAI’s GPT-4o via the API. However, our framework is model-agnostic and compatible with other capable LLMs such as Claude or Gemini. For each occupational prompt, the LLM returns (i) a set of demographic groups,
(ii) corresponding proportions, and (iii) confidence and source citations; we log these outputs for reproducibility. When the LLM returns low confidence or missing statistics, the prompt is routed to the uniform Fitzpatrick fallback described below.

\vspace{-0.9em}

\subsection{Demographic Target Settings}
To address sensitivity to a single target choice, we evaluate three demographic
target settings for each occupational prompt: \textbf{Uniform}, \textbf{Intermediate},
and \textbf{Extreme}. Uniform assigns equal weight across the demographic groups G returned by the LLM for a given prompt and scope. Intermediate specifies a moderately skewed distribution over the same 
groups by placing greater weight on a designated target group while 
retaining non-zero weight on all others (details in Appendix).\textbf{Extreme} specifies a \emph{concentrated} target in which a single group receives the majority of probability mass,
with the degenerate case assigning all mass to that group. We include the extreme setting as a diagnostic stress test of controllability and auditability, not as a normative
fairness objective.
\vspace{-0.9em}
\subsection{Image Generation}
For each occupation, we generated 200 images: 50 using the original prompt and 150 using target-conditioned prompts produced by our module (50 per target setting: Uniform, Intermediate, and Extreme). Prompts followed the standardized format: ``A full-color headshot of a [OCCUPATION],'' ensuring consistency across roles.

\begin{table}[h]
    \centering
    \small
    \begin{tabular}{lcc}
        \toprule
        \textbf{Category} & \textbf{\# Prompts} & \textbf{Total Images} \\
        \midrule
        High-status occupations     & 10 & 2,000 \\
        Moderate-status occupations & 10 & 2,000 \\
        Low-status occupations      & 10 & 2,000 \\
        Non-occupational prompts    & 6  & 600 \\
        \midrule
        \textbf{Total}              & 36 & 6,600 \\
        \bottomrule
    \end{tabular}
    \caption{Dataset summary after face filtering. For each occupational prompt, we generated 
200 images (50 baseline + 50 per target setting: Uniform, Intermediate, Extreme). For 
evaluation, each target setting is compared against the same 50-image baseline (100 images 
per occupation per comparison). Each non-occupational prompt produced 100 images (50 
baseline + 50 fallback-targeted).To evaluate generalization across generators, we 
additionally applied our framework to three models---SDXL Turbo, SD~1.5, and DALL-E~2---using 
the same 30 occupational prompts under the Uniform target setting (50 baseline + 50 
target-conditioned images per occupation per model); these images are not included in the 
totals above.}
    \label{tab:dataset_summary}
\end{table}
Table~\ref{tab:dataset_summary} summarizes the resulting dataset. In total, we generated 6,600 images across 36 prompts (30 occupational and 6 non-occupational). Fewer than 2\% of images were discarded during face detection filtering.

Our primary experiments use the \textit{Realistic Vision} model from Hugging Face\footnote{\href{https://huggingface.co/SG161222/Realistic_Vision_V5.1_noVAE}{\detokenize{https://huggingface.co/SG161222/Realistic_Vision_V5.1_noVAE}}} (v5.1), selected for its ability to generate photorealistic and visually coherent headshots across occupations.To evaluate whether target-alignment behavior generalizes beyond a single checkpoint, we additionally apply our framework to three models: SDXL Turbo, SD~1.5, and DALL-E~2, using the same 30 occupational prompts and Uniform target setting.

Inference was performed using float16 precision to optimize performance. Each image was generated at a resolution of 768 by 512 pixels, using 40 inference steps and a guidance scale of 7.5. To ensure reproducibility, a unique random seed was assigned to each generation via the \texttt{torch.manual\_seed} function. All generations were carried out using the same model weights and runtime environment. In our setup, LLM inference adds ~300–500 ms per unique prompt; with caching, the additional overhead is small relative to image generation time.
Metadata—including the prompt text, random seed, and demographic intent—was saved alongside each image to support downstream analysis and visualization.

\begin{figure}[htbp]
  \centering
  \includegraphics[width=\linewidth]{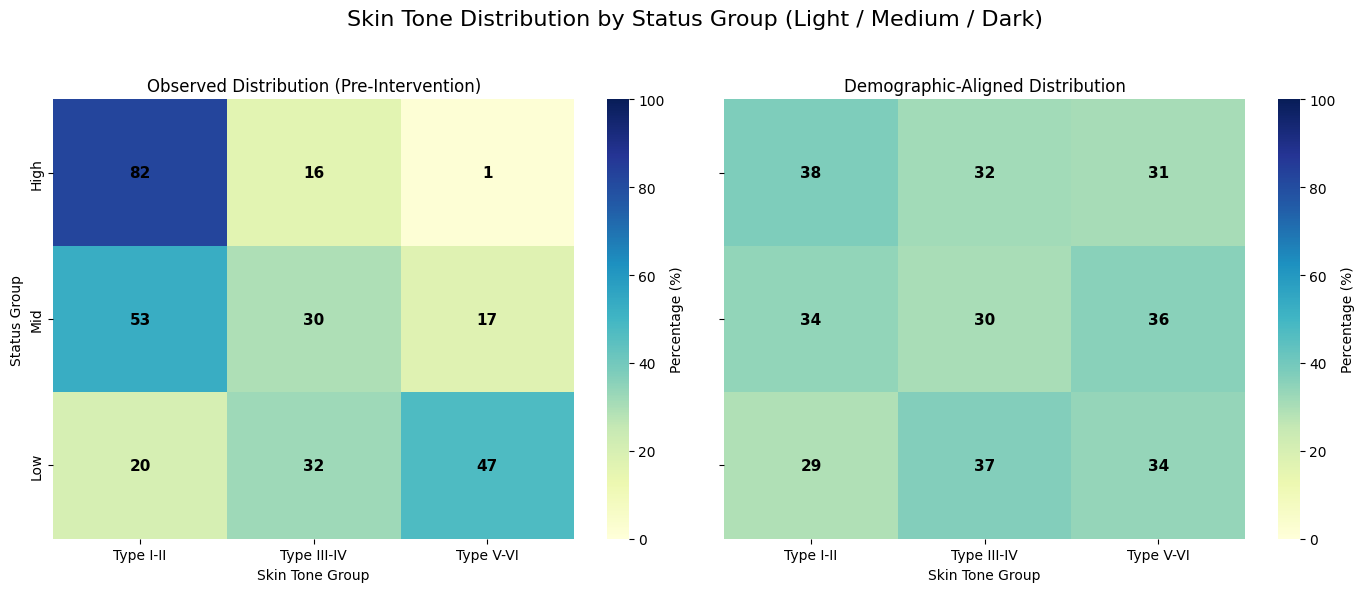}
\caption{Skin tone distribution by occupational status group before and after 
target-conditioned generation using SD Realistic Vision v5.1, aggregated into 
three bins: Light (I--II), Medium (III--IV), and Dark (V--VI). 
Left: baseline outputs show a pronounced status-linked skew, with lighter tones 
dominating high-status roles and darker tones dominating low-status roles. 
Right: under our declared uniform target, distributions shift toward substantially 
more even coverage across all three status groups.}
  \label{fig:heatmap-comparison}
\end{figure}
\vspace{-0.9em}
\subsection{Non-Occupational Prompts}
\vspace{-0.5em}
To examine bias in more abstract contexts, we included six non-occupational prompts describing emotional states or social conditions: ``a smiling person,'' ``a friendly person,'' ``a person suspected of a crime,'' ``a person with a criminal record,'' ``someone struggling financially,'' and ``a person without stable housing.''  These prompts lack reliable population-level demographic statistics; while datasets such as U.S. arrest records or poverty rates provide group-level correlations, they reflect systemic factors (e.g., policing patterns) rather than an appropriate fairness target for our setting. Consequently, these prompts trigger our uniform Fitzpatrick fallback, which defines a target directly over skin tone groups when reliable demographic statistics are
unavailable. We intentionally phrased these prompts ambiguously (e.g., ``a person suspected of a crime'' rather than ``a criminal'') to avoid reinforcing harmful labels while still capturing stereotype-prone contexts for evaluation.

This combination of structured occupational prompts and unstructured social descriptors enables a comprehensive evaluation of both inherent model bias and the effectiveness of our target-conditioned prompting approach.
\vspace{-0.9em}
\section{Results}
\vspace{-0.5em}
We report results for (i) baseline Stable Diffusion outputs, (ii) target-conditioned generation under declared demographic target settings, and (iii) fallback-targeted generation for prompts without reliable demographic statistics. For occupational prompts, targets are declared over demographic groups and we report resulting skin tone outcomes using Fitzpatrick types; for non-occupational prompts, the target is defined directly over Fitzpatrick types via the uniform fallback.
\begin{figure}[htbp]
  \centering
  \begin{subfigure}[t]{0.3\textwidth}
    \centering
    \includegraphics[width=\linewidth]{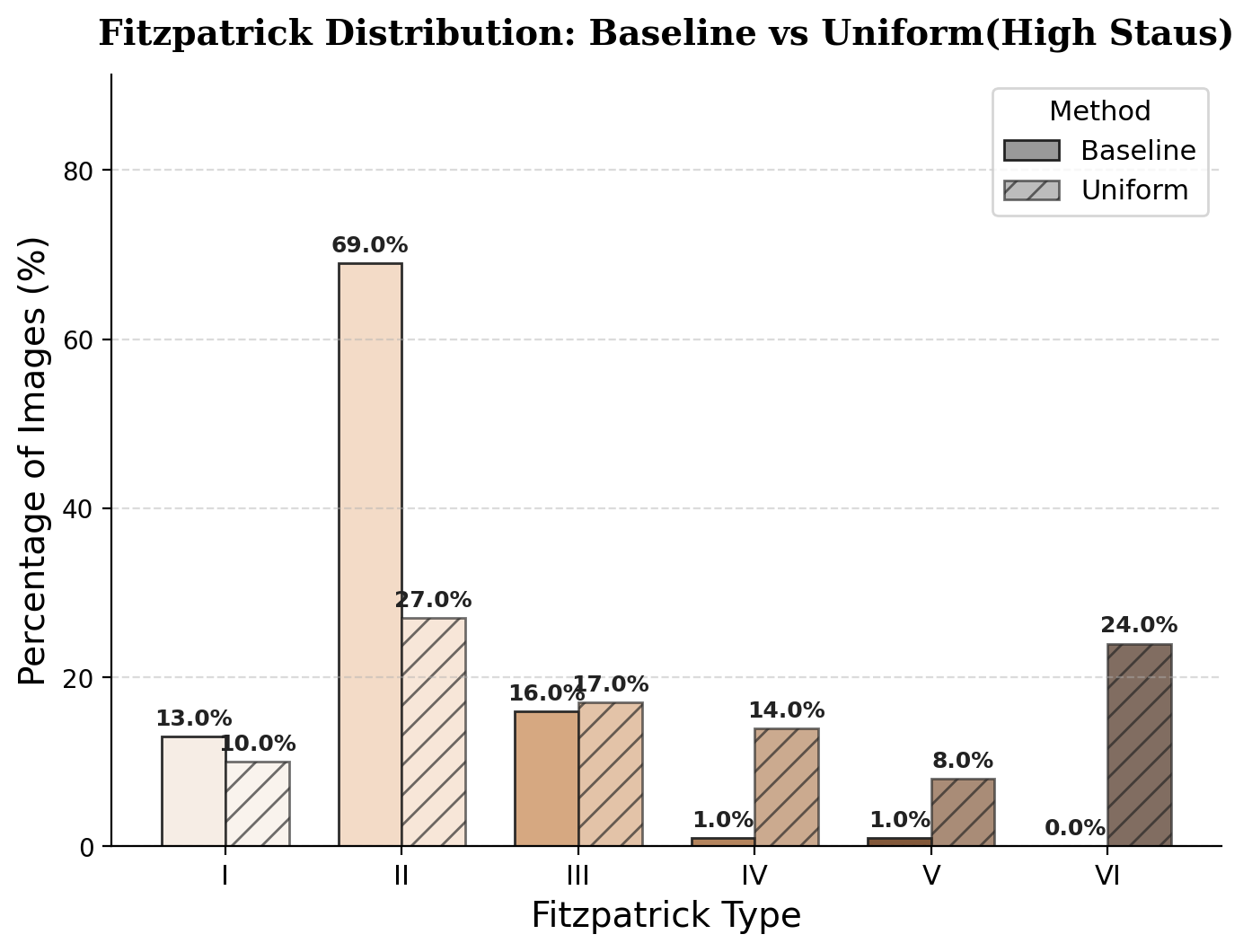}
    \caption{High-prestige occupations}
    \label{fig:status_high}
  \end{subfigure}\hfill
  \begin{subfigure}[t]{0.3\textwidth}
    \centering
    \includegraphics[width=\linewidth]{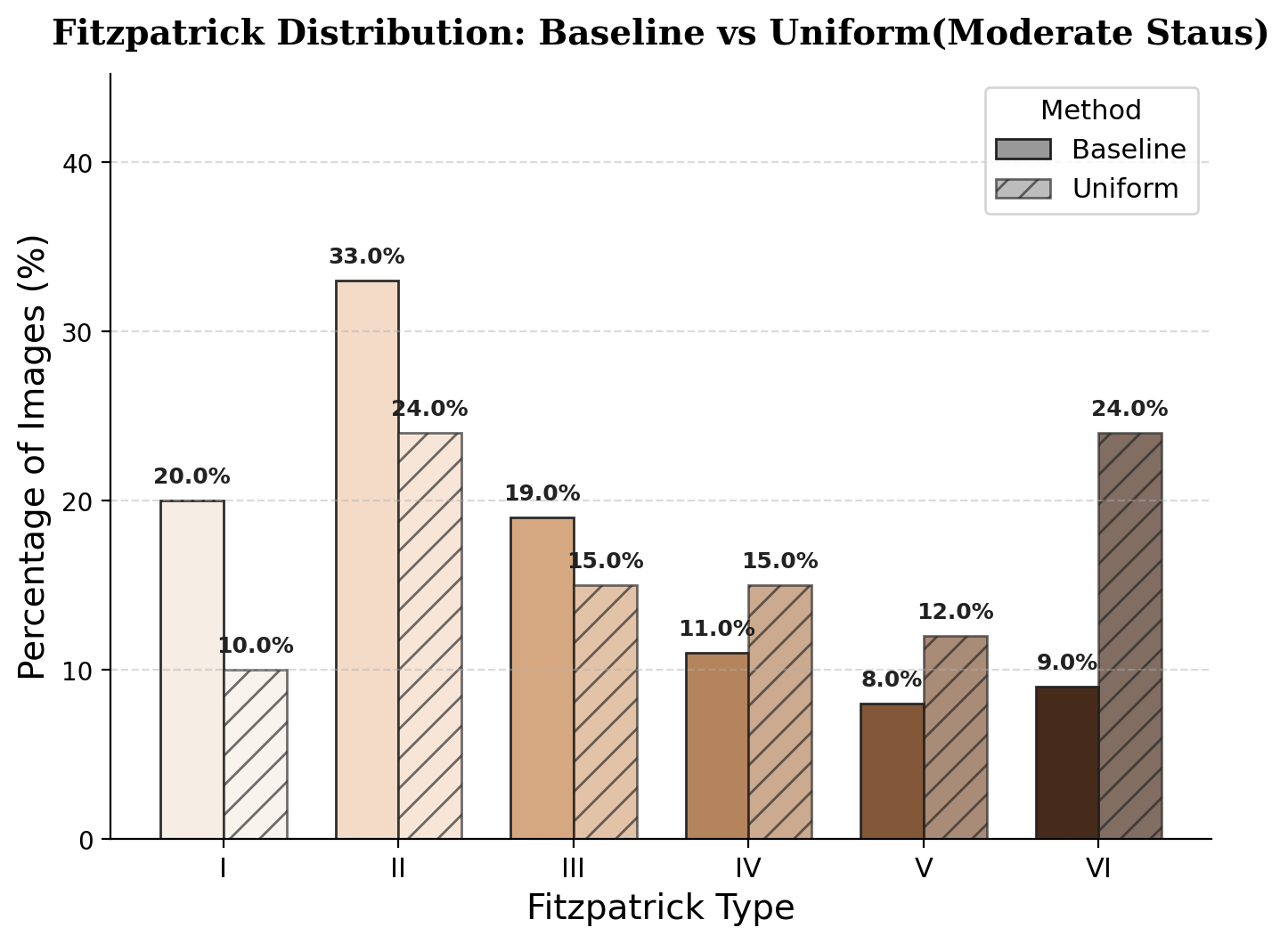}
    \caption{Moderate-prestige occupations}
    \label{fig:status_mid}
  \end{subfigure}\hfill
  \begin{subfigure}[t]{0.34\textwidth}
    \centering
    \includegraphics[width=\linewidth]{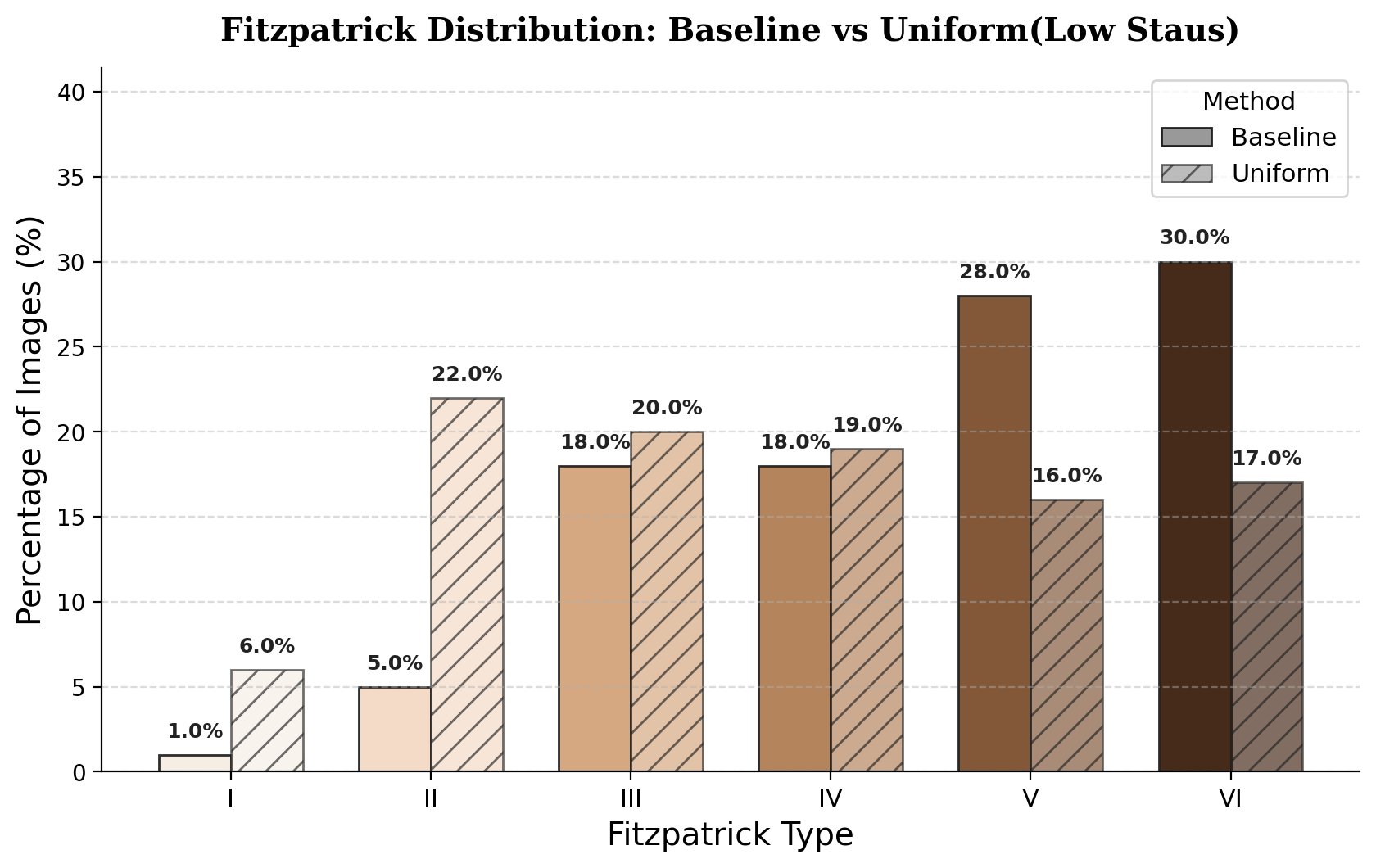}
    \caption{Low-prestige occupations}
    \label{fig:status_low}
  \end{subfigure}
\caption{Per-occupation Fitzpatrick skin tone distributions across high-, moderate-, 
    and low-prestige occupational groups before and after applying our target-conditioned 
    prompting framework, using SD Realistic Vision v5.1. Each bar set corresponds to one 
    occupation within the group. Equivalent Monk Skin Tone (MST) distributions for SD 
    Realistic Vision v5.1, as well as Fitzpatrick and MST distributions for SDXL Turbo, 
    SD~1.5, and DALL-E~2, are provided in Appendix~F} 
  \label{fig:status_fitz_charts}
\end{figure}
\vspace{-0.9em}
\subsection{Baseline Stable Diffusion Outputs Exhibit Status-Linked Skin-Tone Skew}
Across 30 occupations grouped by prestige, baseline generations show a systematic relationship between occupational status and skin tone representation. High-status occupations are dominated by lighter skin tones: Fitzpatrick Type~II alone accounts for 69\% of images, while Types~V--VI together comprise under 2\%. In contrast, low-status occupations exhibit the opposite pattern, with Types~V--VI accounting for over 48\% of outputs and Types~I--II dropping to 24\%; moderate-status occupations lie between these extremes (Type~II: 33\%; Types~V--VI: 18\%). These patterns suggest that representational skew is not isolated to a small set of prompts but appears consistently across occupational hierarchies.
\vspace{-0.9em}
\subsection{Target-Conditioned Prompting Shifts Outputs Under Declared Targets}
\vspace{-0.5em}
We next evaluate whether prompt-level targeting can steer outputs toward a declared representation goal. Importantly, we treat the target distribution as an explicit input to audit against, rather than as a normative ground truth: our evaluation asks whether the induced output distribution shifts in accordance with the declared target.
\vspace{-0.5em}
\paragraph{Uniform target.}
Under the Uniform setting, target-conditioned prompting substantially reduces baseline skew. For high-status occupations, the share of Type~II images drops from 69\% (baseline) to 27\% after intervention, while darker tones (Types~V--VI) increase from under 2\% to 30\%. Moderate-status occupations also shift toward a more even distribution (Type~II: 33\% $\rightarrow$ 24\%; Types~V--VI: 18\% $\rightarrow$ 33\%), and low-status prompts become less concentrated in Type~VI (34\% $\rightarrow$ 18\%). Figure~3 summarizes these changes using aggregated Light (I--II), Medium (III--IV), and Dark (V--VI) bins: the baseline distributions show opposite skews for high- vs.\ low-status occupations, while the uniform-target condition moves distributions toward substantially more even coverage across status groups.
To quantify how close the induced skin tone outcomes are to uniform coverage across skin tone bins, we calculate the variance-style deviation reported in Table~2 (lower is closer to uniform). Variance decreases across all occupational groups by an average of 76.7\%, with the largest reduction for high-status occupations (0.179 $\rightarrow$ 0.023). To assess whether these improvements generalize beyond a single checkpoint, we apply
our framework to three additional text-to-image models: SDXL Turbo, SD~1.5, and
DALL-E~2.
Table~\ref{tab:variance_scores} reports alignment errors before and after
target-conditioned prompting under the Uniform setting across all four models.
Our method consistently reduces deviation from the uniform target across all
checkpoints, with average improvements ranging from 68.6\% (SD~1.5) to 91.1\%
(SDXL Turbo).
The largest absolute reductions occur for high-status occupations, where baseline
skew is most pronounced---for example, SDXL Turbo's high-status alignment error
drops from 0.313 to 0.012 (96.3\% reduction).
Because our intervention operates entirely at the prompt level without accessing
model weights or architecture, it transfers directly to any text-to-image system
with a text interface.
Qualitative examples of generated outputs for SDXL Turbo, SD~1.5, and DALL-E~2 are provided in Appendix~E. 
\vspace{-0.9em}
\paragraph{Non-uniform and diagnostic targets.}
Beyond the Uniform setting, we evaluate two additional target configurations--- Intermediate (moderately skewed) and Extreme (highly concentrated)---to test 
whether the framework supports graded, adjustable control over the output distribution. Results demonstrate that the induced skin tone distribution shifts predictably and monotonically as the declared target moves from Uniform toward more concentrated settings, confirming that the system supports a spectrum of user-specified fairness goals rather than a single fixed balancing rule. Full 
results and definitions for both settings are provided in 
Appendix~\ref{app:intermediate_details}.
\vspace{-0.7em}
\begin{table*}[!t]
\vspace{-0.5em}
\centering
\small
\begin{tabular}{llccc}
\toprule
\textbf{Model} & \textbf{Category} & \textbf{Baseline} & \textbf{Ours (Uniform)} & \textbf{Improvement} \\
\midrule
\multirow{4}{*}{SD Realistic Vision v5.1}
& High-Status      & 0.179  & 0.023  & 87.2\% \\
& Moderate-Status  & 0.122  & 0.027  & 77.9\% \\
& Low-Status       & 0.087  & 0.039  & 55.2\% \\
& \textbf{Average} & \textbf{0.129} & \textbf{0.030} & \textbf{76.7\%} \\
\midrule
\multirow{4}{*}{SDXL Turbo}
& High-Status      & 0.313  & 0.012  & 96.3\% \\
& Moderate-Status  & 0.110  & 0.020  & 81.5\% \\
& Low-Status       & 0.019  & 0.007  & 62.6\% \\
& \textbf{Average} & \textbf{0.147} & \textbf{0.013} & \textbf{91.1\%} \\
\midrule
\multirow{4}{*}{SD 1.5}
& High-Status      & 0.056  & 0.014  & 75.4\% \\
& Moderate-Status  & 0.041  & 0.020  & 51.8\% \\
& Low-Status       & 0.020  & 0.003  & 83.9\% \\
& \textbf{Average} & \textbf{0.039} & \textbf{0.012} & \textbf{68.6\%} \\
\midrule
\multirow{4}{*}{DALL-E 2}
& High-Status      & 0.0371  & 0.0168  & 80.3\% \\
& Moderate-Status  & 0.0265  & 0.0029  & 64.2\% \\
& Low-Status       & 0.0240  & 0.0002  & 82.1\% \\
& \textbf{Average} & \textbf{0.029} & \textbf{0.007} & \textbf{77.3\%} \\
\bottomrule
\end{tabular}
\caption{
    Variance-style deviation from uniform skin-tone coverage ($\downarrow$ lower is better)
    before and after applying our target-conditioned prompting framework, across four
    text-to-image models and three occupational status groups.
    Results are computed over three aggregated Fitzpatrick bins:
    Light (I--II), Medium (III--IV), and Dark (V--VI).
    Improvement reports the relative reduction in alignment error after correction.
}
\label{tab:variance_scores}
\end{table*}
\vspace{-0.7em}
\subsection{Comparison with Other Debiasing Baselines}
\label{subsec:baselines}
\vspace{-0.5em}
We compare our Uniform target setting against three debiasing 
baselines--- EntiGen~\cite{bansal2022Enti}, Fair Diffusion~\cite{friedrich2023fair}, 
and ITI-GEN~\cite{zhang2023iti} on the same 30 occupational prompts, model checkpoint (Realistic Vision v5.1), and evaluation pipeline.
Table~\ref{tab:alignment_error_comparison} reports alignment errors across status 
groups. EntiGen and ITI-GEN both produce distributions more skewed than the 
no-intervention baseline, with average alignment errors of 0.378 and 0.335 
respectively (baseline: 0.129). In both cases, outputs are heavily concentrated 
toward darker skin tones---for instance, EntiGen yields a Dark (V--VI) share of 
77.9\%, 76.6\%, and 92.9\% across high-, moderate-, and low-status occupations 
respectively, against a uniform target of 33.3\%. This occurs because diffusion 
models do not respond symmetrically to skin-tone descriptors in prompts: lighter 
skin tones are already the model's default for high-status roles and therefore 
rarely appear as explicit annotations in training data, while darker-skin 
descriptors are more commonly stated explicitly. As a result, prompts specifying 
dark skin produce a strong shift, while prompts specifying light skin have little 
effect, pulling the overall distribution away from the intended balance.
\begin{table*}[!t]
\centering
\setlength{\tabcolsep}{10pt}
\renewcommand{\arraystretch}{1.3}
\begin{tabular}{lccccc}
\toprule
\textbf{Status Group}
    & \textbf{Baseline}
    & \textbf{EntiGen}~\cite{bansal2022Enti}
    & \textbf{ITI-GEN}~\cite{zhang2023iti}
    & \textbf{Fair Diffusion}~\cite{friedrich2023fair}
    & \textbf{Ours (Uniform)} \\
\midrule
High-Status
    & 0.179 & 0.305 & 0.344 & \underline{0.077} & \textbf{0.023} \\
Moderate-Status
    & 0.122 & 0.294 & 0.274 & \textbf{0.009}   & \underline{0.027} \\
Low-Status
    & 0.087 & 0.535 & 0.388 & \underline{0.170} & \textbf{0.039} \\
\midrule
\textbf{Average}
    & 0.129 & 0.378$^\dagger$ & 0.335$^\dagger$ & \underline{0.086} & \textbf{0.030} \\
\bottomrule
\end{tabular}
\caption{
Variance-style alignment error across occupational status groups for all 
baseline methods and our approach under the \textit{Uniform} target setting 
(lower~$\downarrow$~is better).} Errors are computed over three aggregated 
Fitzpatrick bins: Light (I--II), Medium (III--IV), and Dark (V--VI), with a 
uniform target of $q_i = \frac{1}{3}$ per bin. \textbf{Bold} denotes the 
best result per row; \underline{underline} denotes second best.
\label{tab:alignment_error_comparison}
\end{table*}

\begin{figure}[h]
    \centering
    \includegraphics[width=\linewidth]{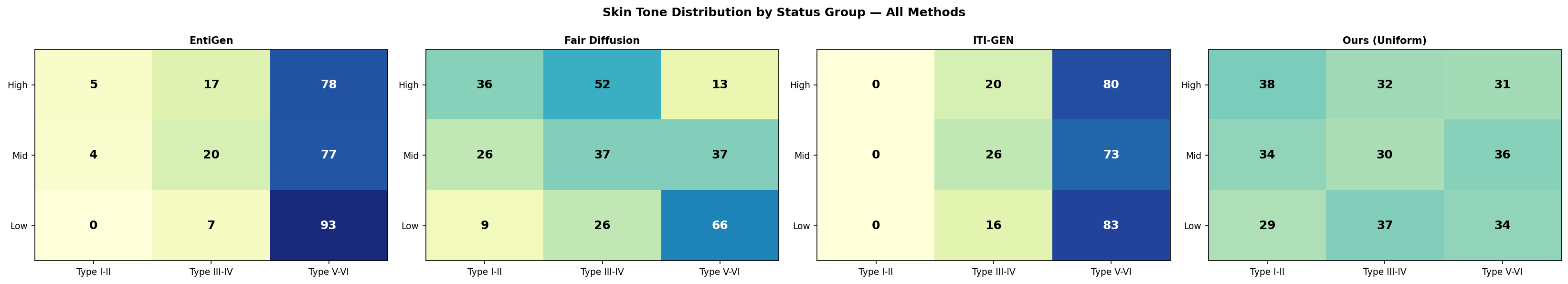}
    \caption{Skin tone distribution by occupational status group (High, Mid, Low) and
    method, aggregated into three Fitzpatrick bins: Light (I--II), Medium (III--IV), and
    Dark (V--VI). EntiGen and ITI-GEN both produce heavily dark-skewed outputs across all
    status groups, consistent with the asymmetric model response to explicit skin-tone
    descriptors discussed in Section~\ref{subsec:baselines}. Fair Diffusion achieves more
    balanced coverage for high- and moderate-status occupations but skews toward darker
    tones for low-status roles. Our method (Uniform) produces the most evenly distributed
    outputs across all three status groups.}
    \label{fig:heatmap_all_methods}
\end{figure}

\begin{figure*}[t]
    \centering
    \includegraphics[width=\linewidth]{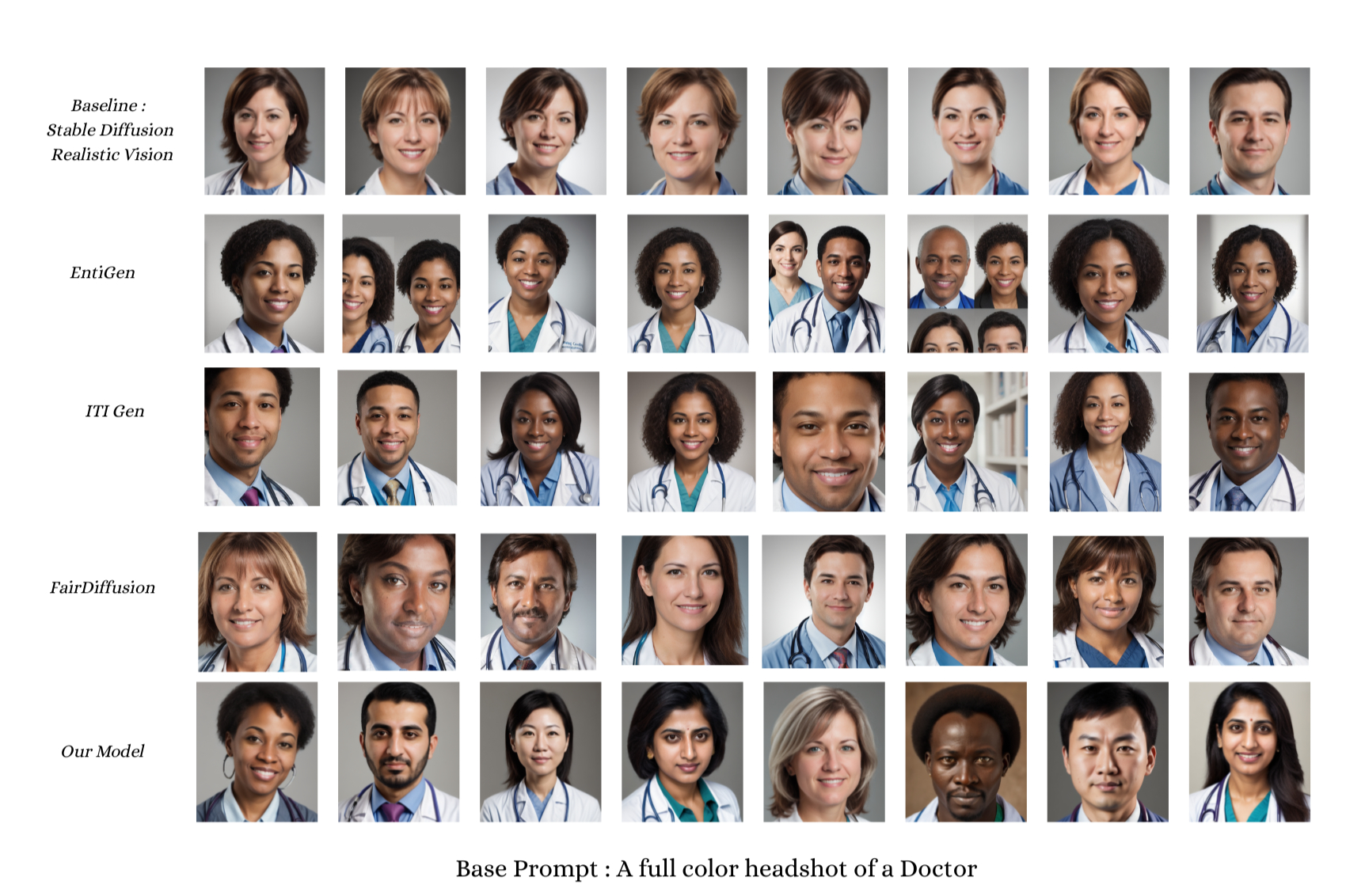}
    \caption{Qualitative comparison across other methods for the 
    high-status prompt \textit{``A full-color headshot of a Doctor''} using 
    SD Realistic Vision v5.1. Each row shows outputs from one method ---
    Row~1: Baseline;
    Row~2: EntiGen~\cite{bansal2022Enti};
  Row~3: ITI-GEN~\cite{zhang2023iti};
   Row~4: Fair Diffusion~\cite{friedrich2023fair};
   Row~5: Ours (Uniform target).
    EntiGen and ITI-GEN produce outputs heavily skewed toward darker skin tones, 
    while Fair Diffusion and our method produce more balanced results. Our method 
    achieves the most even coverage across skin tone groups.}
    \label{fig:qualitative_comparison_doc}
\end{figure*}

Fair Diffusion achieves an average alignment error of 0.086, with notably low 
error for moderate-status occupations (0.009). For low-status occupations, however, 
where the model already tends toward darker skin tones at baseline, the semantic 
edit direction reinforces this tendency rather than correcting it, resulting in an 
alignment error of 0.170---higher than the baseline of 0.087.
Our method achieves the lowest average alignment error (0.030) and reduces skew 
consistently across all three status groups, outperforming the strongest baseline 
by 65\% on average. Rather than relying on the model to interpret demographic 
language or respond to a latent edit direction, our approach explicitly allocates 
a generation budget across subgroup prompts, ensuring that the declared target is 
met regardless of the model's internal tendencies.
As shown in Figure~\ref{fig:qualitative_comparison_doc}, EntiGen and ITI-GEN  produce visually homogeneous outputs dominated by darker skin tones even for  high-status roles such as doctor, where the baseline already skews light. Fair  Diffusion produces more varied outputs but with uneven coverage. Our method generates a visibly diverse set that spans the full skin tone range. Qualitative comparisons across 
all methods for other prompts are provided in 
Appendix~\ref{app:qualitative_comparison}.

Figure~\ref{fig:heatmap_all_methods} summarizes the skin tone distributions produced by 
each method across occupational status groups. The heatmap makes the asymmetry stark: 
EntiGen and ITI-GEN collapse toward darker tones regardless of occupational status, 
while Fair Diffusion shows uneven correction that deteriorates for low-status roles. 
Our method is the only one that achieves near-uniform coverage across all three status 
groups and all three skin tone bins.

\vspace{-0.9em}
\subsection{Bias Correction in Non-Occupational Prompts}
\begin{figure}[t]
    \centering
    \includegraphics[width=0.95\linewidth]{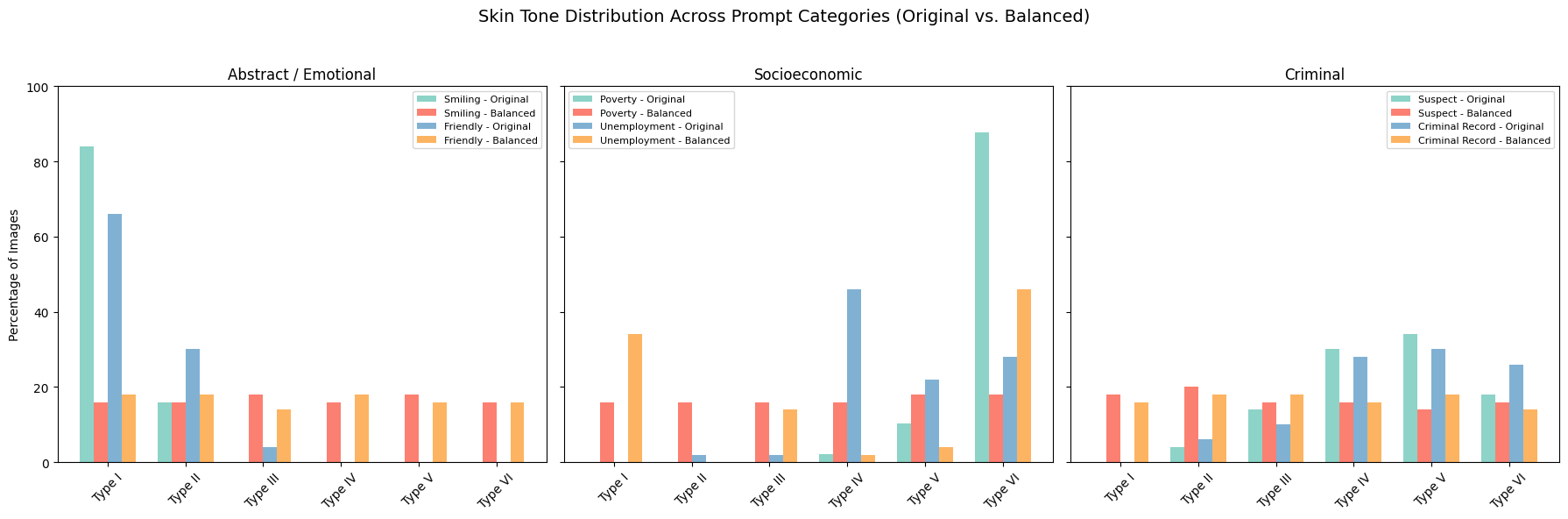}  \caption{Skin tone distributions for three non-occupational prompt groups: (a) emotional/abstract, (b) socioeconomic, and (c) criminal stereotype.}
  \label{fig:socioeconomic_results}
\end{figure}

\begin{figure}[b]
    \centering
    \includegraphics[width=0.95\linewidth]{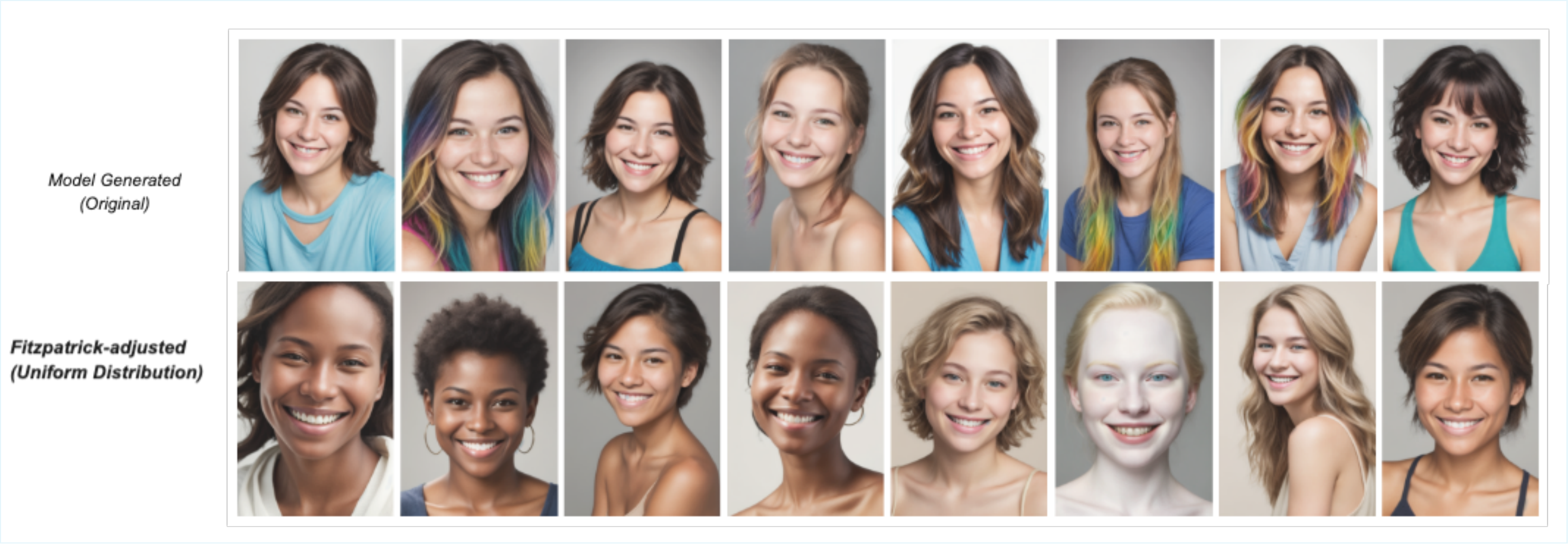}
    \caption{Visual comparison of model-generated and fallback-targeted image outputs for the non-occupational prompt \textit{``A full-color headshot of someone smiling''}. Top: original diffusion model outputs. Bottom: outputs using prompts modified with a uniform Fitzpatrick skin tone distribution.}
    \label{fig:qualitative_identity_comparison}
\end{figure}

\vspace{-0.5em}
We evaluate the Fitzpatrick fallback target on prompts where the LLM returns low
confidence or no demographic data. These include abstract descriptors (e.g., emotions
or personality traits) as well as stereotype-prone social contexts, where reliable
population-level demographic statistics are not available.
As shown in Figure~\ref{fig:socioeconomic_results}a, baseline generations for abstract
prompts such as ``a smiling person'' are dominated by lighter tones (over 80\% in
Fitzpatrick Types~I--II). Under the fallback condition, the system declares a uniform
target over Fitzpatrick Types~I--VI (16.7\% each) and the induced output distribution
moves close to this target, yielding near-uniform coverage across all six types.
A qualitative example is shown in Figure~\ref{fig:qualitative_identity_comparison}.
For stereotype-linked prompts such as ``a person suspected of a crime,'' baseline outputs
exhibit the opposite trend, skewing toward darker tones (Types~IV--VI). The fallback
condition reduces deviation from the uniform Fitzpatrick target, bringing each skin type
within 2\% of the 16.7\% target share (Figure~\ref{fig:socioeconomic_results}c), illustrating
robustness across semantically different prompt types.

\subsubsection{Socioeconomic Prompts}
\label{subsubsec:nonocc}
Figure~\ref{fig:socioeconomic_results}b reports Fitzpatrick skin-tone distributions for
socioeconomic prompts such as ``Someone struggling financially'' and ``a person
without stable housing.'' Baseline generations show skews consistent with common visual
stereotypes associated with poverty. Under the fallback condition, distributions shift
toward near-uniform coverage across all six Fitzpatrick types. Qualitative examples in
Figure~\ref{fig:poverty_examples} illustrate this change: the top row shows baseline outputs,
while the bottom row shows fallback-targeted outputs.
\vspace{-0.9em}

\subsubsection{Criminal-Context Prompts}
Figure~\ref{fig:socioeconomic_results}c shows skin-tone distributions for criminal-context
prompts (e.g., ``a person suspected of a crime,'' ``a person with a criminal record'').
Baseline outputs exhibit a pronounced skew toward darker tones. Under the fallback target,
outputs shift toward the declared uniform distribution. Figure~\ref{fig:crime_examples}
provides qualitative examples (top: baseline; bottom: fallback-targeted).

\vspace{-0.9em}
\paragraph{Generality beyond skin tone.}
The same inference-time targeting mechanism can be applied to other attributes by
declaring an attribute-specific target and generating subgroup prompt variants.
Additional qualitative examples (gender presentation) are included in Appendix~\ref{app:gender_demo}.

\section{Limitations}

Our approach has several limitations. First, demographic targets obtained or structured via an LLM may be inaccurate or reflect biases in sources and model training data. While we log confidence and use a fallback when statistics are unavailable or low-confidence, errors in inferred targets can still propagate to generated outputs.

Second, we evaluate representation using automated skin tone estimation mapped to the Fitzpatrick scale. This provides an operational measure of visual appearance but is coarse and sensitive to factors such as lighting, stylization, and face-detection failures; it also does not capture the full complexity of identity or intersectional attributes.Additionally, our evaluation does not assess concept bleeding---the 
possibility that demographic descriptors in subgroup prompts may unintentionally alter other visual attributes (e.g., setting, clothing, or expression) beyond skin tone. We flag this as a direction for future work.

Finally, since the method operates at inference time without access to training data or model weights, it cannot remove underlying dataset or model biases. It should be viewed as complementary to training-time mitigation and governance efforts, providing a transparent, auditable mechanism for target-conditioned generation rather than a guarantee of fairness.
\vspace{-0.9em}
\section{Ethical Considerations}
\vspace{-0.5em}
This work addresses representational skew in text-to-image generation by making a representation target explicit and auditable. Any target distribution is normative: our system does not claim that user-specified or statistics-derived targets are ``correct''; it operationalizes a declared target and measures alignment. Because
targets are user-controlled, the mechanism could also be misused to enforce exclusionary preferences; we therefore frame the contribution as transparency and controllability rather than a fairness guarantee. Demographic retrieval via LLMs may be biased or erroneous, so we log confidence and sources and use a predefined fallback when reliable statistics are unavailable. 
\vspace{-0.9em}
\section{Conclusion}
\vspace{-0.5em}
Text-to-image systems can reproduce systematic skews in who is visually 
represented, even for seemingly neutral prompts. We presented an inference-time, 
user-controlled framework that makes representation targets explicit and 
operationalizes them through LLM-guided prompt rewriting, without modifying the 
underlying diffusion model. Across occupational and non-occupational prompts, 
target-conditioned prompting substantially shifts measured skin tone outcomes 
relative to baseline generation, and supports graded control under different 
declared target settings. Applied across four text-to-image models---SD Realistic 
Vision v5.1, SDXL Turbo, SD~1.5, and DALL-E~2---our framework consistently reduces 
deviation from the uniform target, demonstrating that prompt-level intervention 
transfers directly across generators without modification. In direct comparison with 
inference-time debiasing baselines, our approach achieves the lowest alignment error 
across all occupational status groups, outperforming the strongest baseline by 65\% 
on average.

Rather than prescribing a single notion of fairness, our approach treats the 
representation target as an input that can be inspected, logged, and audited. This 
reframing clarifies where normative choices enter the pipeline and enables downstream 
accountability: users can compare outputs against the declared target, assess 
trade-offs, and surface disagreements about what counts as ``appropriate'' 
representation in a given context. Future work should examine how users specify 
targets in practice, extend the framework to multiple attributes with careful 
interaction design, and improve auditing when targets and evaluations are defined 
over different demographic abstractions.

\bibliographystyle{ACM-Reference-Format}
\bibliography{samples/sample-base}

@article{chen2023pixart,
  title={Pixart Fast training of diffusion transformer for photorealistic text-to-image synthesis},
  author={Chen, Junsong and Yu, Jincheng and Ge, Chongjian and Yao, Lewei and Xie, Enze and Wu, Yue and Wang, Zhongdao and Kwok, James and Luo, Ping and Lu, Huchuan and others},
  journal={arXiv preprint arXiv:2310.00426},
  year={2023}
}

@article{chardon1991skin,
  title={Skin colour typology and suntanning pathways},
  author={Chardon, Alain and Cretois, Isabelle and Hourseau, Colette},
  journal={International journal of cosmetic science},
  volume={13},
  number={4},
  pages={191--208},
  year={1991},
  publisher={Wiley Online Library}
}

@incollection{treiman2019occupational,
  title={Occupational prestige in comparative perspective},
  author={Treiman, Donald J},
  booktitle={Social Stratification, Class, Race, and Gender in Sociological Perspective, Second Edition},
  pages={260--263},
  year={2019},
  publisher={Routledge}
}

@article{ramesh2022hierarchical,
  title={Hierarchical text-conditional image generation with clip latents},
  author={Ramesh, Aditya and Dhariwal, Prafulla and Nichol, Alex and Chu, Casey and Chen, Mark},
  journal={arXiv preprint arXiv:2204.06125},
  year={2022}
}

@inproceedings{rombach2022high,
  title={High-resolution image synthesis with latent diffusion models},
  author={Rombach, Robin and Blattmann, Andreas and Lorenz, Dominik and Esser, Patrick and Ommer, Bj{\"o}rn},
  booktitle={Proceedings of the IEEE/CVF conference on computer vision and pattern recognition},
  pages={10684--10695},
  year={2022}
}

@article{goodfellow2014generative,
  title={Generative adversarial nets},
  author={Goodfellow, Ian and Pouget-Abadie, Jean and Mirza, Mehdi and Xu, Bing and Warde-Farley, David and Ozair, Sherjil and Courville, Aaron and Bengio, Yoshua},
  journal={Advances in neural information processing systems},
  volume={27},
  year={2014}
}

@article{reed2016generative,
  title={Generative adversarial text to image synthesis},
  author={Reed, Scott and Akata, Zeynep and Yan, Xinchen and Logeswaran, Lajanugen and Schiele, Bernt and Lee, Honglak},
  journal={arXiv preprint arXiv:1605.05396},
  year={2016}
}

@article{ho2020denoising,
  title={Denoising diffusion probabilistic models},
  author={Ho, Jonathan and Jain, Ajay and Abbeel, Pieter},
  journal={Advances in Neural Information Processing Systems},
  volume={33},
  pages={6840--6851},
  year={2020}
}

@article{song2020score,
  title={Score-based generative modeling through stochastic differential equations},
  author={Song, Yang and Sohl-Dickstein, Jascha and Kingma, Diederik P and Kumar, Abhishek and Ermon, Stefano and Poole, Ben},
  journal={arXiv preprint arXiv:2011.13456},
  year={2020}
}

@article{ramesh2021zero,
  title={Zero-shot text-to-image generation},
  author={Ramesh, Aditya and Pavlov, Mikhail and Goh, Gabriel and Gray, Scott and Voss, Chelsea and Radford, Alec and Chen, Mark and Sutskever, Ilya},
  journal={arXiv preprint arXiv:2102.12092},
  year={2021}
}

@article{saharia2022photorealistic,
  title={Photorealistic text-to-image diffusion models with deep language understanding},
  author={Saharia, Chitwan and Chan, William and Saxena, Saurabh and Li, Lala and Whang, Jay and Denton, Emily L and Ghasemipour, Kamyar and Gontijo Lopes, Raphael and Karagol Ayan, Burcu and Salimans, Tim and others},
  journal={Advances in Neural Information Processing Systems},
  volume={35},
  pages={36479--36494},
  year={2022}
}

@article{peebles2023scalable,
  title={Scalable diffusion models with transformers},
  author={Peebles, William and Xie, Saining},
  journal={arXiv preprint arXiv:2212.09748},
  year={2023}
}

@article{bianchi2023easily,
  title={Easily accessible text-to-image generation amplifies demographic stereotypes at large scale},
  author={Bianchi, Federico and Kalluri, Pratyusha and Durmus, Esin and Ladhak, Faisal and Cheng, Myra and Nozza, Debora and Hashimoto, Tatsunori and Jurafsky, Dan and Zou, James and Caliskan, Aylin},
  journal={arXiv preprint arXiv:2211.03759},
  year={2023}
}

@article{ghosh2023person,
  title={'Person'== light-skinned, western man, and sexualization of women of color: Stereotypes in stable diffusion},
  author={Ghosh, Sourojit and Caliskan, Aylin},
  journal={arXiv preprint arXiv:2310.19981},
  year={2023}
}

@article{cho2022dall,
  title={DALL-EVAL: Probing the reasoning skills and social biases of text-to-image generation models},
  author={Cho, Jaemin and Zala, Abhaysinh and Bansal, Mohit},
  journal={arXiv preprint arXiv:2202.04053},
  year={2022}
}

@article{wu2023stable,
  title={Stable diffusion exposed: Gender bias from prompt to image},
  author={Wu, Yankun and Nakashima, Yuta and Garc{\'\i}a, Noa},
  journal={arXiv preprint arXiv:2308.03399},
  year={2023}
}

@inproceedings{bansal2022Enti,
  title={How well can text-to-image generative models understand ethical natural language interventions?},
  author={Bansal, Hritik and Yin, Da and Monajatipoor, Masoud and Chang, Kai-Wei},
  booktitle={Proceedings of the 2022 Conference on Empirical Methods in Natural Language Processing},
  pages={1358--1370},
  year={2022}
}

@article{naik2023social,
  title={Social biases through the text-to-image generation lens},
  author={Naik, Ranjita and Nushi, Besmira},
  journal={arXiv preprint arXiv:2304.06034},
  year={2023}
}

@article{luccioni2023stable,
  title={Stable bias: Evaluating societal representations in diffusion models},
  author={Luccioni, Sasha and Akiki, Christopher and Mitchell, Margaret and Jernite, Yacine},
  journal={arXiv preprint arXiv:2303.11408},
  year={2023}
}

@article{chen2024would,
  title={Would deep generative models amplify bias in future models?},
  author={Chen, Tianwei and Hirota, Yusuke and Otani, Mayu and Garc{\'\i}a, Noa and Nakashima, Yuta},
  journal={arXiv preprint arXiv:2404.08242},
  year={2024}
}

@article{friedrich2023fair,
  title={Fair diffusion: Instructing text-to-image generation models on fairness},
  author={Friedrich, Felix and Brack, Manuel and Struppek, Lukas and Hintersdorf, Dominik and Schramowski, Patrick and Luccioni, Sasha and Kersting, Kristian},
  journal={arXiv preprint arXiv:2302.10893},
  year={2023}
}

@misc{google_skintone_scale,
  author       = {{Google}},
  title        = {Skin Tone by Google},
  howpublished = {\url{https://skintone.google/}},
  note         = {Accessed: 2026-03-23}
}

@article{zhang2023iti,
  title={ITI-GEN: Inclusive text-to-image generation},
  author={Zhang, Cheng and Chen, Xuanbai and Chai, Siqi and Wu, Chen Henry and Lagun, Dmitry and Beeler, Thabo and De la Torre, Fernando},
  journal={arXiv preprint arXiv:2309.05569},
  year={2023}
}

@article{chinchure2023tibet,
  title={TIBET: Identifying and evaluating biases in text-to-image generative models},
  author={Chinchure, Aditya and Shukla, Pushkar and Bhatt, Gaurav and Salij, Kiri and Hosanagar, Kartik and Sigal, Leonid and Turk, Matthew},
  journal={arXiv preprint arXiv:2312.01261},
  year={2023}
}

@article{dinca2024openbias,
  title={OpenBias: Open-set bias detection in text-to-image generative models},
  author={D'Inca, Moreno and Peruzzo, Elia and Mancini, Massimiliano and Xu, Dejia and Goel, Vidit and Xu, Xingqian and Wang, Zhangyang and Shi, Humphrey and Sebe, Nicu},
  journal={arXiv preprint arXiv:2404.07990},
  year={2024}
}

@article{mehrabi2021survey,
  title={A survey on bias and fairness in machine learning},
  author={Mehrabi, Ninareh and Morstatter, Fred and Saxena, Nripsuta and Lerman, Kristina and Galstyan, Aram},
  journal={ACM computing surveys},
  volume={54},
  number={6},
  pages={1--35},
  year={2021}
}

@book{barocas2017fairness,
  title={Fairness and machine learning},
  author={Barocas, Solon and Hardt, Moritz and Narayanan, Arvind},
  year={2017},
  publisher={fairmlbook. org}
}

@article{fitzpatrick1975soleil,
  title={The validity and practicality of sun-reactive skin types I through VI},
  author={Fitzpatrick, Thomas B},
  journal={Archives of dermatology},
  volume={124},
  number={6},
  pages={869--871},
  year={1988}
}

@article{karkkainen2021fairface,
  title={FairFace: Face attribute dataset for balanced race, gender, and age for bias measurement and mitigation},
  author={K{\"a}rkk{\"a}inen, Kimmo and Joo, Jungseock},
  journal={Proceedings of the IEEE/CVF winter conference on applications of computer vision},
  pages={1548--1558},
  year={2021}
}

@article{merler2019diversity,
  title={Diversity in faces},
  author={Merler, Michele and Ratha, Nalini and Feris, Rogerio S and Smith, John R},
  journal={arXiv preprint arXiv:1901.10436},
  year={2019}
}

@inproceedings{he2024debiasing,
  title={Debiasing text-to-image diffusion models},
  author={He, Ruifei and Xue, Chuhui and Tan, Haoru and Zhang, Wenqing and Yu, Yingchen and Bai, Song and Qi, Xiaojuan},
  booktitle={Proceedings of the 1st ACM Multimedia Workshop on Multi-modal Misinformation Governance in the Era of Foundation Models},
  pages={29--36},
  year={2024}
}

@article{huang2025debiasing,
  title={Debiasing Diffusion Model: Enhancing Fairness through Latent Representation Learning in Stable Diffusion Model},
  author={Huang, Lin-Chun and Tsao, Ching Chieh and Su, Fang-Yi and Chiang, Jung-Hsien},
  journal={arXiv preprint arXiv:2503.12536},
  year={2025}
}

@inproceedings{kim2025rethinking,
  title={Rethinking Training for De-biasing Text-to-Image Generation: Unlocking the Potential of Stable Diffusion},
  author={Kim, Eunji and Kim, Siwon and Park, Minjun and Entezari, Rahim and Yoon, Sungroh},
  booktitle={Proceedings of the Computer Vision and Pattern Recognition Conference},
  pages={13361--13370},
  year={2025}
}

@article{kheya2024pursuit,
  title        = {The Pursuit of Fairness in Artificial Intelligence Models: A Survey},
  author       = {Kheya, Tahsin Alamgir and Bouadjenek, Mohamed Reda and Aryal, Sunil},
  journal      = {arXiv preprint arXiv:2403.17333},
  year         = {2024},
  doi          = {10.48550/arXiv.2403.17333}
}

@inproceedings{verma2018fairness,
  title        = {Fairness Definitions Explained},
  author       = {Verma, Sahil and Rubin, Julia},
  booktitle    = {Proceedings of the International Workshop on Software Fairness (FairWare '18)},
  year         = {2018},
  doi          = {10.1145/3194770.3194776}
}

@inproceedings{saxena2019,
  title        = {How Do Fairness Definitions Fare? Examining Public Attitudes Towards Algorithmic Definitions of Fairness},
  author       = {Saxena, Nripsuta Ani and Huang, Karen and DeFilippis, Evan and Radanovic, Goran and Parkes, David C. and Liu, Yang},
  booktitle    = {Proceedings of the AAAI/ACM Conference on AI, Ethics, and Society (AIES '19)},
  year         = {2019},
  doi          = {10.1145/3306618.3314248}
}

@inproceedings{fraser2023,
  title     = {Diversity is Not a One-Way Street: Pilot Study on Ethical Interventions for Racial Bias in Text-to-Image Systems},
  author    = {Fraser, Kathleen C. and Kiritchenko, Svetlana and Nejadgholi, Isar},
  booktitle = {Proceedings of the 14th International Conference on Computational Creativity (ICCC)},
  year      = {2023},
  pages     = {288--292}
}

@article{wan2024,
  title={Survey of Bias In Text-to-Image Generation: Definition, Evaluation, and Mitigation},
  author={Wan, Yixin and Subramonian, Arjun and Ovalle, Anaelia and Lin, Zongyu and Suvarna, Ashima and Chance, Christina and Bansal, Hritik and Pattichis, Rebecca and Chang, Kai-Wei},
  journal={arXiv preprint arXiv:2404.01030},
  year={2024}
}

@inproceedings{prerak2024,
  title={Addressing Bias in Text-to-Image Generation: A Review of Mitigation Methods},
  author={Prerak, Shah},
  booktitle={2024 Third International Conference on Smart Technologies and Systems for Next Generation Computing (ICSTSN)},
  year={2024},
  doi={10.1109/ICSTSN61422.2024.10671230}
}

@article{saunders2004fairgo,
  title        = {What is Fair About a 'Fair Go'?},
  author       = {Saunders, Peter},
  journal      = {Policy},
  volume       = {20},
  number       = {1},
  pages        = {3--10},
  year         = {2004}
}

@inproceedings{deng2020retinaface,
  title={Retinaface: Single-shot multi-level face localisation in the wild},
  author={Deng, Jiankang and Guo, Jia and Ververas, Evangelos and Kotsia, Irene and Zafeiriou, Stefanos},
  booktitle={Proceedings of the IEEE/CVF conference on computer vision and pattern recognition},
  pages={5203--5212},
  year={2020}
}

\appendix

\section{Appendix: Prompt List}

\subsection{Occupational Prompts}
We used 30 occupational prompts grouped into three status categories: High, Moderate, and Low. Each prompt follows the standardized template: 
\textit{"A full-color headshot of a [OCCUPATION]"}.

\begin{table}[h!]
\centering
\small
\begin{tabular}{p{0.28\linewidth} p{0.28\linewidth} p{0.28\linewidth}}
\toprule
\textbf{High-Status} & \textbf{Moderate-Status} & \textbf{Low-Status} \\
\midrule
Doctor               & Teacher          & Construction worker \\
Lawyer               & Nurse            & Janitor \\
University professor & Police officer   & Factory worker \\
CEO                  & Firefighter      & Farm worker \\
Scientist            & Social worker    & Maintenance worker \\
Engineer             & Paramedic        & Landscaper \\
Judge                & Librarian        & Food service worker \\
Surgeon              & Counselor        & Warehouse worker \\
Architect            & Postal worker    & Security guard \\
Financial advisor    & Bus driver       & Delivery driver \\
\bottomrule
\end{tabular}
\caption{List of occupational prompts grouped by social prestige categories. Each prompt follows the format \textit{``A full-color headshot of a [OCCUPATION]''}.}
\label{tab:occupational_prompts}
\end{table}

\subsection{Non-Occupational Prompts}

In addition to occupational prompts, we evaluated demographic bias on six non-occupational prompts that reflect abstract, socioeconomic, and stereotype-related contexts. Each follows the same template: 
\textit{"A full-color headshot of [DESCRIPTION]"}.

\begin{table}[h!]
\centering
\small
\begin{tabular}{p{0.4\linewidth} p{0.55\linewidth}}
\toprule
\textbf{Category} & \textbf{Prompts} \\
\midrule
Criminal stereotype & A person suspected of a crime; a person with a criminal record \\
Socioeconomic       & Someone struggling financially; person without stable housing \\
Abstract/Emotional  & Someone smiling; a friendly person \\
\bottomrule
\end{tabular}
\caption{Non-occupational prompts used for bias evaluation. Each follows the format \textit{``A full-color headshot of [DESCRIPTION]''}.}
\label{tab:nonoccupational_prompts}
\end{table}

To avoid reinforcing harmful stereotypes, we intentionally phrased certain non-occupational prompts in an ambiguous and neutral manner. For example, rather than using explicit labels such as ``a criminal'' or ``a person in poverty,'' we adopted softer alternatives like ``a person suspected of a crime'' and ``a person struggling financially.''

This approach allows us to evaluate bias in stereotype-prone contexts without directly encoding stigmatizing language, aligning with best practices in fairness research.

\subsection{Full Occupation List with SIOPS Scores}
Table~\ref{tab:occupations} lists all occupations used in our experiments with their corresponding SIOPS scores.

\begin{table}[h!]
    \centering
    \small
    \begin{tabular}{llc}
        \toprule
        \textbf{Occupation} & \textbf{Status} & \textbf{SIOPS Score} \\
        \midrule
        Lawyer & High & 78 \\
        Engineer & High & 74 \\
        Judge & High & 77 \\
        Doctor & High & 78 \\
        CEO & High & 75 \\
        Architect & High & 71 \\
        Scientist & High & 71 \\
        University Professor & High & 71 \\
        Financial Advisor & High & 69 \\
        \midrule
        Teacher (High School) & Moderate & 61 \\
        Nurse & Moderate & 66 \\
        Police Officer & Moderate & 60 \\
        Firefighter & Moderate & 62 \\
        Social Worker & Moderate & 58.2 \\
        Counselor & Moderate & 56.8 \\
        Librarian & Moderate & 54 \\
        Paramedic & Moderate & 53.5 \\
        Postal Worker & Moderate & 45.3 \\
        Bus Driver & Moderate & 47 \\
        \midrule
        Delivery Driver & Low & 42.5 \\
        Security Guard & Low & 48.1 \\
        Maintenance Worker & Low & 44 \\
        Landscaper & Low & 43.7 \\
        Construction Worker & Low & 28 \\
        Factory Worker & Low & 22 \\
        Farm Worker & Low & 22 \\
        Warehouse Worker & Low & 28 \\
        Janitor & Low & 15 \\
        Food Service Worker & Low & 35 \\
        \bottomrule
    \end{tabular}
    \caption{Full list of occupations with their status categories and SIOPS scores.}
    \label{tab:occupations}
\end{table}

\section{Appendix: Demographic Target Settings}
\label{app:targets}

This appendix specifies how we instantiate the three target settings used in
\S5.3. Let $G=\{g_1,\dots,g_m\}$ denote the set of demographic groups returned by
the LLM for a given prompt and scope, along with a normalized distribution
$r$ over $G$ (e.g., proportions estimated from external statistics). Our module
constructs a \emph{declared target} distribution $q$ over the same set $G$ and
allocates a generation budget of $N$ images across subgroup prompts according to
$N_i \approx q(g_i)\,N$ (rounding as needed).

\subsection{Uniform Target}
\label{app:uniform}
The Uniform setting assigns equal weight to all groups in $G$:
\begin{equation}
q_{\text{uni}}(g_i) = \frac{1}{m}, \quad i=1,\dots,m.
\end{equation}
This target is independent of $r$ and tests whether the system can produce
approximately even representation across the groups returned for a prompt.

\subsection{Intermediate Target}
\label{app:intermediate}
The Intermediate setting produces a \emph{moderately skewed} target that preserves
support over all groups while increasing mass on the majority group. Let
\begin{equation}
g^{\ast} = \arg\max_{g \in G} r(g)
\end{equation}
denote the majority group under the LLM-returned distribution. We define Intermediate
as a convex mixture of the LLM distribution and the Uniform target:
\begin{equation}
q_{\text{int}} = \alpha\, r + (1-\alpha)\, q_{\text{uni}},
\end{equation}
where $\alpha \in (0,1)$ controls the degree of skew. Unless otherwise noted, we
use \textbf{$\alpha=0.5$}, which yields a target that is more concentrated than
Uniform while avoiding single-group collapse.

\paragraph{Implementation note.}
We apply no thresholding: all groups in $G$ retain non-zero mass under
$q_{\text{int}}$, ensuring every group receives a non-zero allocation.

\subsection{Extreme Target}
\label{app:extreme}
The Extreme setting specifies a \emph{concentrated} target in which a single focal
group receives the majority of probability mass. Let $g^\dagger \in G$ denote the
selected focal group and let $\alpha \in (0.5,1]$ control the concentration. Let
$s$ be a reference distribution over $G$ (we use the LLM-inferred distribution $r$).
We define:
\begin{equation}
q_{\text{ext}}(g;\alpha,g^\dagger)=
\begin{cases}
\alpha, & g=g^\dagger,\\[4pt]
(1-\alpha)\,\dfrac{s(g)}{1-s(g^\dagger)}, & g\neq g^\dagger.
\end{cases}
\label{eq:extreme}
\end{equation}
When $\alpha=1$, Eq.~\ref{eq:extreme} reduces to the degenerate single-group target
that assigns all mass to $g^\dagger$.

\paragraph{Choosing $g^\dagger$.}
Unless otherwise stated, we select $g^\dagger$ from the LLM-returned set $G$ as
(i) the majority group $g^{\ast}$, or (ii) a non-majority group (e.g., the smallest
non-zero group among the top-$k$ groups). We report the selected focal group
alongside the logged target for transparency.

\subsection{From Targets to Generation Budgets}
\label{app:budget}
Given a target $q$ and a total budget of $N$ images for an occupation under a
target-conditioned setting, we allocate:
\begin{equation}
N_i = \mathrm{round}\big(q(g_i)\,N\big)
\end{equation}
and adjust by $\pm1$ as needed to ensure $\sum_i N_i = N$ (e.g., distributing any
remainder to groups with the largest fractional parts). Each subgroup prompt
corresponding to $g_i$ is then used to generate $N_i$ images.

\paragraph{Relation to evaluation.}
Targets $q$ are declared over demographic labels (race/ethnicity) returned by the
LLM, while our automated evaluation reports skin tone outcomes. We therefore use
Uniform/Intermediate/Extreme targets to study how changing the declared target
influences observed skin tone distributions and to support auditing via logged
targets and allocations.

\section{Appendix: Intermediate Target Reporting}
\label{app:intermediate_details}

This appendix summarizes how we report outcomes under the \emph{Intermediate}
target setting. The operational definition of Intermediate is given in
Appendix~\ref{app:intermediate}.

\subsection{Why Intermediate is a Controllability Test}
\label{app:int_rationale}
Intermediate targets are declared over demographic labels (race/ethnicity), while
our automated evaluation reports \emph{skin tone} (Fitzpatrick). Because these
representations are not in one-to-one correspondence, we do not expect perfect
matching between a demographic-label target and a single Fitzpatrick bin.
Instead, Intermediate serves as a test of \emph{graded controllability}: whether
moving from baseline $\rightarrow$ Uniform $\rightarrow$ Intermediate (non-uniform)
produces predictable, directional shifts in skin tone outcomes.
\subsection{Reporting Intermediate Outcomes in Skin-Tone Space}
\label{app:int_reporting}
For each occupation, we compute the empirical skin-tone distribution over Fitzpatrick types and aggregate them into three bins: Light (I--II), Medium (III--IV), and Dark(V--VI). 
\vspace{-0.9em}
\section{Intermediate and Extreme Target Results}
\label{app:intermediate_results}

\subsection{Intermediate Target}

We test a non-uniform \emph{Intermediate} setting that specifies a moderately 
skewed target over the demographic groups returned by the LLM for each 
occupation (formal definition in Appendix~\ref{app:intermediate}). Because 
subgroup prompts are expressed using demographic labels (e.g., race/ethnicity) 
while our evaluation measures \emph{skin tone}, we treat this experiment as a 
\emph{controllability} test: whether changing the declared target produces a 
predictable, graded shift in observed skin tone outcomes.

Across occupational groups, Intermediate targets yield distributions that are 
less concentrated than baseline but do not enforce full uniformity. The induced 
skin tone distribution typically falls between the baseline skew and the Uniform 
condition, indicating that the module supports \emph{graded control} rather than 
a single fixed balancing rule. Under Intermediate targets, high-status 
occupations exhibit an average Light/Medium/Dark distribution of 
$(0.482,\,0.256,\,0.262)$ (mean$\pm$std across occupations: 
$0.482\pm0.096$, $0.256\pm0.072$, $0.262\pm0.094$). Relative to the Uniform 
condition, this outcome retains moderate skew toward lighter tones while 
increasing representation of medium and darker tones compared to baseline, 
consistent with the intended ``in-between'' behavior of Intermediate targets. 
We observe the same graded trend for moderate- and low-status occupations, 
indicating that the module supports adjustable representation targets rather 
than enforcing a single fixed balancing rule.
\begin{figure}[h]
    \centering
    \includegraphics[width=\linewidth]{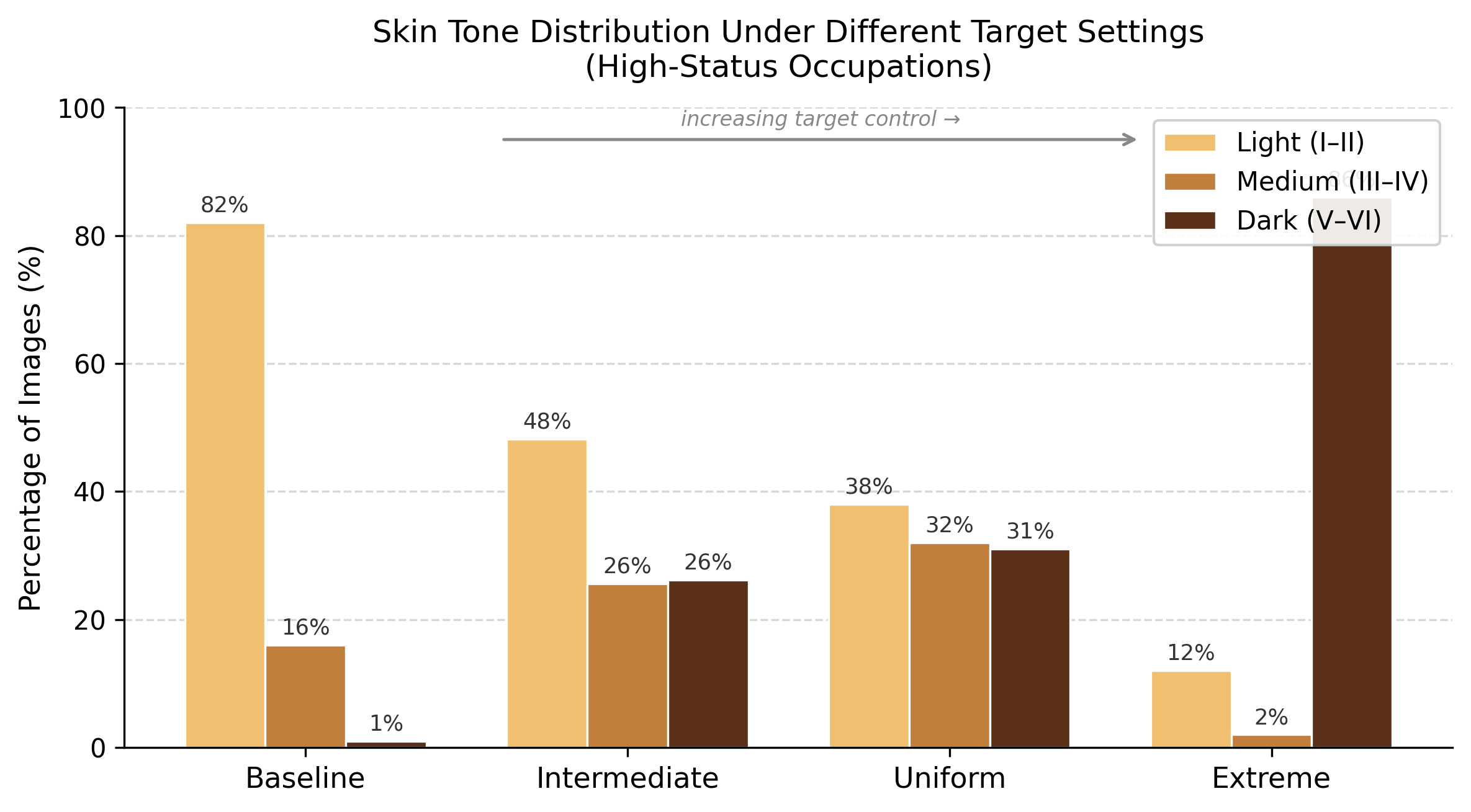}
    \caption{Skin tone distribution (Light I--II, Medium III--IV, Dark V--VI) 
    for high-status occupations across four declared target settings: Baseline 
    (no intervention), Intermediate (moderately skewed), Uniform (equal weight), 
    and Extreme (diagnostic stress test).The progression demonstrates that the 
    framework supports graded, monotonic control over the output distribution 
    rather than a single fixed balancing rule.}
    \label{fig:target_progression}
\end{figure}
\subsection{Extreme Target (Diagnostic)}

We evaluate an \emph{Extreme} (highly concentrated) target as a diagnostic 
stress test of controllability and auditability, rather than as a normative 
fairness objective. In this setting, we ask whether declaring a strongly skewed 
demographic target produces a correspondingly strong shift in observed skin tone 
outcomes.For high-status prompts under an extreme target (Black=80\%, Indian=10\%, 
White=5\%, Asian=5\%), the resulting Fitzpatrick distribution becomes heavily 
concentrated in darker tones: Type~VI accounts for 43/50 images (86\%), with 
the remaining images spread across Types~I--III (Type~I=10\%, Type~II=2\%, 
Type~III=2\%; Types~IV--V=0\%). Aggregating to Light/Medium/Dark bins, this 
corresponds to (12\%, 2\%, 86\%). While perfect collapse to a single Fitzpatrick 
bin is not expected---since targets are declared over demographic labels while 
evaluation measures skin tone---the strong directional shift toward darker tones 
demonstrates that extreme targets induce large, measurable changes in the output 
distribution, confirming the system's auditability under stress conditions.

\section{ Additional Results : Beyond Occupational Prompts}
\begin{figure*}[h]
    \centering
    \includegraphics[width=0.95\linewidth]{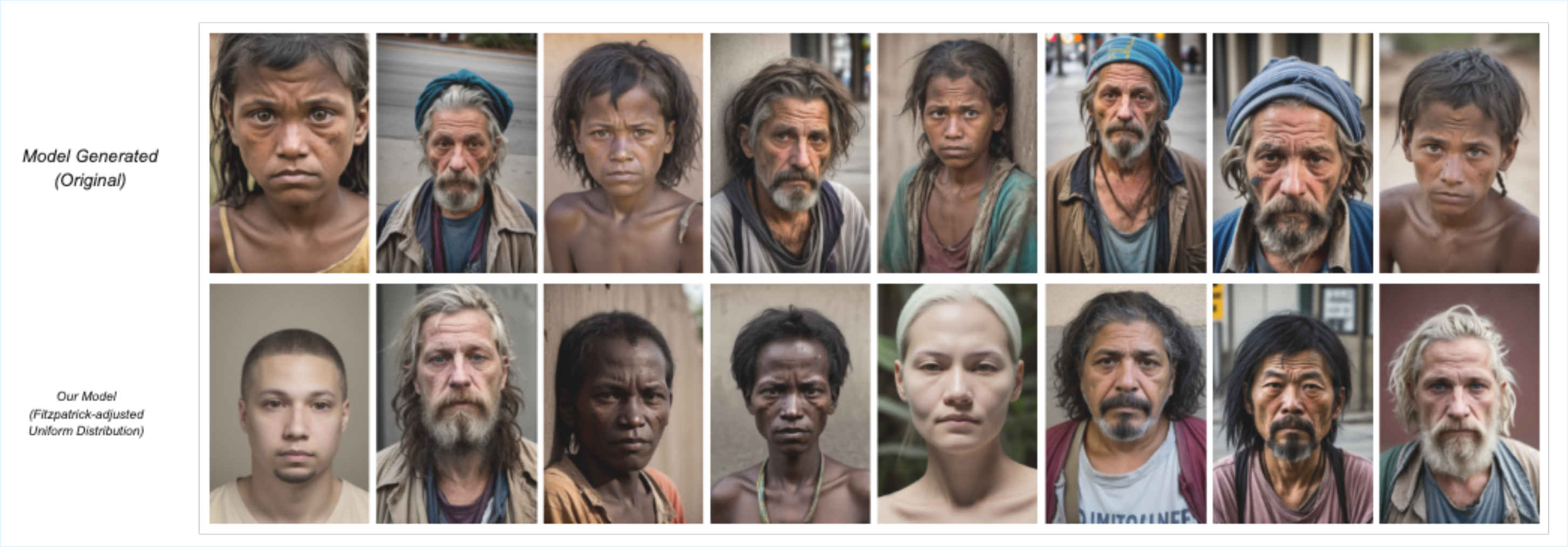}
   
    \caption{Visual comparison of model-generated and fallback-targeted image outputs for the non-occupational prompt “A full-color
headshot of someone struggling financially”. Original outputs (first row) reveal biased distributions. After applying the fallback strategy (second row), generations achieve near-uniform balance while maintaining image quality and diversity.}
    \label{fig:poverty_examples}
\end{figure*}

\begin{figure}[h]
    \centering
    \includegraphics[width=\linewidth]{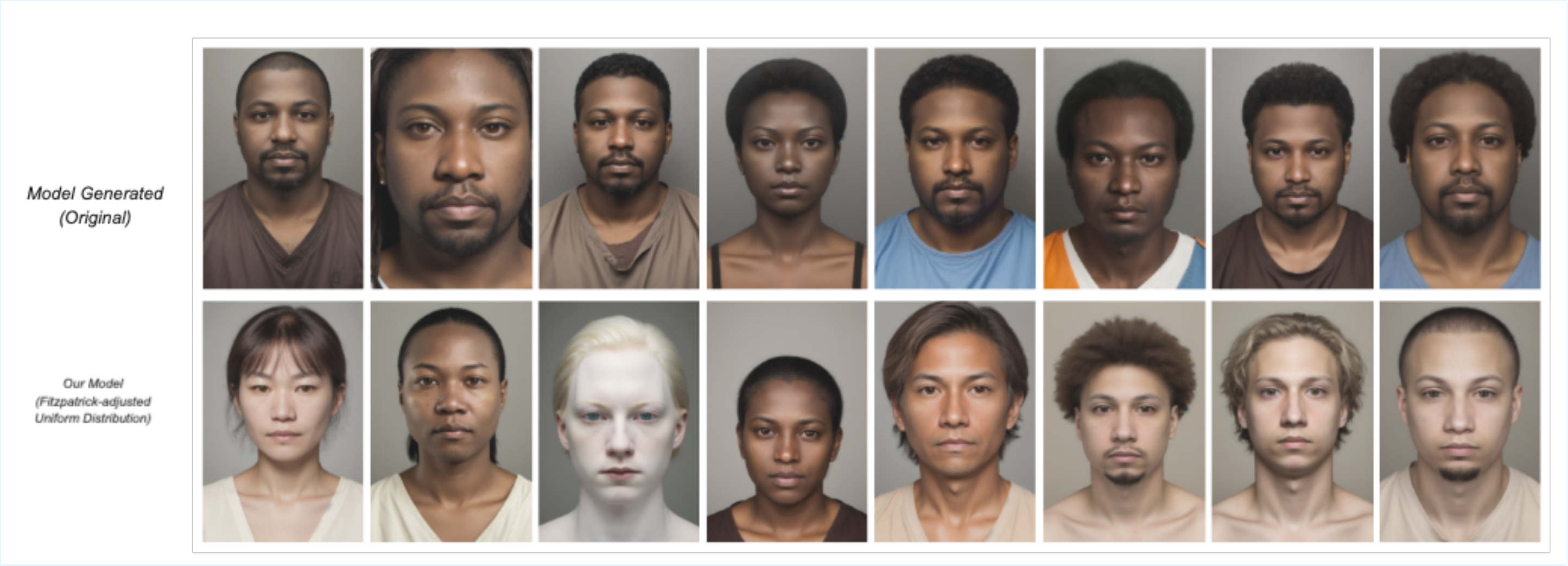}

    \caption{Visual comparison of model-generated and fallback-targeted image outputs for the non-occupational prompt “A full-color
headshot of a person suspected of crime”. The first row shows original model outputs, which exhibit a strong skew toward darker skin tones. The second row displays results after applying the Fitzpatrick fallback strategy, yielding a balanced set across all six skin tone types.}
    \label{fig:crime_examples}
\end{figure}

Figures~\ref{fig:poverty_examples} and~\ref{fig:crime_examples} show qualitative 
examples of baseline and fallback-targeted outputs for two stereotype-prone 
non-occupational prompts. In both cases, baseline generations exhibit pronounced 
skin tone skew consistent with common visual stereotypes---lighter tones dominating 
socioeconomic prompts and darker tones dominating criminal-context prompts. Under 
the uniform Fitzpatrick fallback target, outputs shift toward near-uniform coverage 
across all six skin tone types while maintaining image quality and diversity. These 
examples complement the quantitative results reported in Section~\ref{subsubsec:nonocc}
\vspace{0.9em}

\section{Additional Results: Generalization Beyond Skin Tone}
\label{app:gender_demo}

Although our main study focuses on skin tone, the same inference-time mechanism extends
to other attributes by declaring an additional target distribution and generating
subgroup-specific prompt variants. We illustrate this with gender presentation as an
additional target attribute (uniform target for this demonstration).

\subsection{Occupational Prompts}
\begin{figure*}[h]
    \centering
    \includegraphics[width=\linewidth]{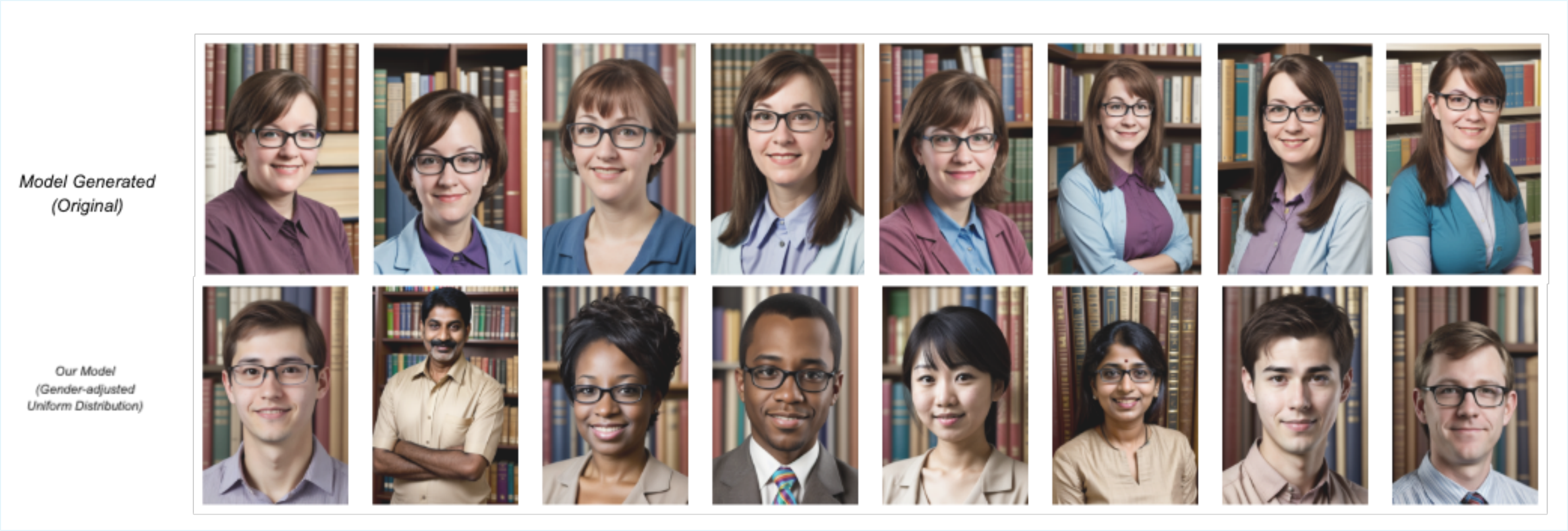}
    \includegraphics[width=\linewidth]{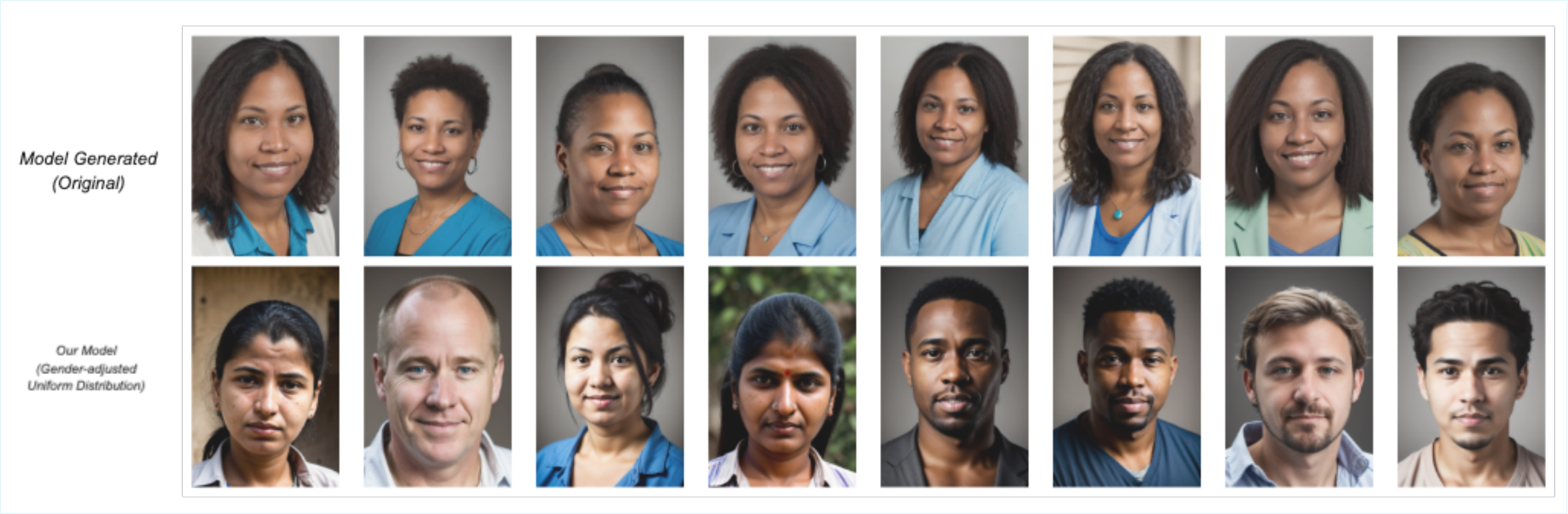}
    \caption{Target-conditioned prompting for gender presentation in occupational prompts.
    Baseline outputs (top) for ``librarian'' and ``social worker'' are visually female-skewed.
    Under a declared uniform gender target (bottom), outputs become more mixed.}
    \label{fig:gender_female_skew}
\end{figure*}

For some occupations (e.g., ``librarian'' and ``social worker''), baseline generations are
visually dominated by female-presenting faces. Under a declared uniform gender target,
the framework produces a more mixed set of gender-presenting outputs without retraining
or access to model parameters (Fig.~\ref{fig:gender_female_skew}). These examples are
included to demonstrate controllability and generality of the targeting mechanism.

\subsection{Non-Occupational Prompts}
\begin{figure*}[h]
    \centering
    \includegraphics[width=\linewidth]{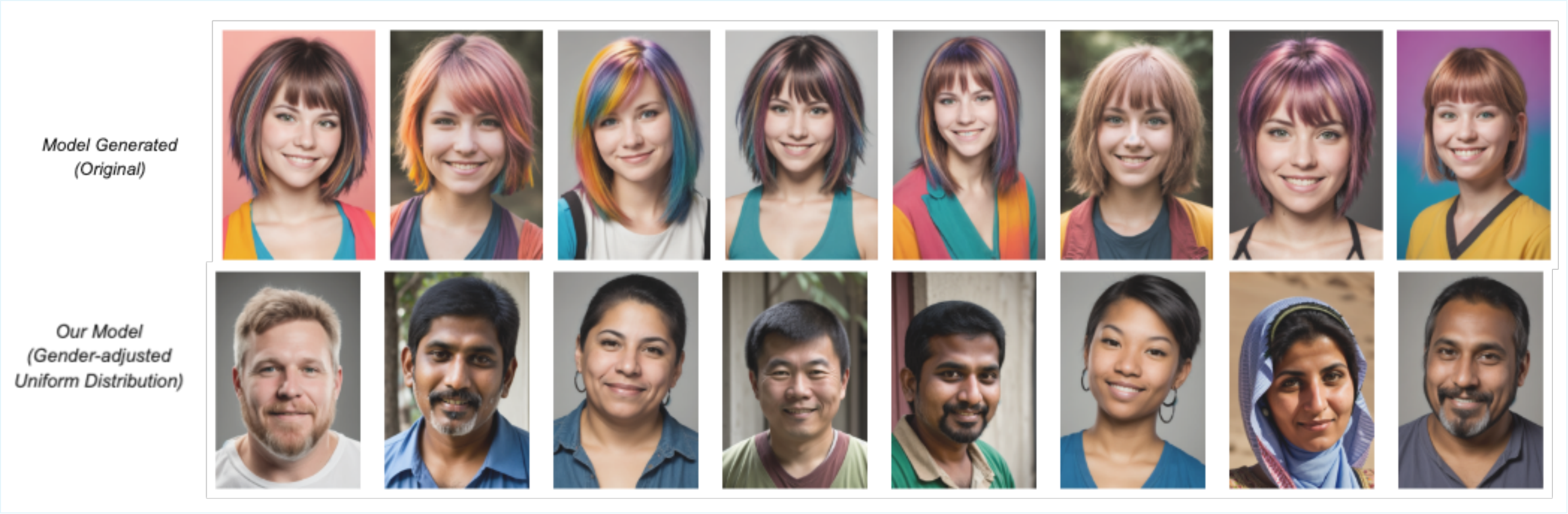}
    \caption{Target-conditioned prompting for gender presentation in a non-occupational prompt.
    Baseline outputs for ``a friendly person'' (top) are visually female-skewed. Under a
    declared uniform gender target (bottom), outputs become more mixed.}
    \label{fig:friendly_gender}
\end{figure*}

Gender skews also appear in abstract prompts such as ``a friendly person,'' where reliable
demographic statistics are unavailable. Using the same approach with a declared uniform
gender target, the framework produces a more mixed set of gender-presenting outputs
(Fig.~\ref{fig:friendly_gender}).

\paragraph{Limitation.}
These demonstrations operationalize gender as perceived presentation and use a binary
target for simplicity; extending to other gender identities requires careful definition
of categories and evaluation protocols.
\clearpage

\section{Qualitative Results Across T2I Models}
\label{app:qualitative_models}
\begin{figure*}[h!]
    \centering
    \includegraphics[width=\linewidth]{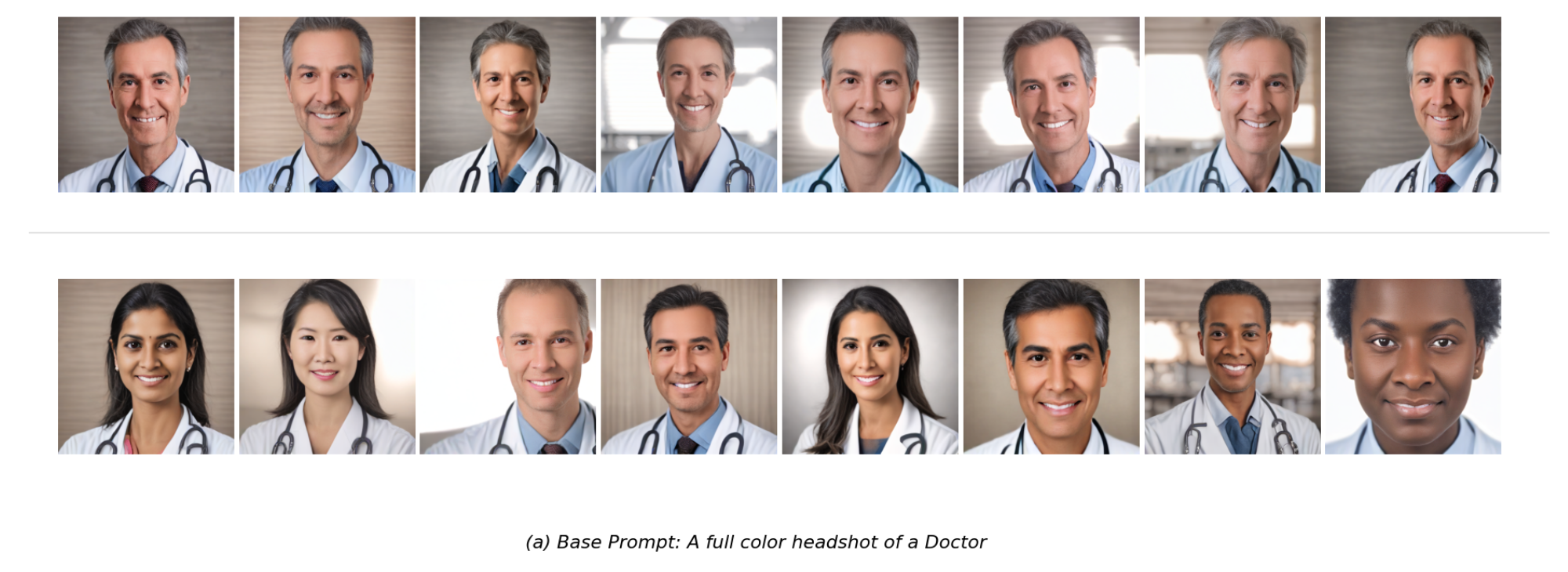}
    \vspace{0.5em}
    \includegraphics[width=\linewidth]{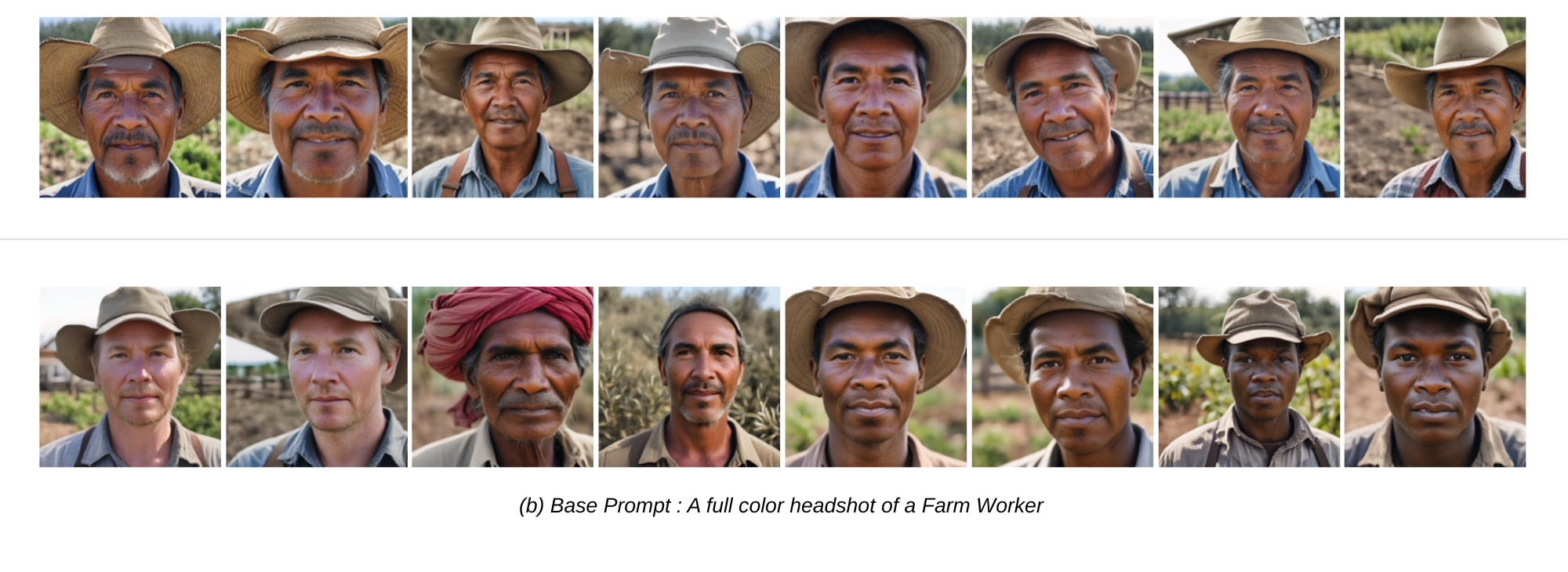}
    \vspace{0.5em}
    \includegraphics[width=\linewidth]{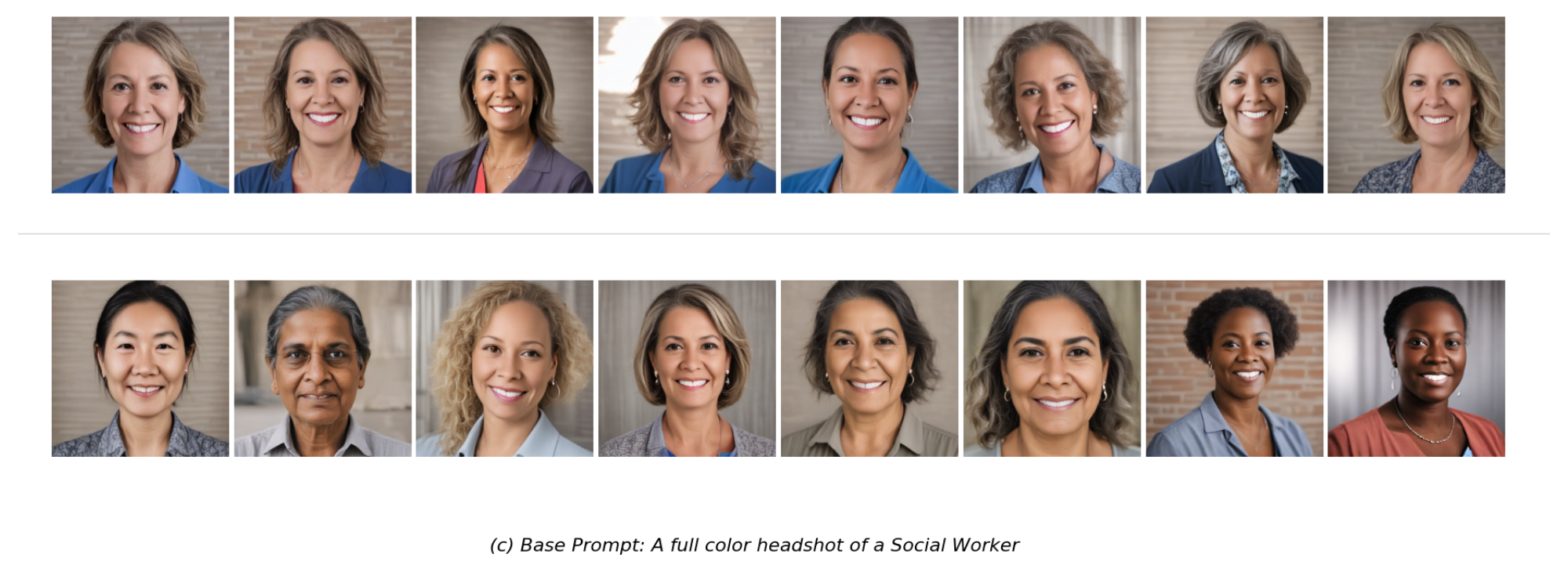}
    \caption{Qualitative examples for SDXL Turbo across representative high-,
    moderate-, and low-status occupational prompts.
    Top rows: baseline outputs.
    Bottom rows: outputs under our declared uniform target.}
    \label{fig:appendix_sdxl}
\end{figure*}

\begin{figure*}[h!]
    \centering
    \includegraphics[width=\linewidth]{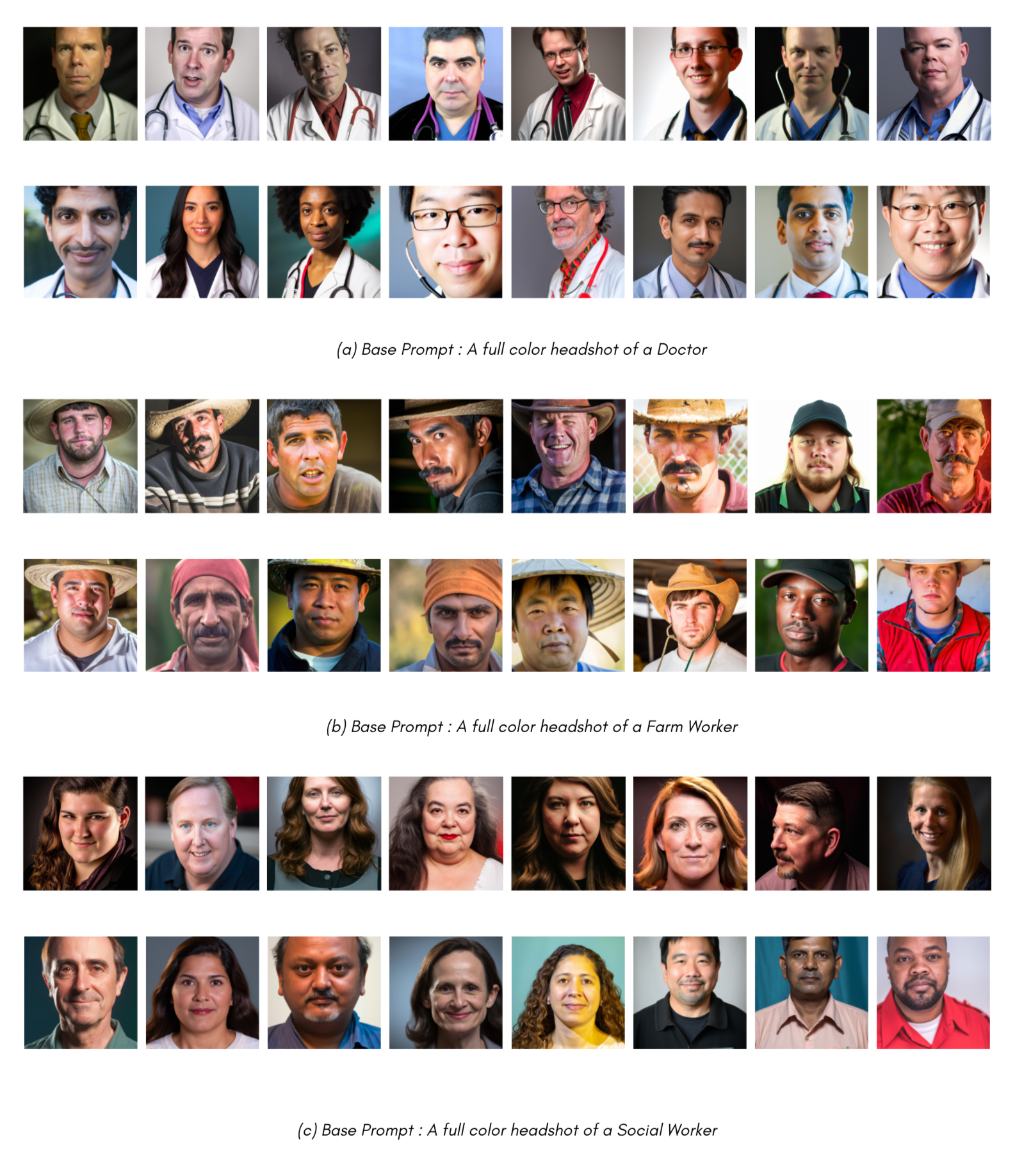}
    \caption{Qualitative comparison of occupational image generations before and after
    prompt adjustment for DALL-E~2. The top rows show the original model outputs, while
    the bottom rows show the fairness-controlled outputs produced by our method.}
    \label{fig:appendix_dalle3}
\end{figure*}

Figures~\ref{fig:appendix_sdxl}--\ref{fig:appendix_dalle3} show qualitative
examples of baseline and target-conditioned outputs generated with SDXL Turbo,
SD~1.5, and DALL-E~2 under the Uniform target setting, complementing the
SD Realistic Vision v5.1 examples shown in Figure~2 of the main paper.
In all cases, our method produces visibly more diverse skin-tone representations
relative to the no-intervention baseline, consistent with the quantitative
results reported in Table~\ref{tab:variance_scores}.

\section{Skin Tone Distributions Across T2I Models}
\label{app:skin_tone_models}

\begin{figure*}[h!]
    \centering
    \includegraphics[width=0.32\linewidth]{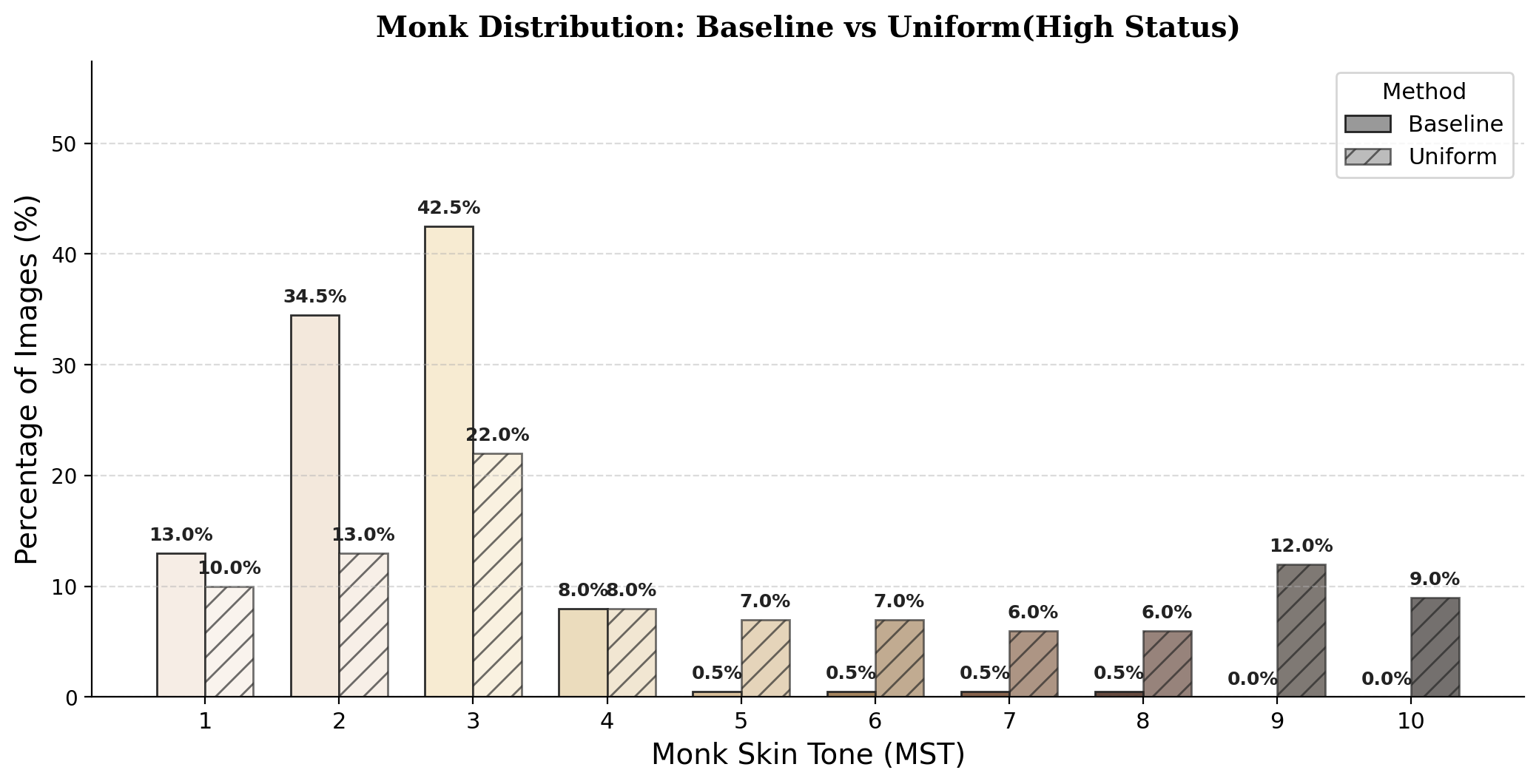}
    \hfill
    \includegraphics[width=0.32\linewidth]{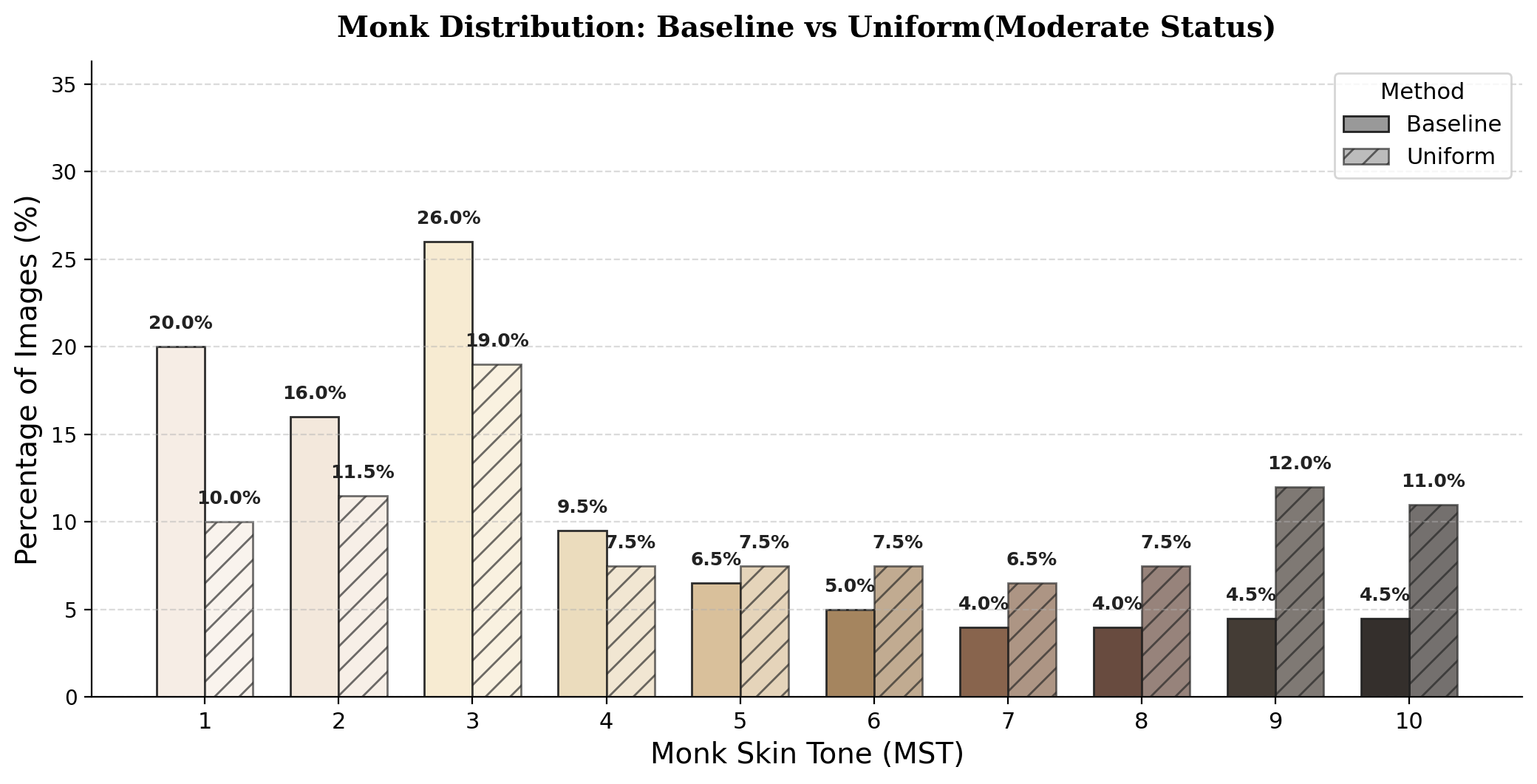}
    \hfill
    \includegraphics[width=0.32\linewidth]{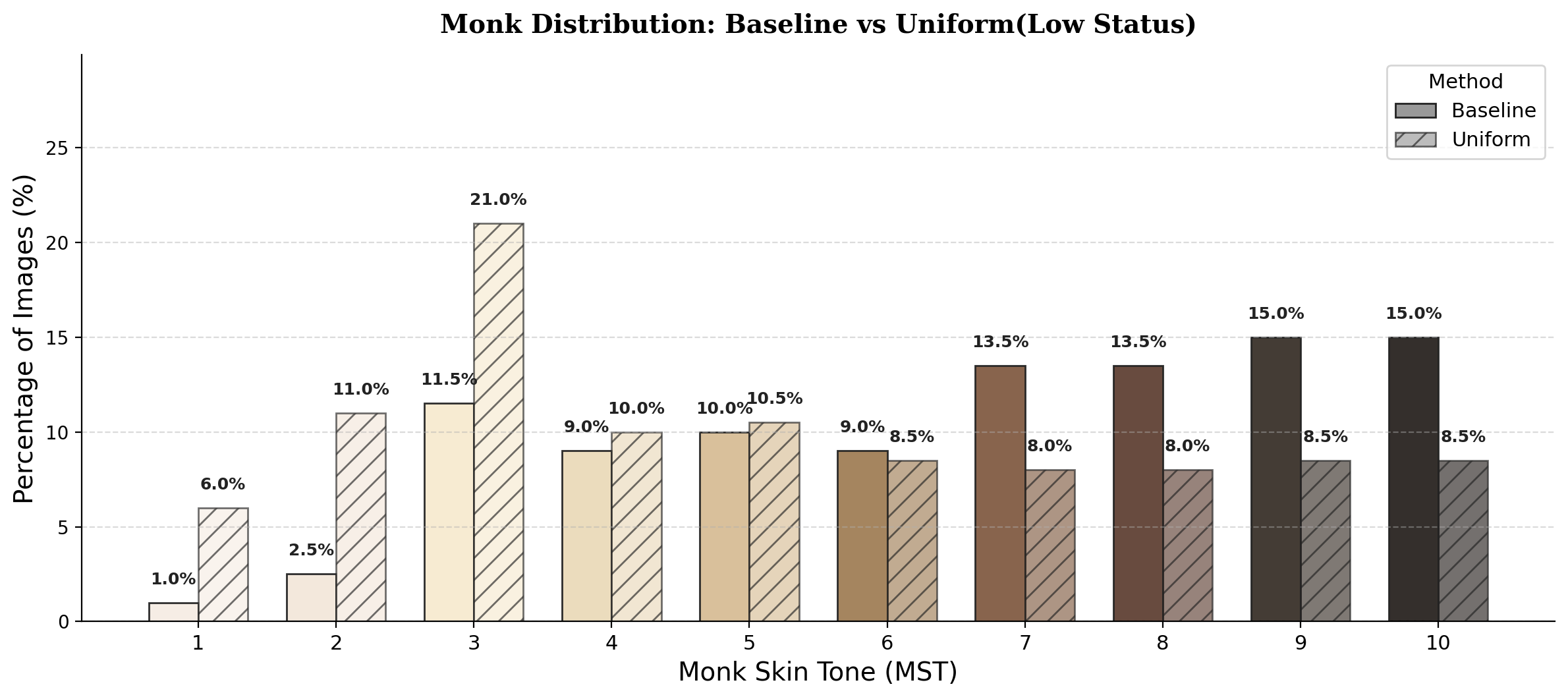}
    \caption{Monk Skin Tone distributions (MST 1--10) across high-, moderate-, and
    low-status occupational groups before (solid) and after (hatched) applying our
    target-conditioned prompting framework. Model: SD Realistic Vision v5.1.} Results
    are consistent with the Fitzpatrick distributions shown in
    Figure~\ref{fig:status_fitz_charts}, confirming that the observed shifts hold
    across both skin tone scales.
    \label{fig:appendix_monk_rv}
\end{figure*}

\begin{figure*}[h!]
    \centering
    \includegraphics[width=0.32\linewidth]{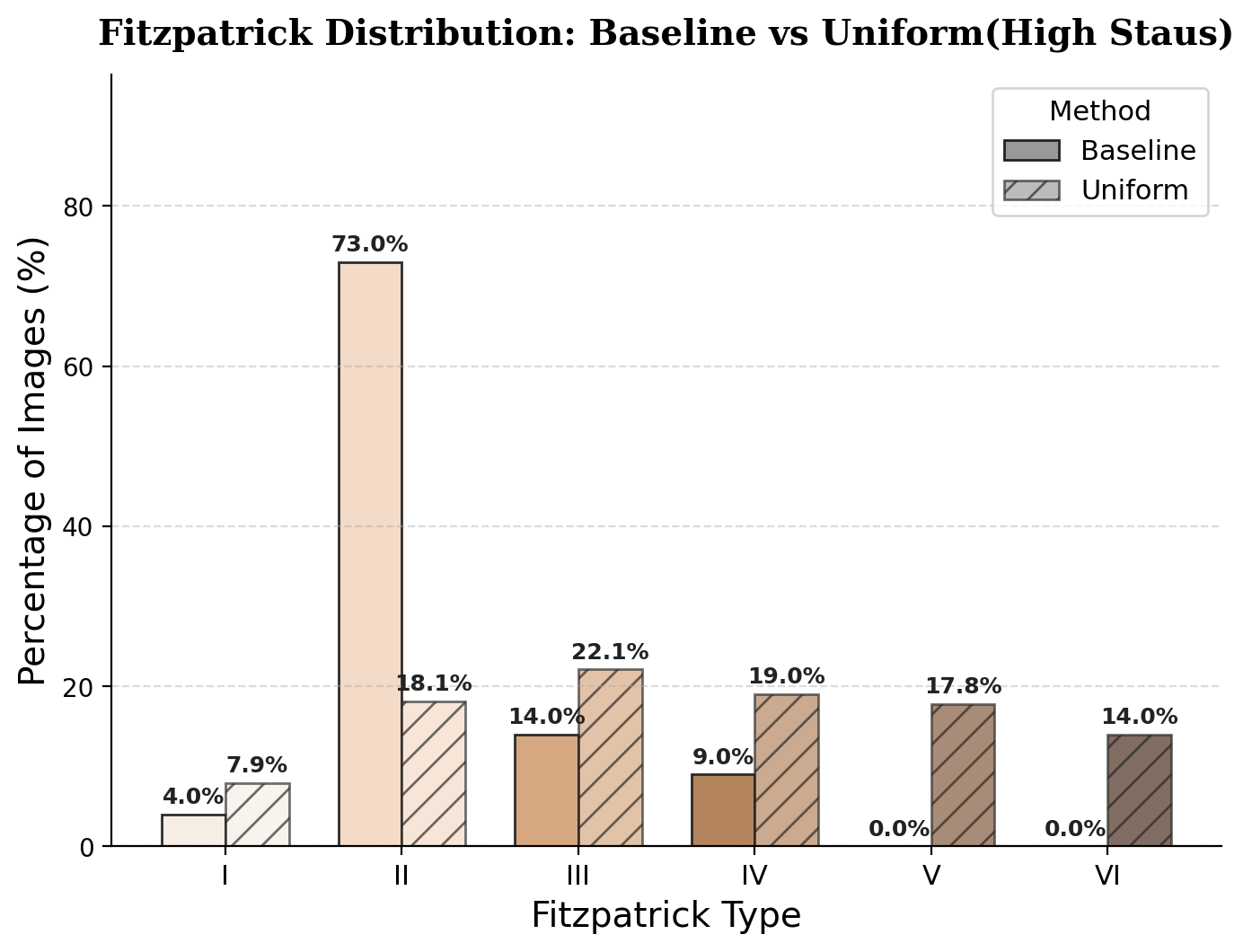}
    \hfill
    \includegraphics[width=0.32\linewidth]{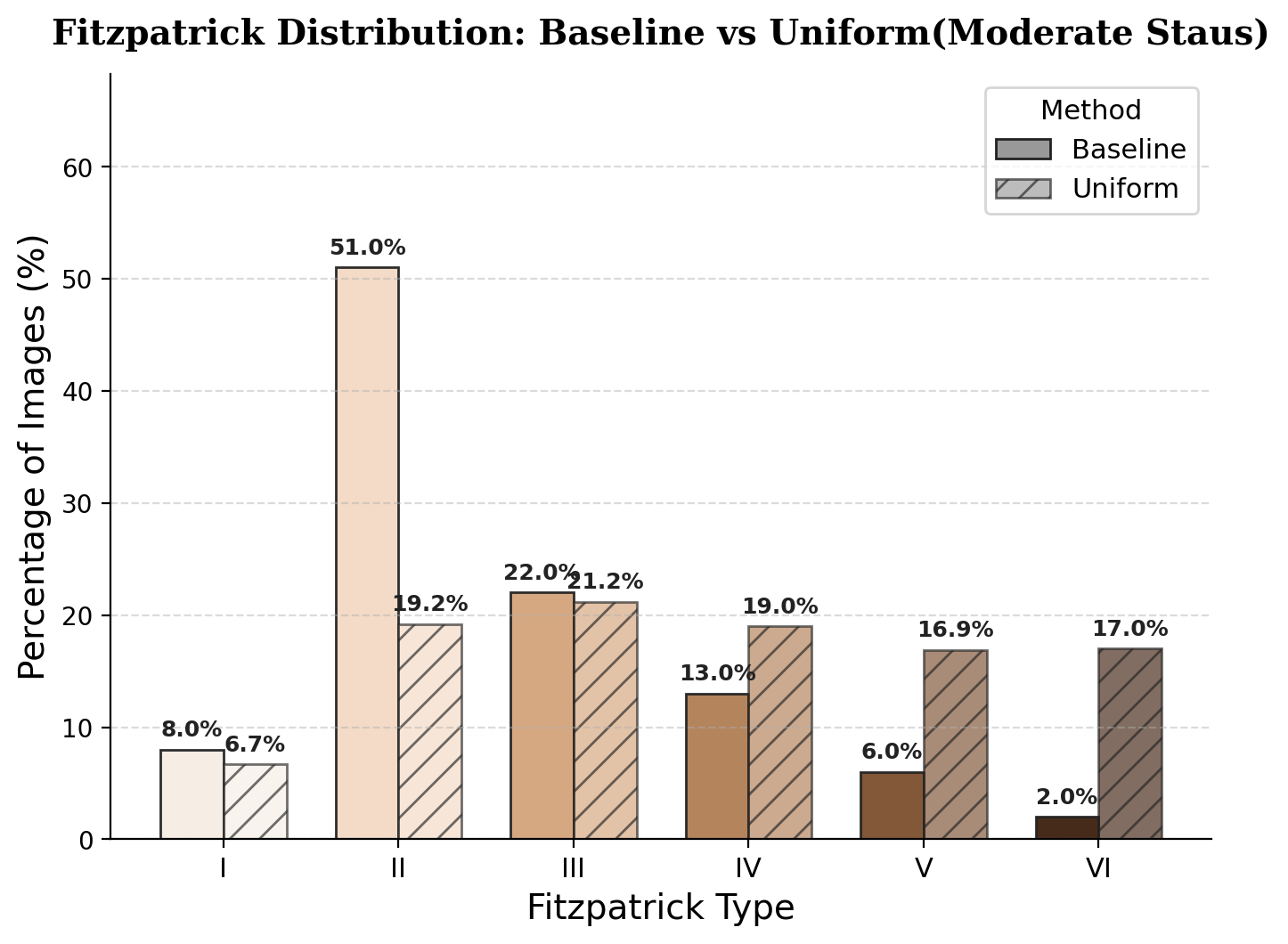}
    \hfill
    \includegraphics[width=0.32\linewidth]{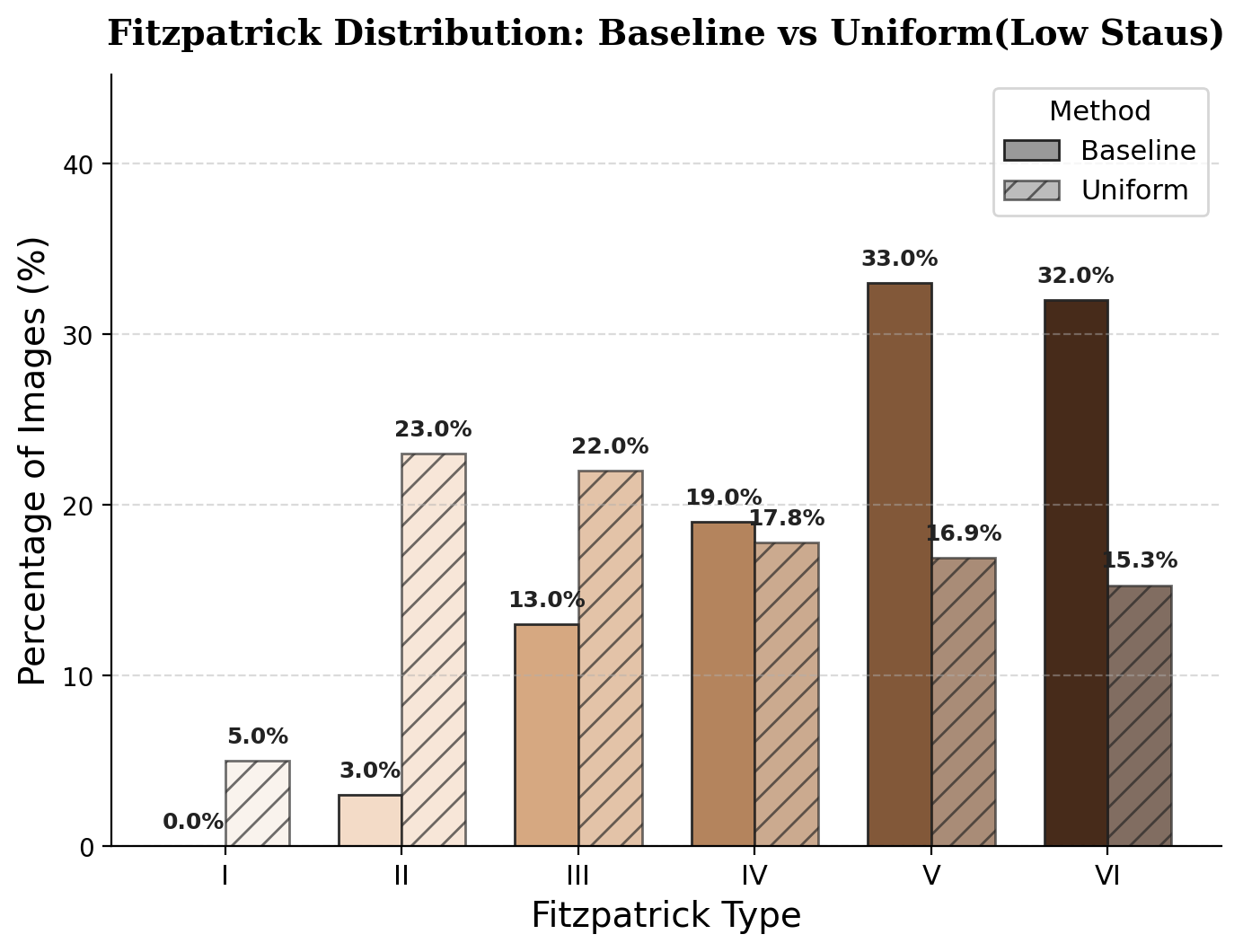}
    \caption{Fitzpatrick skin tone distributions (Types I--VI) across high-, moderate-,
    and low-status occupational groups before (solid) and after (hatched) applying our
    target-conditioned prompting framework. Model: SDXL Turbo. Baseline outputs show a
    pronounced status-linked skew; our uniform target substantially reduces concentration
    at both ends of the scale.}
    \label{fig:appendix_fitz_sdxl}
\end{figure*}

\begin{figure*}[h!]
    \centering
    \includegraphics[width=0.32\linewidth]{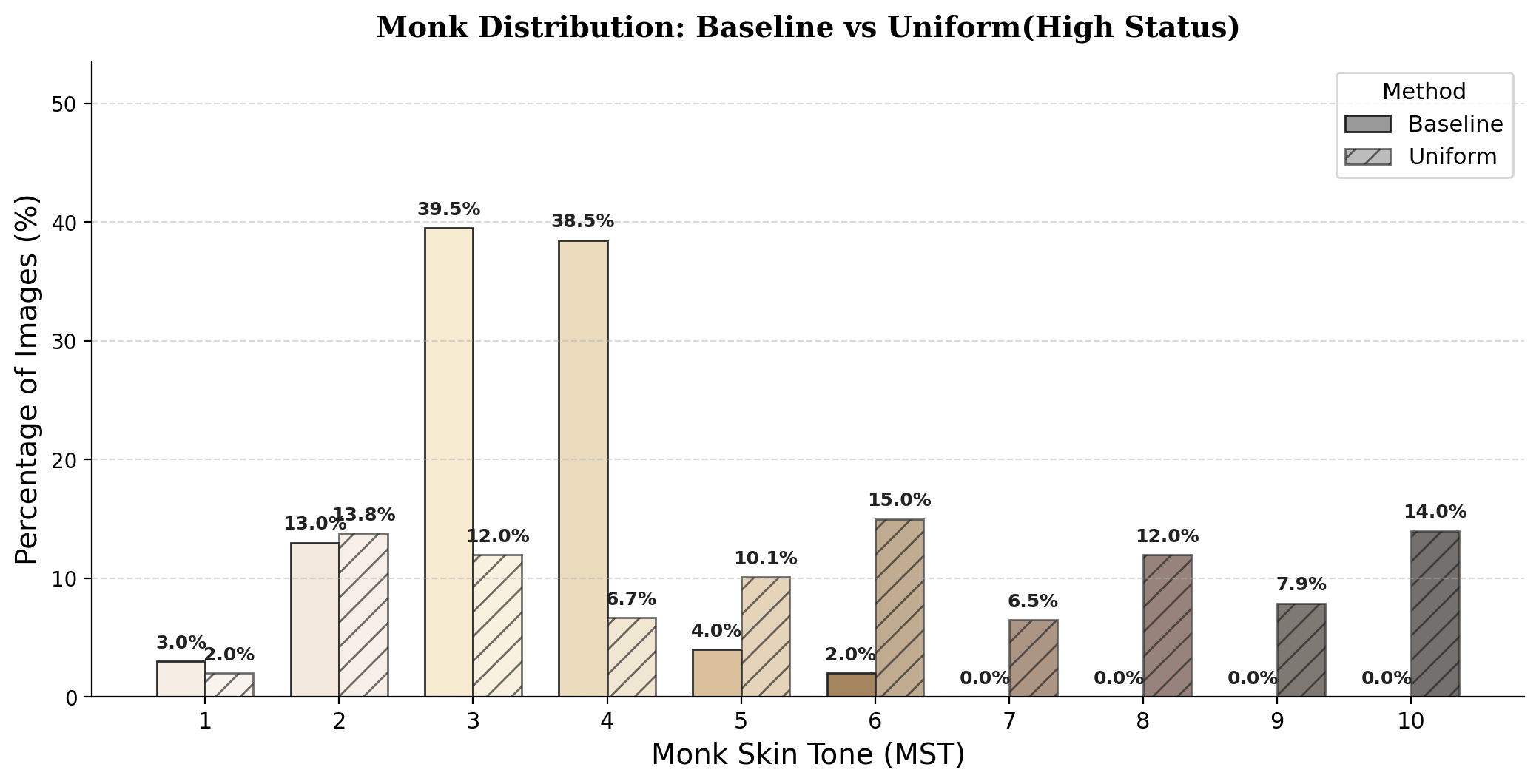}
    \hfill
    \includegraphics[width=0.32\linewidth]{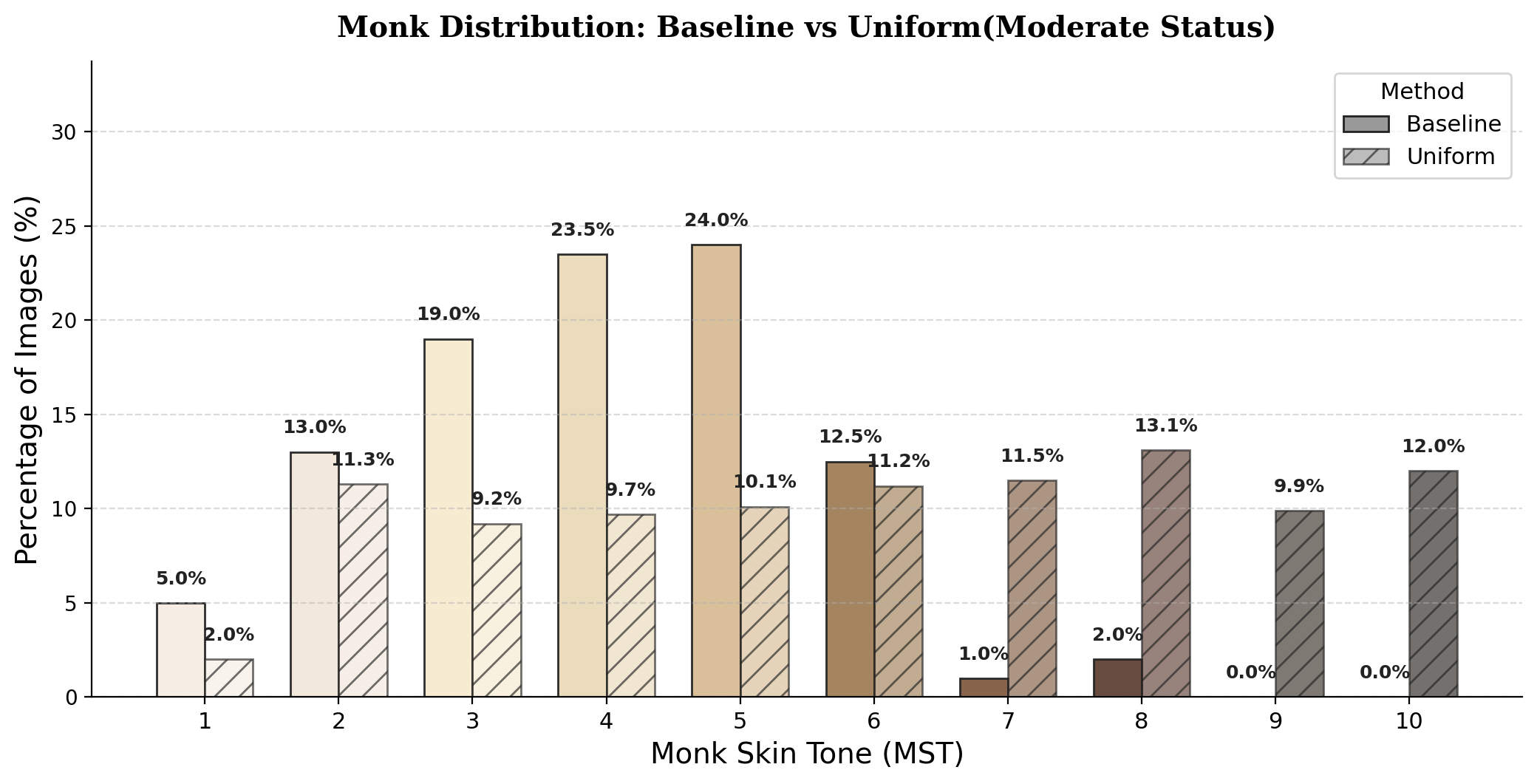}
    \hfill
    \includegraphics[width=0.32\linewidth]{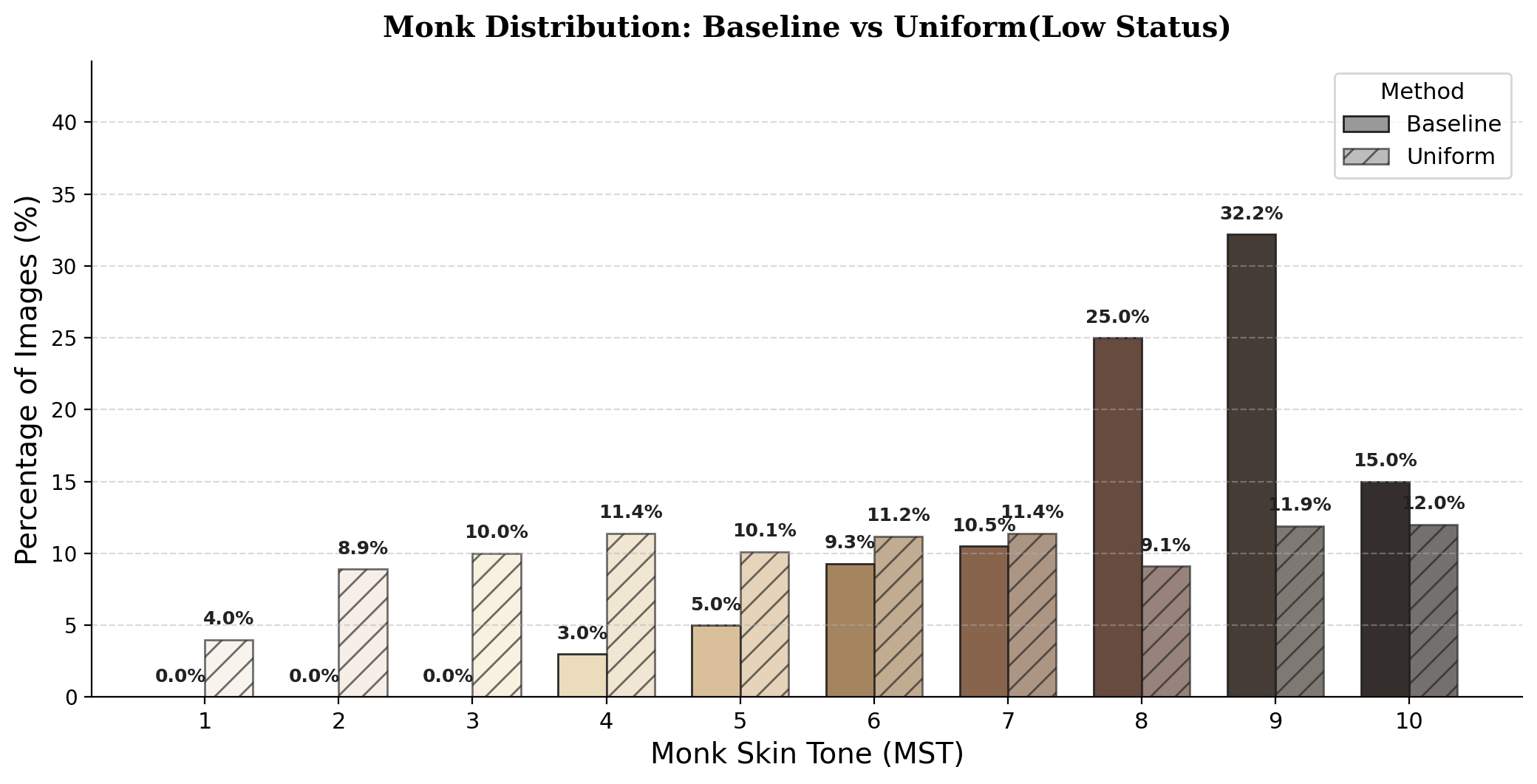}
    \caption{Monk Skin Tone distributions (MST 1--10) across high-, moderate-,
    and low-status occupational groups before (solid) and after (hatched) applying our
    target-conditioned prompting framework. Model: SDXL Turbo. Results are consistent
    with the Fitzpatrick distributions above, confirming that the observed shifts hold
    across both skin tone scales.}
    \label{fig:appendix_monk_sdxl}
\end{figure*}

\begin{figure*}[h!]
    \centering
    \includegraphics[width=0.32\linewidth]{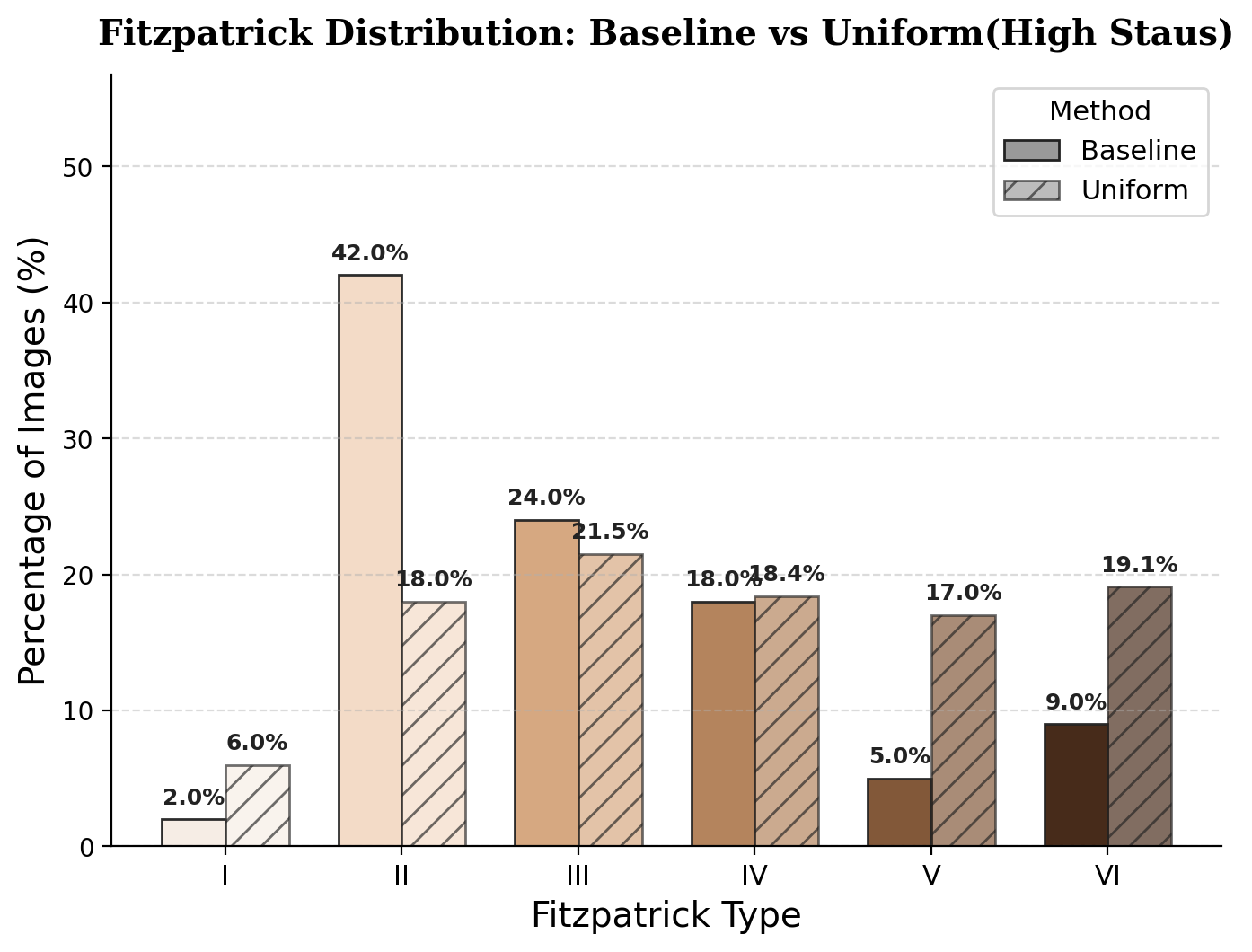}
    \hfill
    \includegraphics[width=0.32\linewidth]{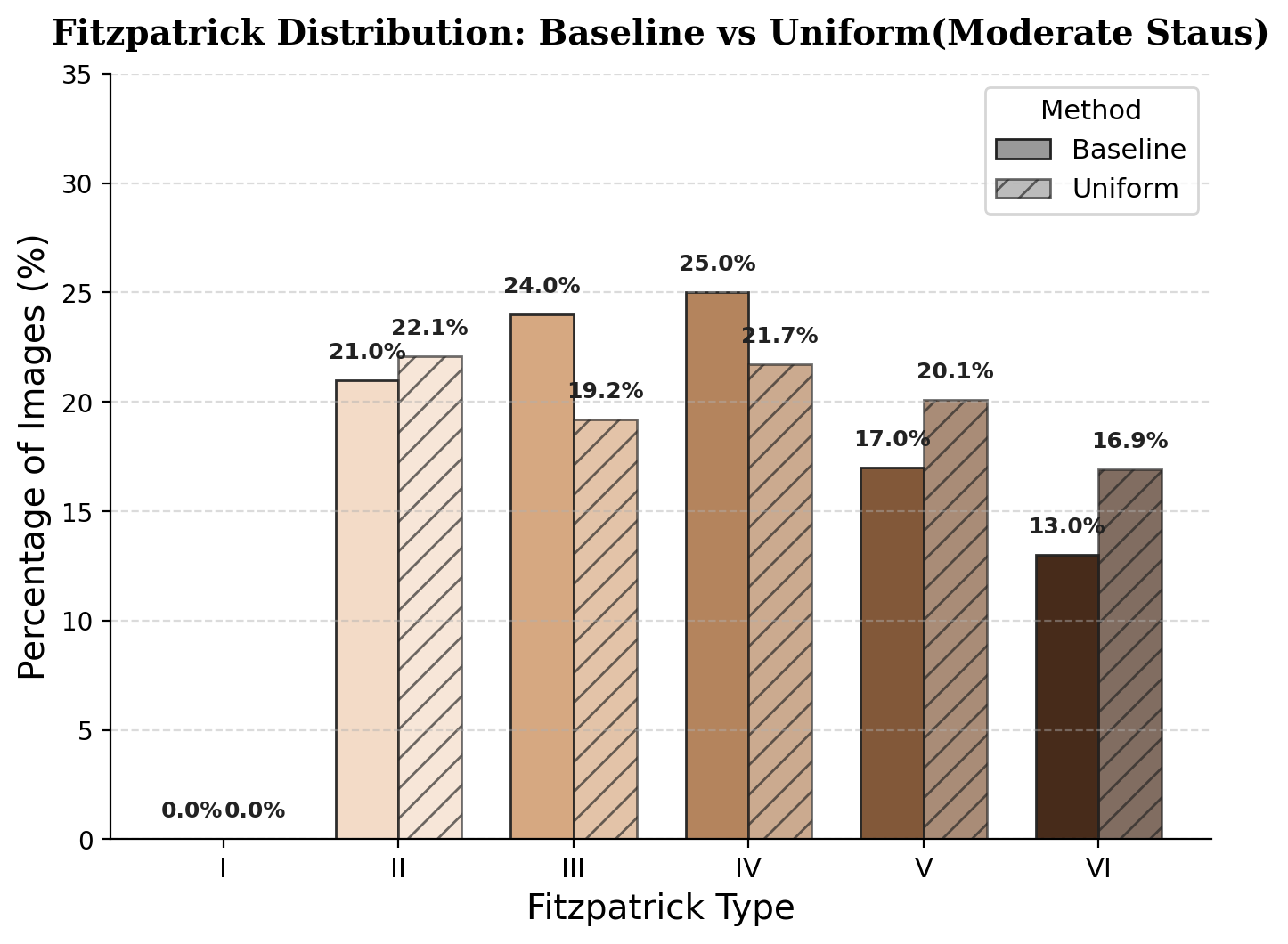}
    \hfill
    \includegraphics[width=0.32\linewidth]{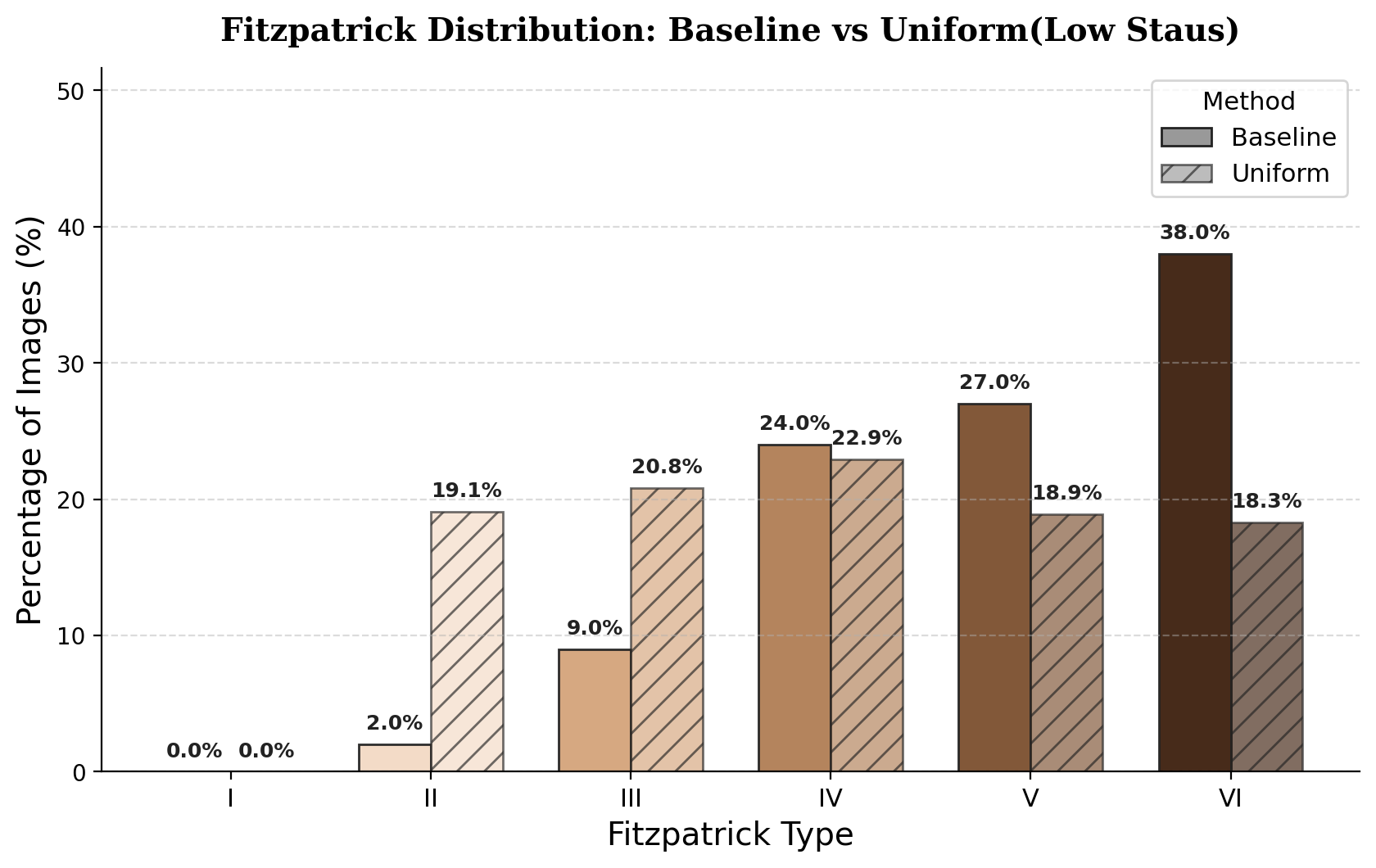}
    \caption{Fitzpatrick skin tone distributions (Types I--VI) across high-, moderate-,
    and low-status occupational groups before (solid) and after (hatched) applying our
    target-conditioned prompting framework. Model: Stable Diffusion 1.5. Baseline outputs
    show a pronounced status-linked skew; our uniform target substantially reduces
    concentration at both ends of the scale.}
    \label{fig:appendix_fitz_sd15}
\end{figure*}

\begin{figure*}[h!]
    \centering
    \includegraphics[width=0.32\linewidth]{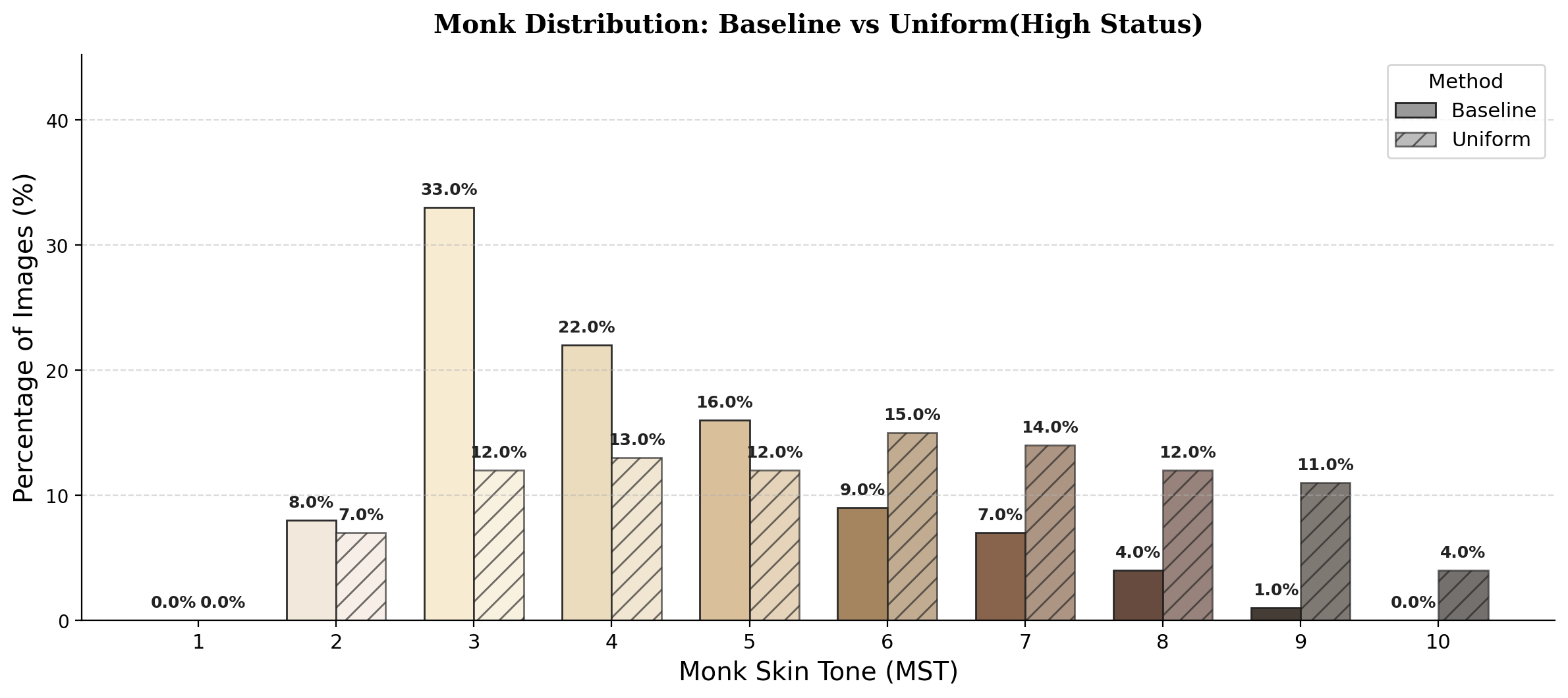}
    \hfill
    \includegraphics[width=0.32\linewidth]{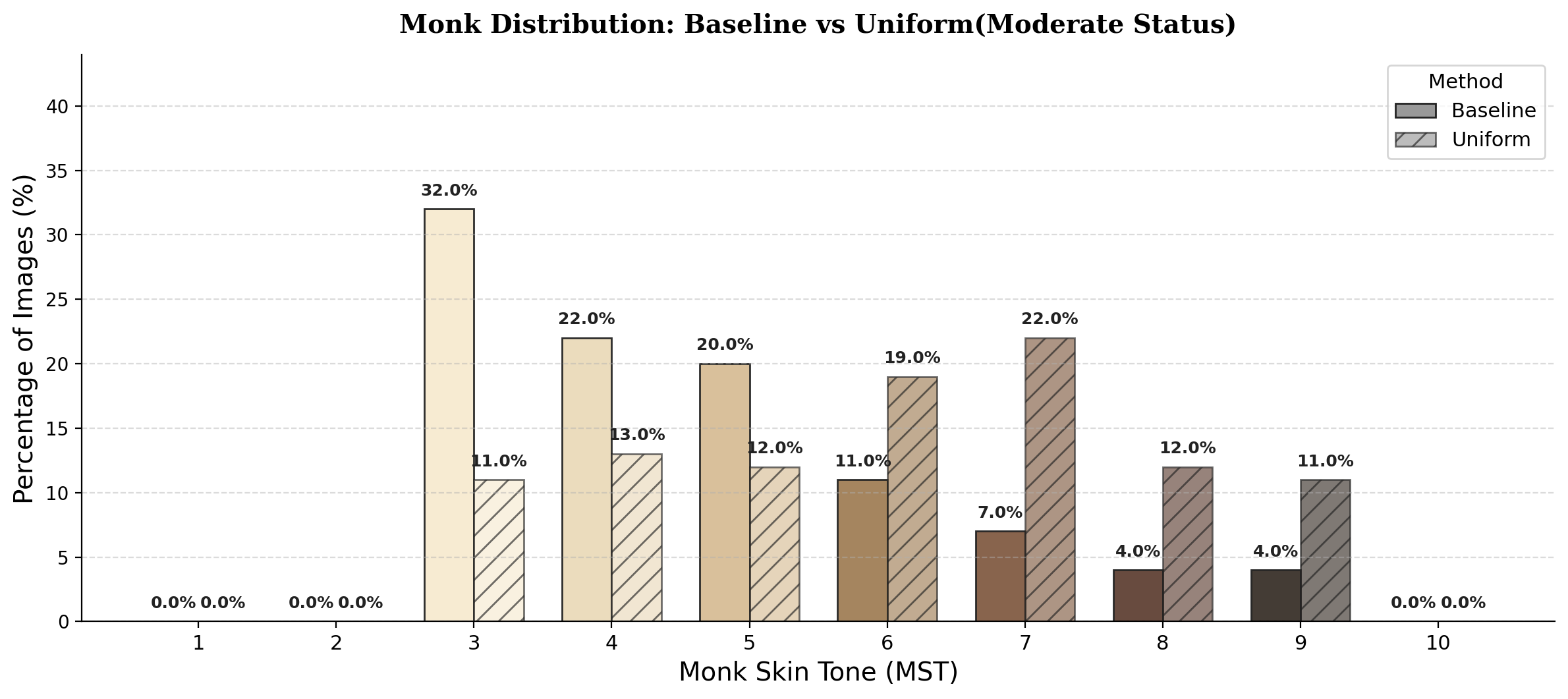}
    \hfill
    \includegraphics[width=0.32\linewidth]{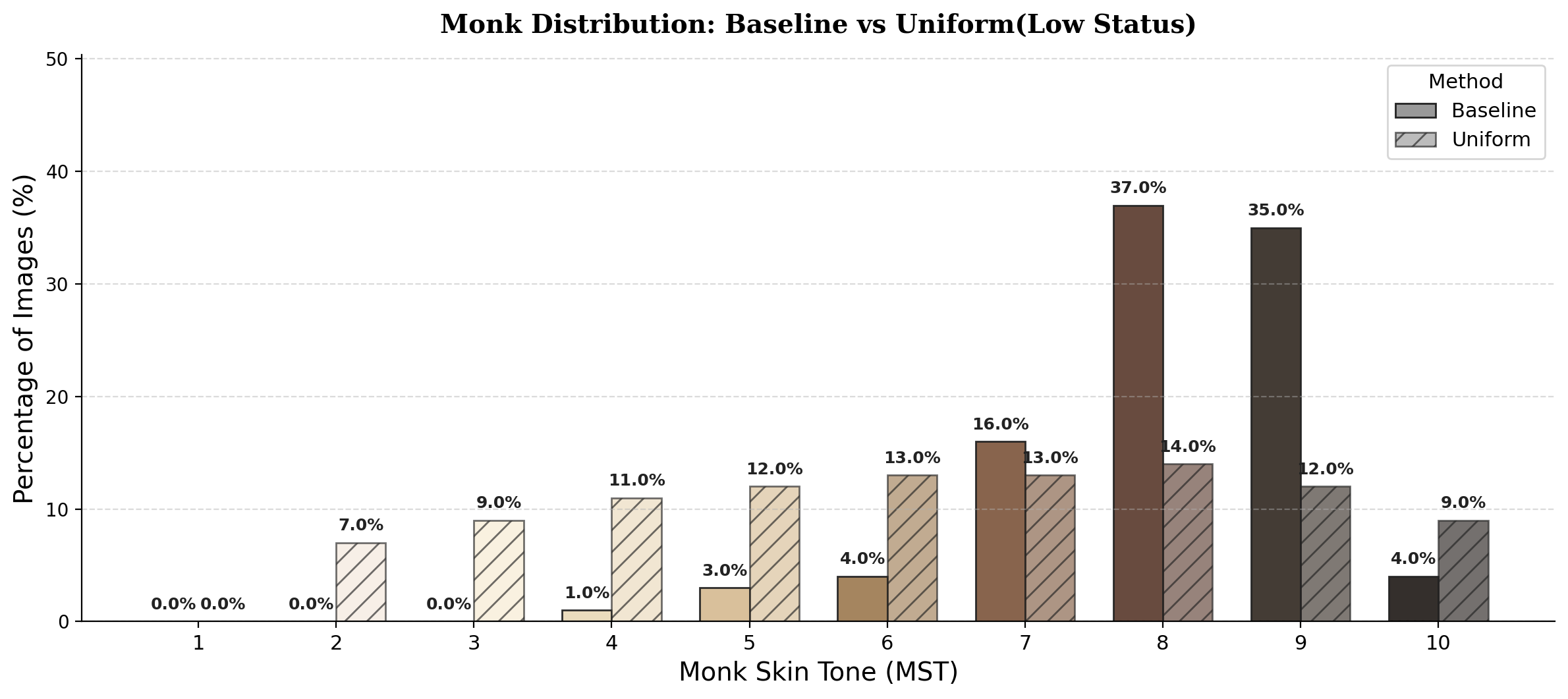}
    \caption{Monk Skin Tone distributions (MST 1--10) across high-, moderate-,
    and low-status occupational groups before (solid) and after (hatched) applying our
    target-conditioned prompting framework. Model: Stable Diffusion 1.5. Results are
    consistent with the Fitzpatrick distributions above, confirming that the observed
    shifts hold across both skin tone scales.}
    \label{fig:appendix_monk_sd15}
\end{figure*}

\begin{figure*}[h!]
    \centering
    \includegraphics[width=0.32\linewidth]{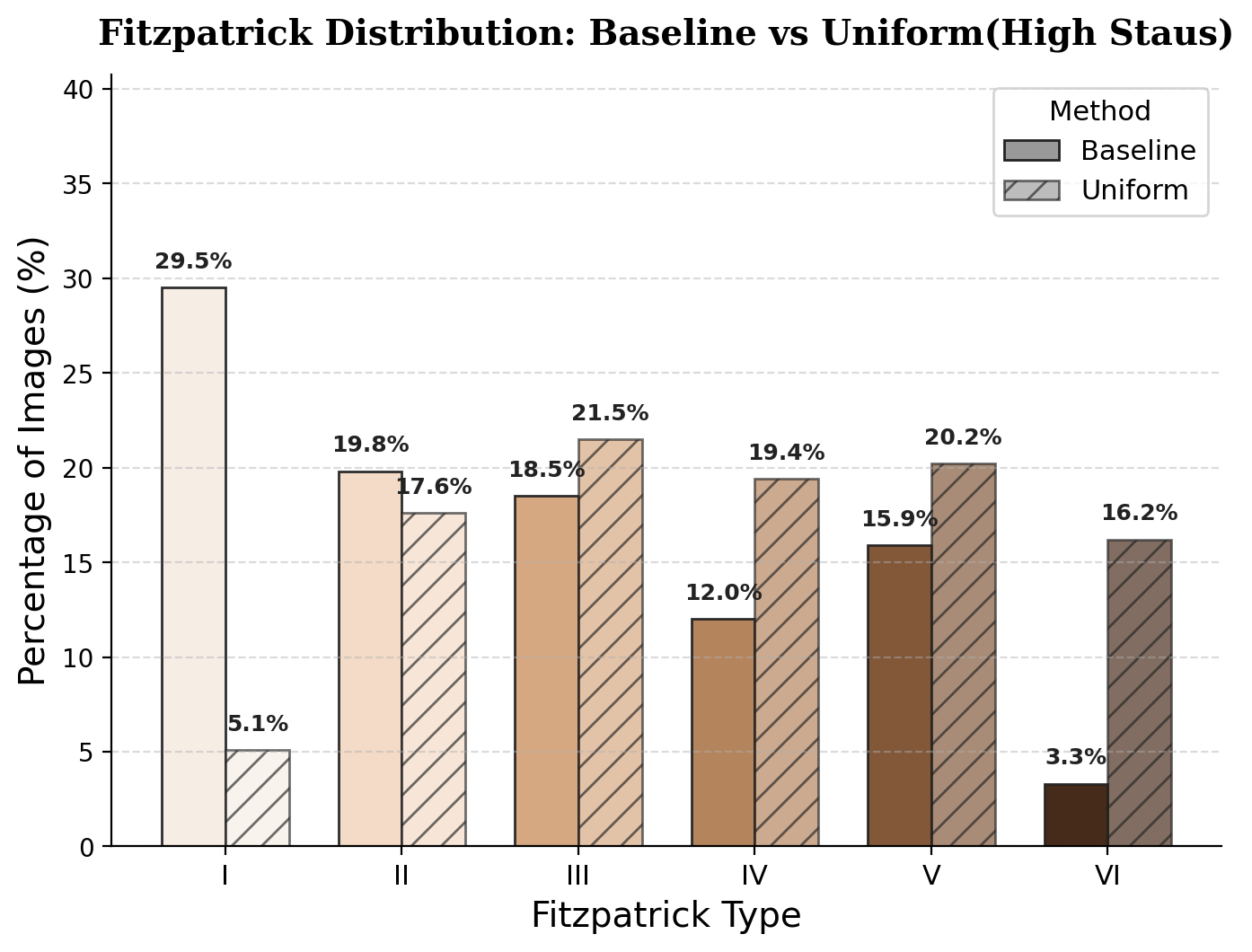}
    \hfill
    \includegraphics[width=0.32\linewidth]{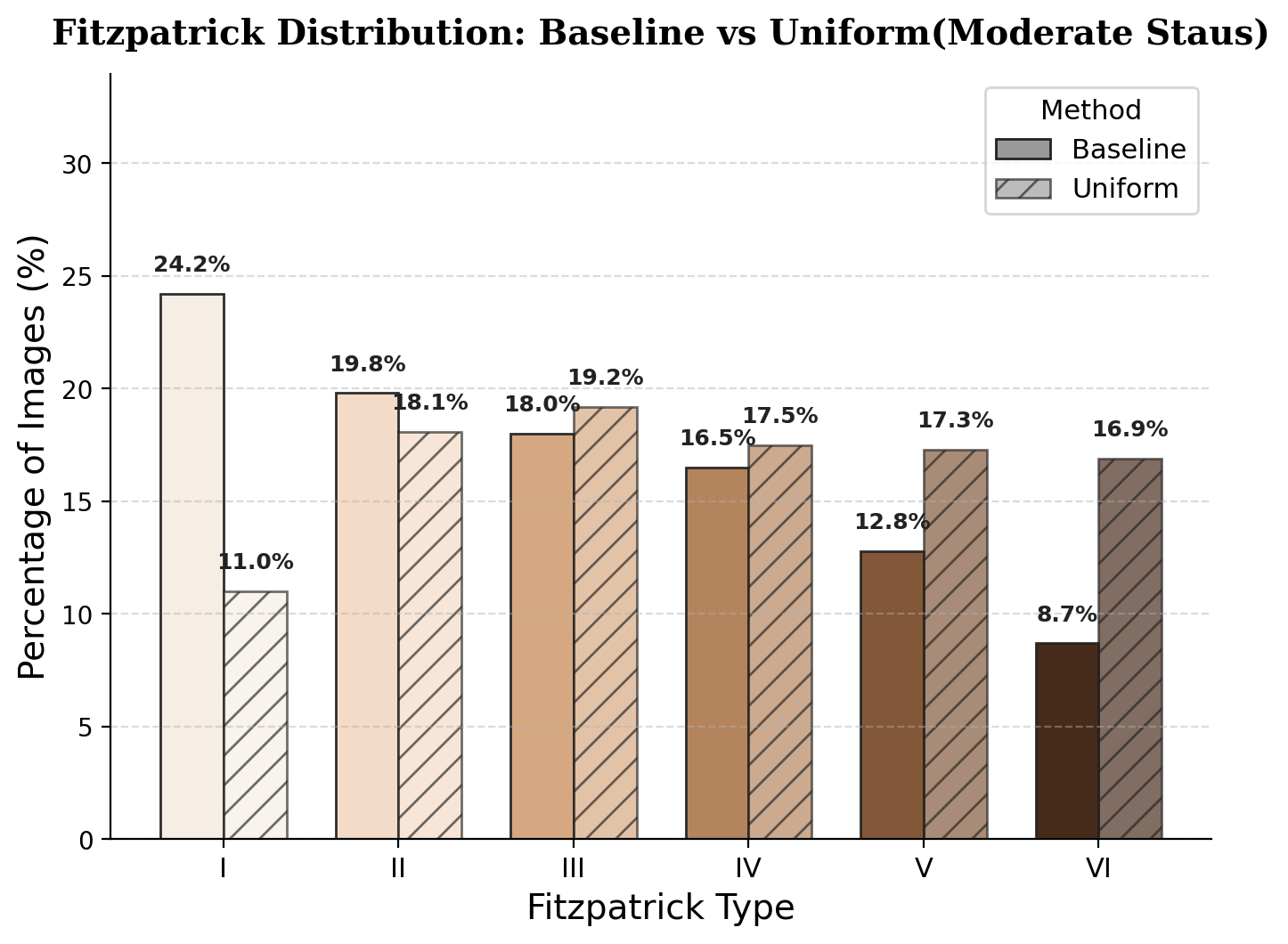}
    \hfill
    \includegraphics[width=0.32\linewidth]{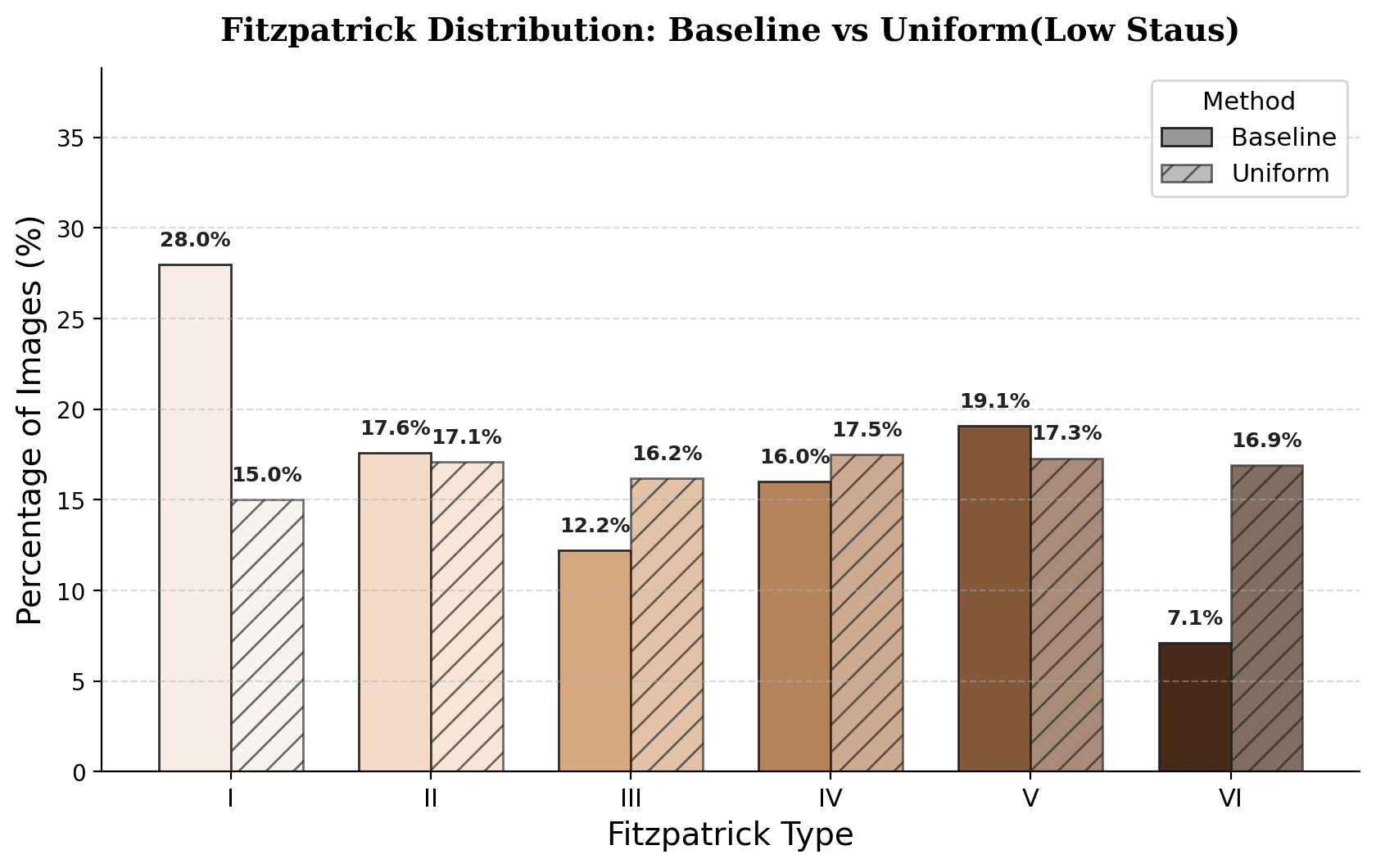}
    \caption{Fitzpatrick skin tone distributions (Types I--VI) across high-, moderate-,
    and low-status occupational groups before (solid) and after (hatched) applying our
    target-conditioned prompting framework. Model: DALL-E~2. Baseline outputs show a
    pronounced status-linked skew; our uniform target substantially reduces concentration
    at both ends of the scale.}
    \label{fig:appendix_fitz_dalle2}
\end{figure*}

\begin{figure*}[h!]
    \centering
    \includegraphics[width=0.32\linewidth]{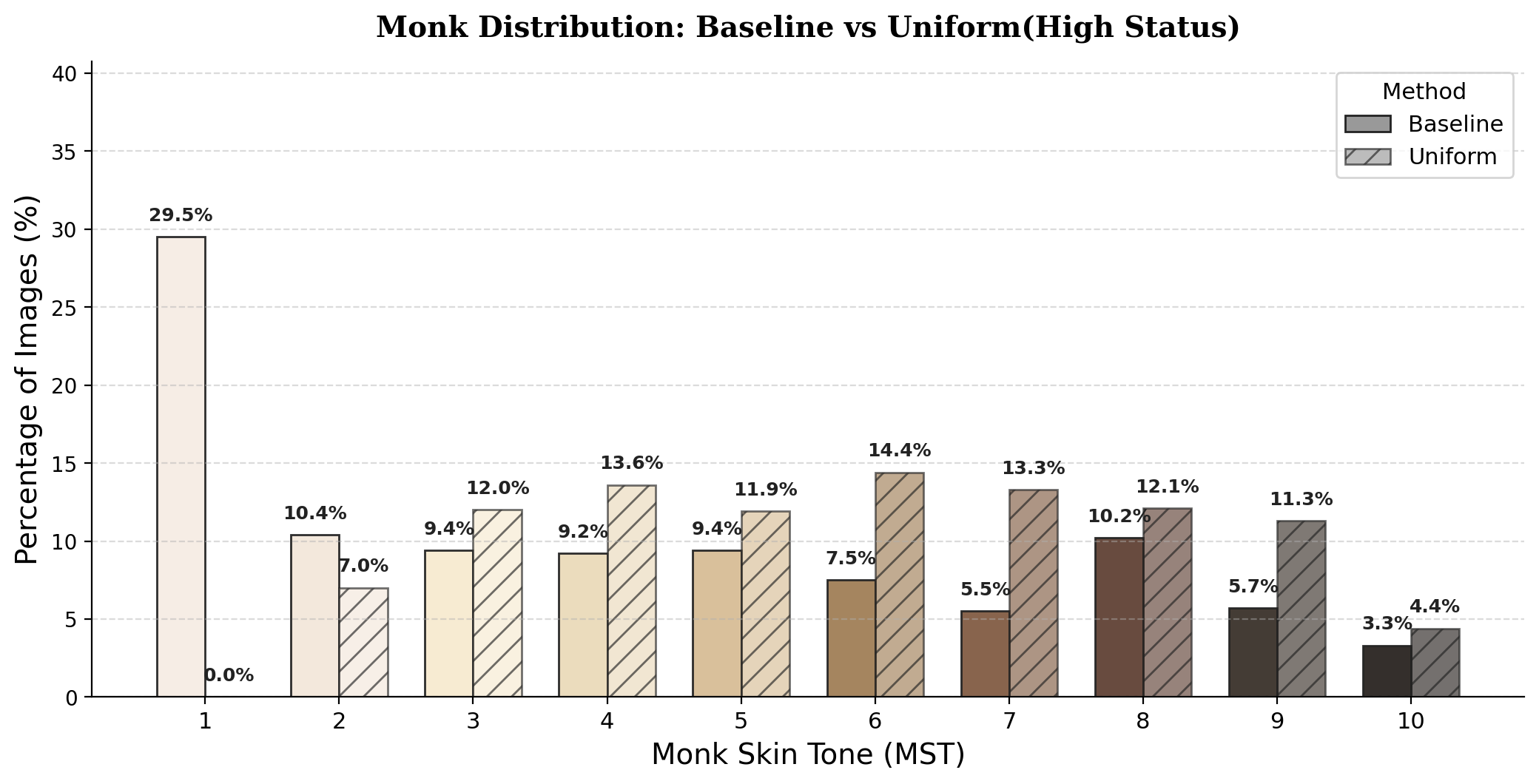}
    \hfill
    \includegraphics[width=0.32\linewidth]{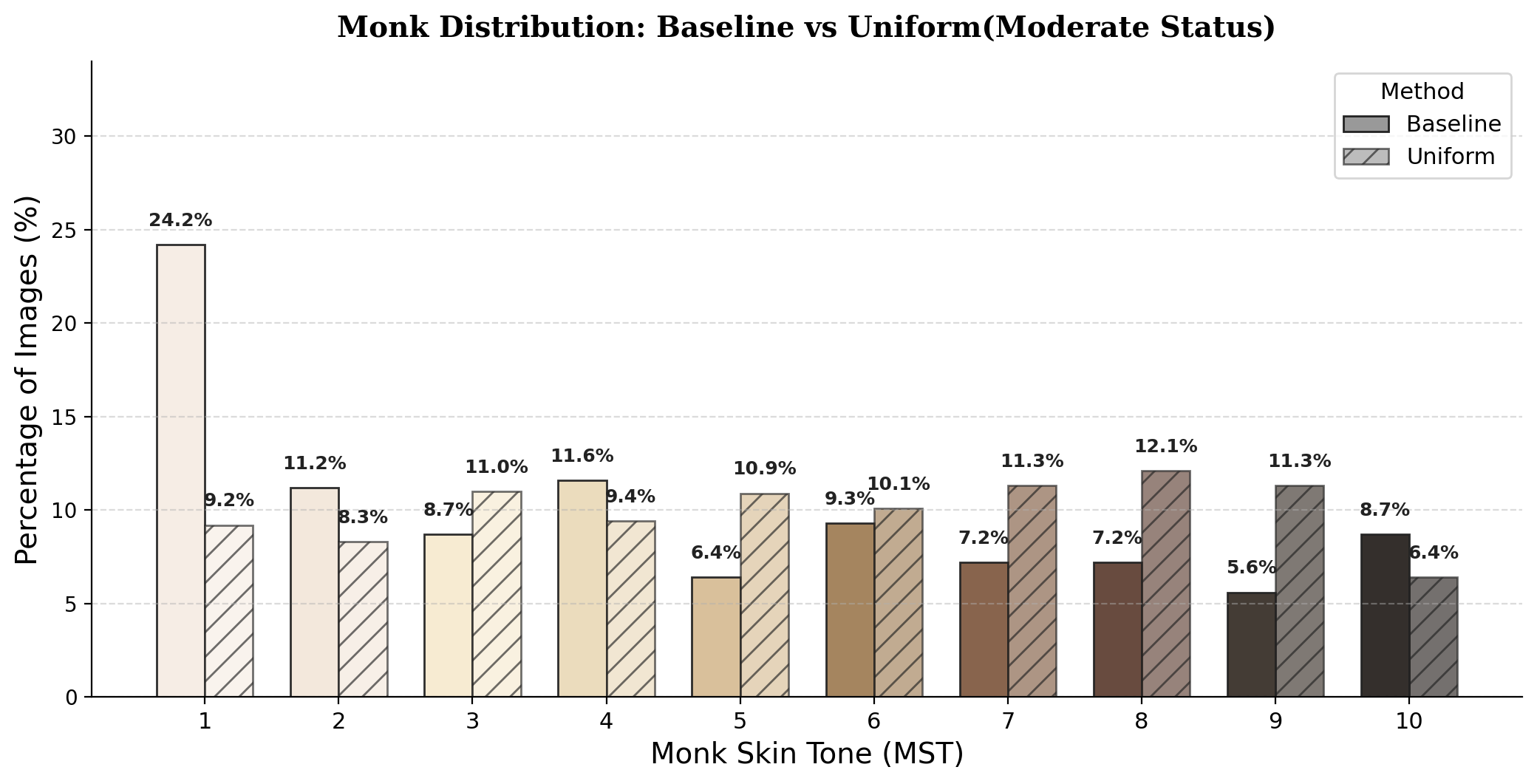}
    \hfill
    \includegraphics[width=0.32\linewidth]{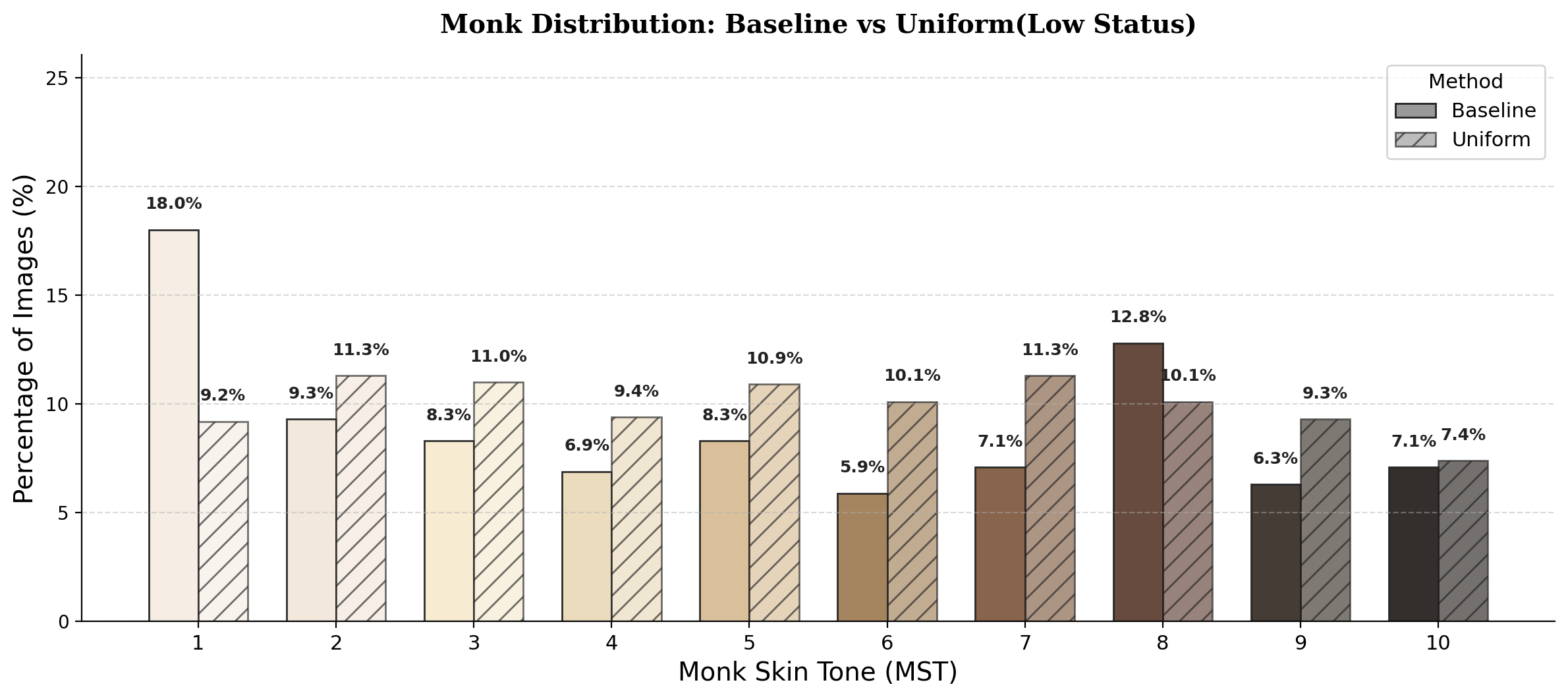}
    \caption{Monk Skin Tone distributions (MST 1--10) across high-, moderate-,
    and low-status occupational groups before (solid) and after (hatched) applying our
    target-conditioned prompting framework. Model: DALL-E~2. Results are consistent with
    the Fitzpatrick distributions above, confirming that the observed shifts hold across
    both skin tone scales.}
    \label{fig:appendix_monk_dalle}
\end{figure*}

Figures~\ref{fig:appendix_monk_rv}--\ref{fig:appendix_monk_dalle} show Fitzpatrick
and Monk Skin Tone (MST)~\cite{google_skintone_scale} distributions before and after
applying our target-conditioned prompting framework, across SD Realistic Vision v5.1,
SDXL Turbo, SD~1.5, and DALL-E~2. For SD Realistic Vision v5.1, the Fitzpatrick
distributions are shown in Figure~\ref{fig:status_fitz_charts} of the main paper;
Figure~\ref{fig:appendix_monk_rv} provides the corresponding MST distributions. For
each model, we report the Fitzpatrick scale (Types I--VI) and the Monk Skin Tone scale
(MST 1--10). Both scales consistently show that baseline outputs are concentrated toward
lighter tones for high-status occupations and darker tones for low-status occupations,
while our uniform target substantially reduces these skews across all status groups.

\clearpage

\section{Qualitative Comparison Across Inference-Time Methods}
\label{app:qualitative_comparison}

\begin{figure*}[t]
    \centering
    \includegraphics[width=\linewidth,height=0.46\textheight,keepaspectratio]{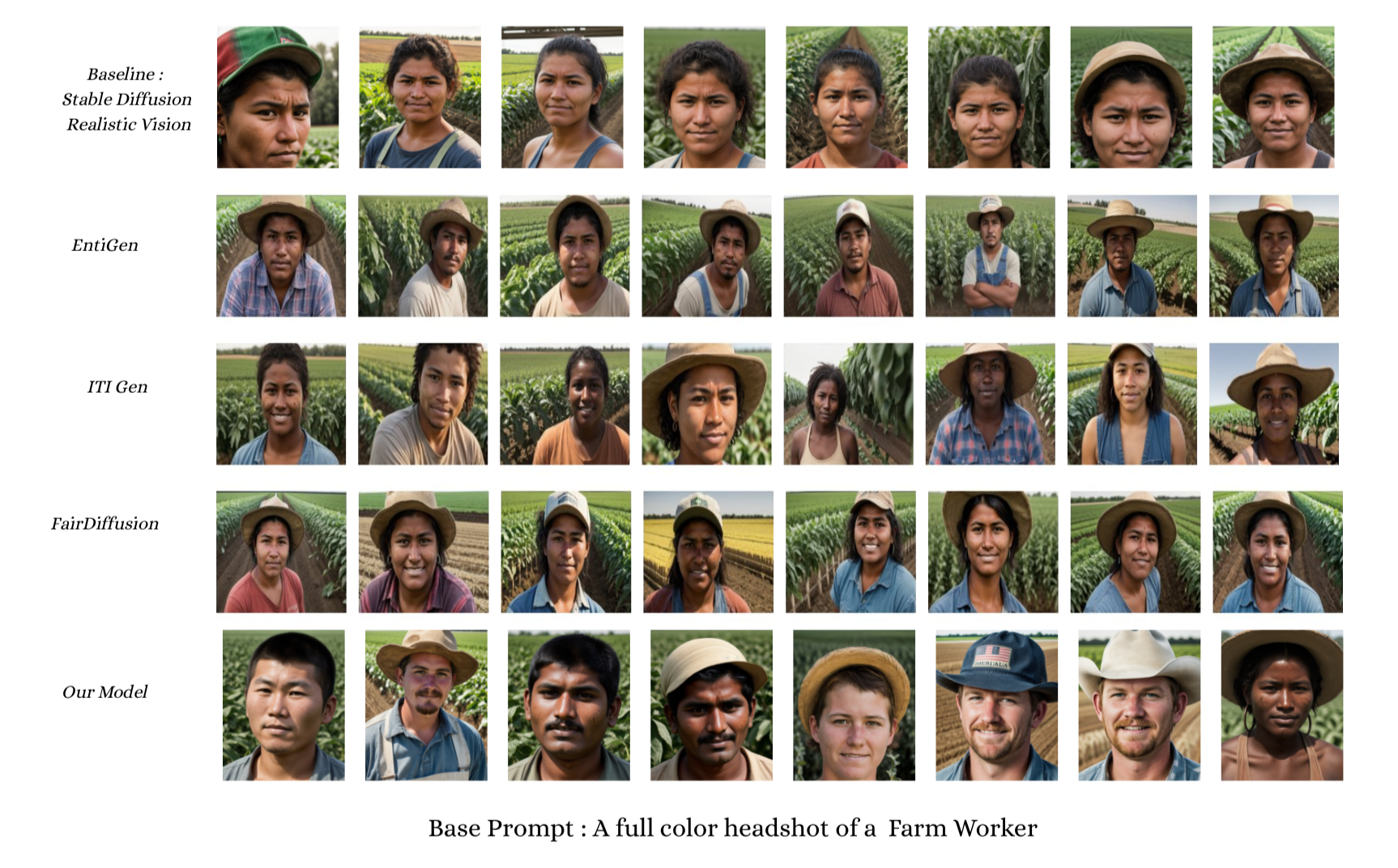}
    \vspace{-0.9em}
    \includegraphics[width=\linewidth,height=0.48\textheight,keepaspectratio]{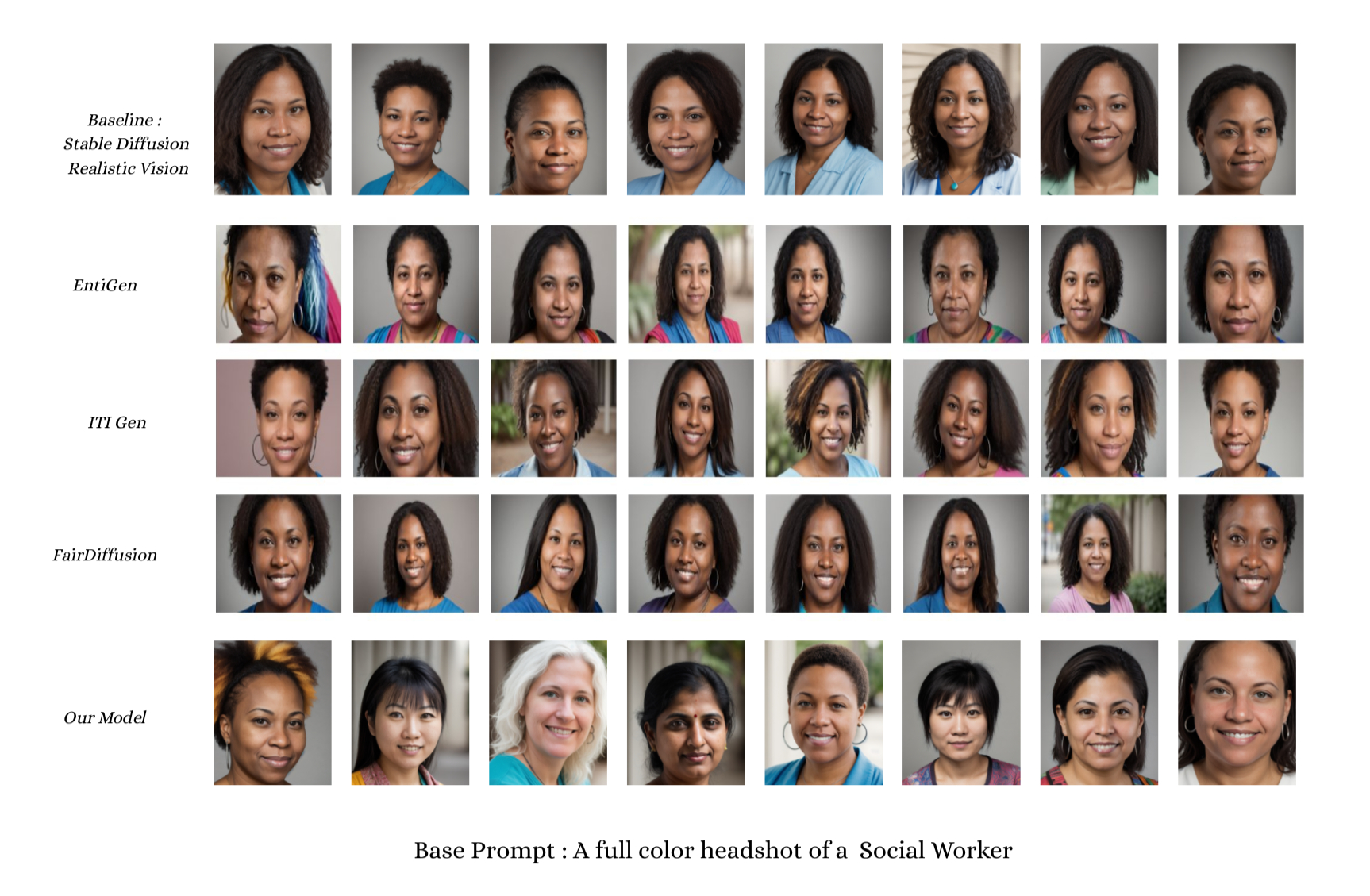}
    \caption{Qualitative comparison across all methods for  occupational prompts
        using SD Realistic Vision v5.1:
        \textit{``A full-color headshot of a Farm Worker''} and\textit{``A full-color headshot of a Social Worker''}.
        Each row shows outputs from one method:
        Row~1: Baseline;
        Row~2: EntiGen~\cite{bansal2022Enti};
        Row~3: ITI-GEN~\cite{zhang2023iti};
        Row~4: Fair Diffusion~\cite{friedrich2023fair};
        Row~5: Ours (Uniform target).
        }
    \label{fig:qualitative_comparison_fw_sw}
\end{figure*}

Figure~\ref{fig:qualitative_comparison_fw_sw}provide a qualitative 
comparison of all inference-time methods evaluated in 
Section~\ref{subsec:baselines}, using SD Realistic Vision v5.1 across three
occupational prompts for -, moderate-, and low-status roles.


\end{document}